\documentclass[10pt]{article} 
\usepackage[accepted]{tmlr}%%

\usepackage{bbm}
\usepackage{url}
\usepackage{pdflscape}
\usepackage{afterpage}
\usepackage{rotating}
\usepackage{array}
\usepackage{colortbl}
\usepackage{float}
\usepackage{hyperref}

\usepackage{etoc}
\usepackage{booktabs}
\usepackage{amsmath}
\usepackage{multirow}
\DeclareMathOperator*{\argmax}{arg\,max}

\definecolor{red_pastel}{RGB}{210, 42, 42} 
\definecolor{blue_pastel}{RGB}{4, 54, 183} 
\definecolor{blue_mpl}{RGB}{31, 119, 180} 
\definecolor{orange_mpl}{RGB}{255, 165, 0}

\usepackage{pifont}% http://ctan.org/pkg/pifont
\newcommand{\cmark}{\ding{51}}%
\usepackage{subfig}

\def\month{08}
\def\year{2026}
\def\openreview{\url{https://openreview.net/forum?id=J7EIFs4sjx}}

\title{\textsc{Alert}: Learning Trigger Functions 
for Early Classification of Time Series using Deep-RL}
\author{\name Aur\'elien Renault \email aurelien.renault@orange.com \\
\addr Orange Research \\%, Châtillon, France \\
\addr AgroParisTech %, Palaiseau, France \\
\AND
\name Alexis Bondu \email alexis.bondu@orange.com \\
\addr Orange Research %, Châtillon, France \\
\AND
\name Antoine Cornu\'ejols \email antoine.cornuejols@agroparistech.fr \\
\addr AgroParisTech %, Palaiseau, France\\
\AND 
\name Vincent Lemaire \email vincent.lemaire@orange.com \\
\addr Orange Research} %, Lannion, France

\begin{document}

\maketitle
\begin{abstract}
Early Classification of Time Series (ECTS) is vital in fields like industrial monitoring and medical triage, where quick and accurate predictions are essential. One of the core challenges lies in the trigger function, which decides when to make a prediction, independently of the classifier itself. Most existing methods rely on handcrafted rules, but can data-driven approaches outperform them? This paper introduces \textsc{Alert}, a Deep-RL framework that learns trigger functions from any state representation. Systematic comparisons on 30 datasets show that the design of the state space significantly influences performance. Building on this, we propose \textsc{Alert}$^+$, a simple yet effective variant that consistently outperforms traditional methods in balancing accuracy and delay within an imbalanced misclassification and exponential delay cost setting. \textsc{Alert} and \textsc{Alert}$^+$ are released to support reproducible research and practical applications.
\end{abstract}

\section{Introduction}

Many data-driven decision-making systems operate under \emph{time pressure}, where delaying a prediction can be almost as costly as making an incorrect one. This tension is particularly acute in applications such as real-time monitoring of industrial assets, predictive maintenance, or medical triage, where acting too late may lead to equipment failure, safety incidents, or adverse clinical outcomes. In such settings, the central question is not only \emph{what} to predict, but also \emph{when} to issue a prediction.

Early Classification of Time Series (ECTS) \citep{gupta2020approaches, akasiadis2024framework, renault2024early} explicitly addresses this joint optimization of predictive accuracy and decision timeliness, and belongs to the broader class of early decision-making~\citep{bondu2022open} and anytime prediction problems. A widely adopted and practically attractive way to tackle ECTS is through \emph{separable architectures}: a time series classifier produces a sequence of probabilistic predictions over time, and a \emph{trigger function} decides when the current prediction should be released. This separation is appealing because it decouples the design of powerful classifiers~\citep{bagnall2017great, ruiz2021great} from the design of decision policies that trade off misclassification and delay costs.

\begin{figure}[!th]
    \centering
    \includegraphics[width=0.5\linewidth]{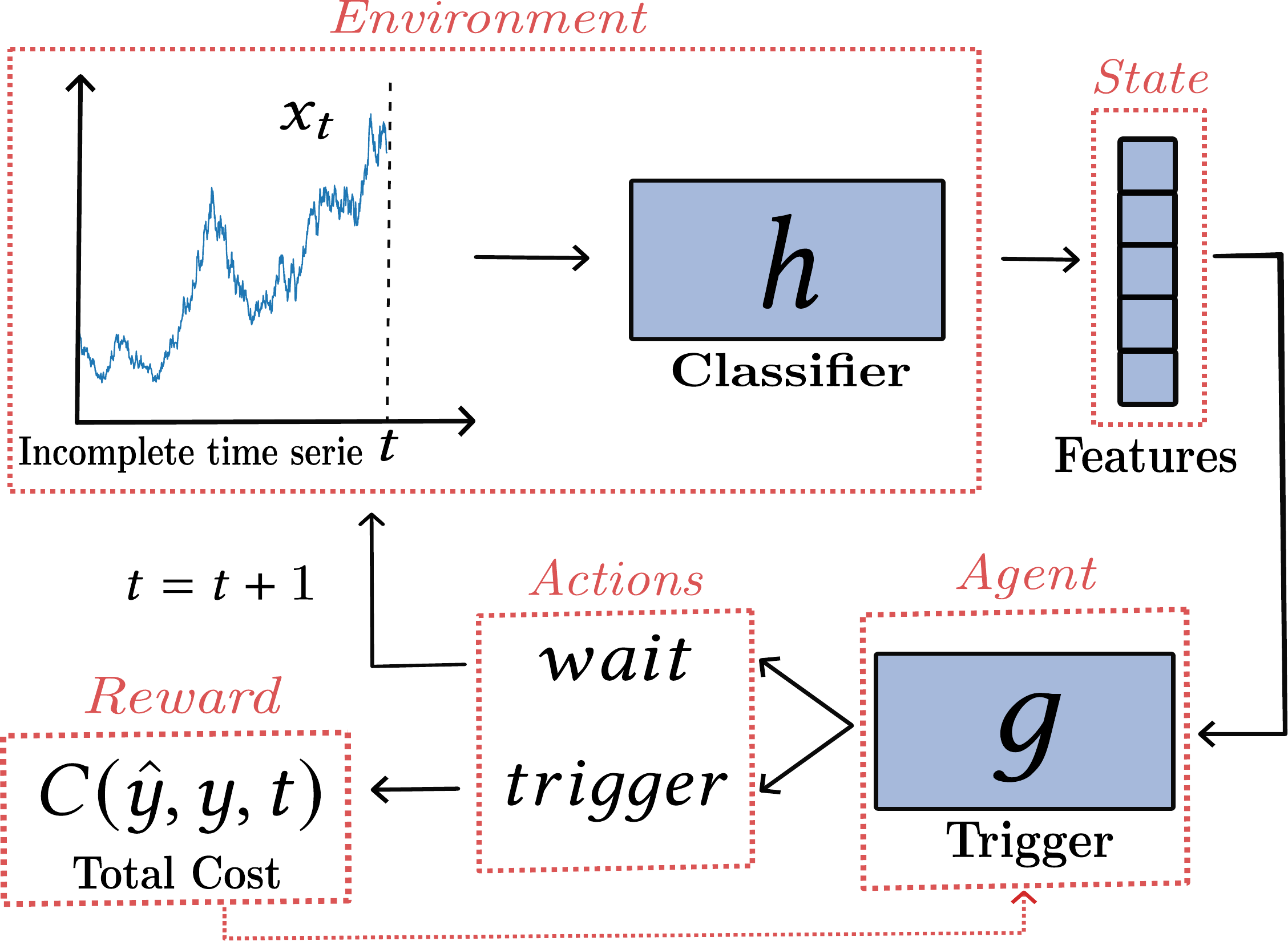}
    \caption{Separable ECTS pipeline and corresponding RL terminology (written in \textcolor{red_pastel}{red}).}
    \label{fig:ects_pipe}
\end{figure}

Most existing separable ECTS methods rely on carefully engineered trigger functions. These range from simple confidence thresholds and stopping rules~\citep{mori2017early, schafer2020teaser}, to more involved cost-based or non-myopic strategies that estimate future misclassification and delay costs~\citep{dachraoui2015early, achenchabe2021early, tavenard2016cost}. While often effective, these approaches are typically predefined and inflexible: they embed strong assumptions about how confidence, time, or expected costs should interact, and they are tailored to specific choices of features (e.g., maximum posterior, margin, or time index). As a result, they raise two key questions:
\begin{enumerate}
    \item Can a \emph{data-driven decision policy} systematically outperform handcrafted triggers when given access to the same information?
    \item How should the \emph{state representation} fed to such a policy be designed in order to leverage the information produced by a classifier most effectively?
\end{enumerate}

Reinforcement Learning (RL) offers a principled framework for sequential decision-making under uncertainty, and appears particularly well-suited to ECTS (see Figure \ref{fig:ects_pipe} for a terminology mapping). At each time step, a triggering agent must decide whether to \emph{wait} for additional observations, incurring further delay costs but potentially improving future classification accuracy, or to \emph{stop} and commit to the current prediction, locking in a misclassification cost if the prediction is wrong. 
This naturally matches the RL formalism, with the state space capturing information about the time series and classifier outputs, actions being $\mathcal{A} = \{\textit{wait}, \textit{trigger}\}$, and rewards reflecting the combined misclassification and delay costs.

However, prior RL-based approaches to ECTS  are typically end-to-end~\citep{martinez2018deep, martinez2020adaptive, wang2024deep, hartvigsen2019adaptive} and tightly couple the triggering policy with a specific encoder or classifier~\citep{hartvigsen2020recurrent, hartvigsen2022stop, huang2022snippet1, huang2022snippet}. This coupling makes it impossible to isolate the impact of the trigger policy itself, and leads to inherently biased comparisons.

In this paper, we bridge this gap by studying RL for \emph{separable} and \emph{classifier-agnostic} ECTS. We introduce \textbf{Alert} (\textbf{A} reinforcement \textbf{L}earning-based \textbf{E}arly classifie\textbf{R}'s \textbf{T}rigger function), a generic Deep-RL framework that learns trigger functions that operate on arbitrary state representations derived from classifier outputs, and potentially on the incoming time series itself. This design enables controlled and fair comparisons, in which RL-based triggers and state-of-the-art handcrafted methods are given exactly the same input information.

Our contributions are threefold:
\begin{description}
  \item \textbf{(C1) A generic RL framework for separable, classifier-agnostic ECTS.} 
  We propose \textsc{Alert}, a Deep-RL framework that learns trigger functions on top of any pre-trained classifier and enables controlled comparisons where RL-based and handcrafted separable methods operate on exactly the same input features and cost functions.
  
  \item \textbf{(C2) A systematic study of state space design for RL-based triggers.}
  Across 30 datasets, we vary the state representation and observe substantial performance differences between alternative designs. This analysis clarifies which signals tend to be most informative for early decision policies and how they interact with different triggering mechanisms.
  
  \item \textcolor{black}{\textbf{(C3) \textsc{Alert}$^+$: a simple RL-based trigger.}
  By combining a small set of standard ECTS features, we obtain \textsc{Alert}$^+$, which achieves the best average rank across our benchmarks and often yields  lower  empirical risk than the strongest separable methods, with statistically significant improvements, depending on the cost setting (see Section~\ref{sec_exp_results}). A Pareto analysis over prediction accuracy and earliness further suggests that \textsc{Alert}$^+$ offers a favorable trade-off over a substantial part of the misclassification--delay spectrum.}
  %\textcolor{red}{By combining a small set of standard ECTS features, we obtain \textsc{Alert}$^+$, which consistently attains the best average rank across our benchmarks and significantly reduces empirical risk compared to the strongest separable methods on a majority of datasets (see Section~\ref{sec_exp_results} for statistical tests). A Pareto analysis over the two conflicting objectives of prediction accuracy and earliness further shows that \textsc{Alert}$^+$ dominates competitors on a large portion of the misclassification--delay trade-off spectrum.}
\end{description}

Beyond raw performance improvements, \textsc{Alert} serves as a unified and modular testbed for ECTS. By explicitly separating the classifier from the triggering policy, it allows us to attribute performance gains either to better predictors or to better decision rules, and to systematically study how state representations, reward designs, and training schemes impact results. The released \textsc{Alert} implementation available at \url{https://github.com/aurelien-renault/ALERT} is intended both as a reproducible research tool and as a practical asset for practitioners seeking to tune trigger functions for their own time series applications.

The remainder of the paper is organized as follows. Section~\ref{background} reviews background on RL in the context of ECTS and summarizes separable ECTS methods, with a focus on the information used to trigger decisions. Section~\ref{sec:proposed_methodology} introduces the \textsc{Alert} framework and discusses its components: state space, reward design, offline training, and model selection. Section~\ref{sec_exp_protocol} details the experimental protocol, including datasets, cost functions, and baselines. Section~\ref{sec_exp_results} presents the empirical results, analyzing the impact of state space design and comparing \textsc{Alert+} against state-of-the-art methods. Section~\ref{sec:conclusion} concludes with a discussion of implications and directions for future RL-based early decision-making systems.

\section{Background}
\label{background}

In ECTS, measurements of an input time series are observed over time. At any time $t < T$, before the last time step $T$, the incomplete time series ${\mathbf x}_t \, = \, \langle {x_1}, \ldots, {x_t} \rangle$ is available where $\{x_i\}_{(1 \leq i \leq t)}$ denotes the time indexed measurements (single or multi-valued). Each input time series belongs to an unknown class $y \in {\mathcal{Y}}$. The task is to make a prediction $\hat{y} \in {\mathcal{Y}}$ 
%about the class of the incoming time series, 
ideally at a time $t^\star \in [1,T]$ 
which optimizes a trade-off between: \textit{(i)} the \textit{misclassification cost} $\mathrm{C}_m(\hat{y}|y): {\mathcal{Y}} \times {\mathcal{Y}} \rightarrow \mathbbm{R}$ of predicting $\hat{y}$ when the true class is $y$; \textit{(ii)} the \textit{delay cost} $\mathrm{C}_d(t) : \mathbbm{R}^+ \rightarrow \mathbbm{R}$ of releasing the prediction at $t \in [1,T]$.

This section presents the background related to both RL and ECTS.  Specifically, in Section \ref{Adapting_RL_to_ECTS}, we begin by discussing the use of RL in a separable ECTS context before examining the existing literature on RL for ECTS.
%, which is not both \textit{separable} and \textit{classifier-agnostic}. 
Then, in Section \ref{sota_ects}, separable ECTS methods are discussed in light of the input information they use to trigger decisions.

\subsection{Adapting RL to ECTS}
\label{Adapting_RL_to_ECTS}

Reinforcement learning \citep{sutton2018reinforcement} aims to learn a function, called a \textit{policy} $\pi$, from states to actions: $\pi: {\mathcal{S}} \rightarrow {\mathcal{A}}$. \textit{Rewards} can be associated with transitions from states $s_t \in {\mathcal{S}}$ to states $s_{t+1} \in {\mathcal{S}}$ under an \textit{action} $a \in {\mathcal{A}}$: $reward(s_t, a, s_{t+1}) \in \mathbbm{R}$. 
%This gain, denoted $G_t$ starting from the state $s_t$, is defined as a function of the rewards from that state (e.g. a discounted sum of the rewards received).  
Given a state $s_t$ and an action $a_t$, the sequence of rewards received after time step $t$ gives rise to a \textit{gain} which classically is a discounted sum of the rewards: $G_t=\sum_{k=0}^\infty \gamma^k r_{t+k+1}$ with $\gamma \in [0, 1]$ the discount factor and $r_{t} = reward(s_t, a_t, s_{t+1})$. %\textcolor{magenta}{k=??}
In all generality, the result of an action $a$ in state $s_t$ may be non-deterministic. An optimal policy $\pi^\star$  maximizes the expected gain from any state $s_t \in {\mathcal{S}}$.  

In the context of separable ECTS, the agent must learn which \textit{action} to take (i.e. decision ``\textit{wait}'' or ``\textit{trigger}'' prediction $\hat{y}$) given its current \textit{state}, i.e. any information provided about the incoming time series $\mathbf{x}_t$ and/or the classifier.
An \textit{episode} is defined as the sequence of states $s_t$ and actions $a_t$ starting from $t=1$ until a prediction is triggered, and necessarily before the deadline $T$. 

%\textcolor{pink!50!orange}{Traditionally, the reward is provided to the learning agent in either one of the following modes. One is to wait for the end of an episode to provide it. This is typically done when learning to play chess, and waiting for the outcome of the play. A second mode is to provide partial rewards during the interaction of the agent with the environment. }
%
%\textcolor{pink!50!orange}{In the context of ECTS, the \textit{first} mode consists in providing the total cost $\mathrm{C}_m(\hat{y}_{\hat{t}}|y) + \mathrm{C}_d(\hat{t})$ (or its opposite, for it to be reward) at the time $\hat{t}$ when the decision is made by the agent. The \textit{second} consists in informing the agent at each time step that waiting entails a delay cost: $\mathrm{C}_d(t) - \mathrm{C}_d(t-1)$ with $\mathrm{C}_d(0) = 0$, and to provide the final (opposite) classification error cost $-\mathrm{C}_m(\hat{y}_{\hat{t}}|y)$ at the time the prediction $\hat{y}_{\hat{t}}$ is made. }

To our knowledge, there are no RL-based ECTS approaches in the literature that are both \textit{separable} and \textit{classifier-agnostic}, which is the scope of this paper. However, this is necessary for a fair comparison with the wide range of state-of-the-art approaches not based on RL. The aim of this paper is to bridge this gap in the literature %\textcolor{magenta}{Phrases ou paragraphe bizarre, ca a été déjà dis, intérêt ? en fait si je comprends bien avant de faire la section 2.2 vous pensez obligatoire de parler de certaines méthodes RL pourquoi ne pas faire une section avec un titre dans ce cas et de le faire vraiment ? là c'est bizarre les 4 paragraphes ci-dessous avant la section 2.2}. \textcolor{red!50!orange}{(AC : je suis d'accord. À partir de ``However, ...'', c'est bizarre et coupe le fil directeur. Je propose la phrase suivante :)}
by presenting a well-founded approach to building such systems. The remainder of this section presents the RL-based approaches in the literature, specifying whether they are separable or not, and which reward function is used. %\textcolor{magenta}{On ne devait pas voir que les séparables?}

%\vspace{7mm}
\paragraph{Review of RL-based ECTS approaches:}

%\vspace{-2mm}
\noindent
\cite{martinez2018deep, martinez2020adaptive} exploit value-based RL in an end-to-end fashion. In any non-separable approaches: (\textit{i}) the RL agent either decides to wait or makes a class prediction and therefore the action space is $\mathcal{A}$ = \{\textit{wait}, $y_0, \ldots, y_K$\} if there are $K$ classes; (\textit{ii}) the state space is defined as the time series themselves. In this approach, the reward function used penalizes the action of waiting and gives equal importance to correct and incorrect predictions. 
%In this approach, the reward function penalizes waiting to make a prediction and gives equal importance to correct and wrong decisions.

EarlyStop-RL \citep{wang2024deep} is a cost-sensitive end-to-end approach, alluding to the possibility of considering any cost function. Indeed, the reward directly corresponds to any misclassification cost and any incremental delay cost, which combine additively.

\textsc{Earliest} and its variants use policy gradient algorithms \citep{hartvigsen2019adaptive, hartvigsen2020recurrent, hartvigsen2022stop} and consist of distinct modules (i.e. an encoder, a triggering agent, and a classifier), each of which operating with its own set of parameters. Thus, these approaches are \textit{separable} even if these modules are jointly trained, but they are \textit{not classifier-agnostic} by design. The reward function used only takes into account the misclassification error, the delay cost being a part of the loss function.       

The SNP approach \citep{huang2022snippet1, huang2022snippet} is a \textit{separable} architecture that consists of (\textit{i}) a specific classification module including an encoder; (\textit{ii}) a RL agent that triggers predictions based on the embedding given by the classification module. This approach is \textit{not classifier-agnostic} by design, and uses a specific reward to favor early and correct decisions with a logarithmic weighting on the trigger time.

\subsection{SOTA: insights from non-RL separable ECTS}
\label{sota_ects}

\textsc{Proba-Threshold} \citep{renault2024early} is a basic method, monitoring the highest conditional probability estimated by the classifier $\max_{y \in {\mathcal{Y}}} p(y|{\mathbf x}_t)$, which is a simple measure of classifier confidence over time. As soon as it exceeds a value, which is a hyperparameter of the method, a prediction is made.

%SR \cite{mori2017early} computes the highest confidence level and the second one for the possible classes in addition to $\frac{t}{T}$. It then checks whether a linear combination of these is positive.

Rather than only considering the largest posterior, \textsc{Stopping Rule} \citep{mori2017early} considers three input variables to trigger decisions $(p_1, p_2, \frac{t}{T})$. The output of the system is defined as:
\begin{equation*}
g(h({\mathbf x}_t))  = \left\{  
    \begin{array}{ll}
        \hat{y} = \displaystyle \argmax_{y \in \mathcal{Y}} p(y|{\mathbf x}_t) & \mbox{if } \gamma_1 \, p_1 + \gamma_2 \, p_2 + \gamma_3 \frac{t}{T} > 0 \\ 
        \emptyset & \mbox{otherwise}
    \end{array} 
    \right.
\label{eq:mori_trigger_function}
\end{equation*}
where $p_1$ is the largest posterior probability  $p(y|{\mathbf x}_t)$ estimated by the classifier $h$, $p_2$ is the margin, i.e., the difference between the two largest posterior probabilities, and $\frac{t}{T}$ represents the proportion of the incoming time series at time $t$. The parameters $\gamma_1$, $\gamma_2$ and $\gamma_3$ are learned from the training set.

%In the ECEC system \cite{lv2019effective}, $h$ computes at time $t$ the history of past classifications $(h(\mathbf{x}_i))_{1\leq i \leq t}$ and transmits it to trigger function. In turn, $g$ implements a criterion that evaluates, in essence, the stability and hence the confidence of the predictions of the classifier, and decides to trigger a prediction when this estimated confidence is above some threshold.

The \textsc{Teaser} method \citep{schafer2020teaser} uses a classifier $h$ that computes the class probabilities for the incoming $\mathbf{x}_t$ and transmits them to the trigger function. In turn, $g$ is a classifier in its own right that classifies the prediction $\hat{y}_t$ as reliable or not. Finally, the trigger function takes the last reliable predictions and their associated times and decides to trigger the prediction $\hat{y}$ only if the same prediction was also given for a number of successive time steps (a hyperparameter to be tuned).

\textsc{Economy} and its variant \citep{dachraoui2015early, tavenard2016cost, achenchabe2021early, zafar2021early} are non-myopic approaches, where the trigger function computes the expected costs, a combination of the misclassification cost and the delay cost, for the current time step $t$ and all future ones. A decision is triggered as soon as the expected cost for the current time step $t$ is the lowest among all future times expected costs. %\textcolor{pink!50!orange}{The most performing variant \textsc{Economy}-$\gamma$ exploits a single variable to trigger predictions encoding confidence levels which are defined by the equal-frequency discretization of the maximum posterior. }

\textsc{Calimera} \citep{bilski2023calimera} is another non-myopic approach that exploits a collection of regressor models, learned in a backward induction fashion as a trigger function. The minimum expected cost occurring in the future time period $[t,T]$ is denoted as $minFutureCost_t$. For a particular time step $t$, the corresponding regressor aims to predict the difference $\Delta = minFuturCost_{t+1} - Cost_t$. Then, \textsc{Calimera} triggers a decision when $\Delta > 0$, i.e. the optimum trigger time is about to be exceeded. The collection of regressors is trained using the following variables: (\textit{i}) all posterior probabilities; (\textit{ii}) the maximum posterior $p_1$; (\textit{iii}) the margin $p_2$.      
 
%Another range of methods is based on Sequential Probability Ratio Test \cite{wald1948optimum}, which, for a given error rate, offers theoretical optimality under i.i.d. assumption between measurements of the time series, as well as an infinite sampling horizon. Recent papers \cite{ebihara2020sequential, ebihara2023toward, ringelearly} try to relax those constraints to make this kind of methods more easily applicable in practice. Those methods fall into the separable realm, where log-likelihood ratios are calculated first and then compared to thresholds.

%Full deep-learning architectures have been recently developed. These can be either separable like the SOCN \cite{lv2023second} algorithm,  the predicted class probabilities sequence from a deep-learning based time series classifier are passed to the triggering function. Then a transformer-based network outputs a confidence scalar, which is itself compared to a threshold in order to trigger the decision. Full deep-learning architectures also can be end-to-end as exemplified with ELECTS \cite{russwurm2023end}, where $h\&g$ outputs both the predicted class value and a probability distribution of triggering the decision. Then, a decision to trigger is sampled as a value using this distribution.

Other separable ECTS approaches include \citep{lv2019effective, lv2023second, schafer2020teaser, ebihara2023toward}. For a more comprehensive description of the ECTS literature, including non-separable approaches, the reader can refer to the survey of
\cite{renault2024early}.

\section{Proposed methodology}
\label{sec:proposed_methodology}

In this section, we present  \textsc{Alert}: a RL-based separable ECTS framework that allows experimenting with various state spaces, reward functions, and classifiers. Its components are described, and possible choices for the state space and the reward functions are discussed.

\subsection{\textsc{Alert}'s pipeline overview}

% \begin{figure*}[!t]
%     \centering
%     \includegraphics[width=1\linewidth]{plots/fig_211.pdf}
%     \caption{\textsc{Alert} training pipeline. \textcircled{1} Example for the state space and the reward function. \textcolor{blue_pastel}{Blue} frame \textcircled{1} denotes that this step is customizable and will vary further in the paper: the values displayed are purely illustrative. \textcircled{2} The generation of the interactions buffer $\beta$. An interaction is defined as a tuple $(s_t, a_t, s_{t+1},r_t)$ containing for a state $s_t$, the associated reward $r_t$ of taking action $a_t$ as well as the next state to which agent transitions $s_{t+1}$. \textcircled{3} The offline training of DDQN.}
%     \label{fig:alert_pipeline}
% \end{figure*}

\textsc{Alert} is a Deep-RL trigger function that can be used in any kind of separable ECTS architecture. %\textsc{Alert} overview is illustrated in Figure \ref{fig:alert_pipeline}. 
 Below, we first outline the main components of \textsc{Alert}:
 %\textcolor{magenta}{du coup on s'attendrait à avoir ensuite 4 sections mais il y en a 5... }: 
\begin{itemize}
    \item \textbf{The state space and reward description}: The first step in designing a RL-based trigger function is to define the state space and what the reward function should be. That is, the inputs of the trigger model and the function to be maximized in order to achieve an effective policy.
    %and what is the function that, once maximized, could lead to a good-performing policy. %\textcolor{magenta}{See section XX for a full ...}
    
    %phrase sur l'encadré bleu ...
    \item \textbf{The experience buffer generation}: Once the setup has been defined, \textsc{Alert} exhaustively generates all possible interactions by simulating ``\textit{wait}'' or ``\textit{trigger}'' actions for every possible time step, for every training sample. 
    \item \textbf{The offline DDQN training}: At this stage, a DDQN model is trained in an offline fashion, i.e. solely based on the buffer generated in the previous step. %\textcolor{magenta}{See section XX for a full ...}
    \item \textbf{The \textit{post hoc} model selection}: Finally, a key component for \textsc{Alert} to work effectively across a various range of different environments is the model selection phase. In this final step, one model is selected from those learned during training to avoid overfitting and improve method stability. %\textcolor{magenta}{See section XX for a full ...}
\end{itemize}

%\subsection{Complete Pipeline}

%In this section, we will go through all the steps needed to learn a Deep RL trigger function for ECTS, using \textsc{Alert}.

\subsection{State space}

In RL, the state, together with the reward signal, provides all the information that the agent uses to decide which next action to take. %Within the context of ECTS, this information is what allows the triggering function $g$ to decide to wait one more time step or to make a prediction $\hat{y}$ about the incoming time series $\mathbf{x}_t$. 
In the context of ECTS, the state space may include any information from the classifiers as well as the time series themselves. \textcolor{black}{In practice, the state representations considered in this work should be viewed as approximately rather than strictly Markovian, since they summarize the information available at time $t$ through a limited set of classifier and time-series-derived features.} It is desirable that the components of the state space be relevant to the classification task to be performed while remaining limited in number. 
%in order to mitigate the \textcolor{gray}{curse of dimensionality} \textcolor{blue}{(compréhensible?)}. 
This compromise is of great importance in the design of effective and efficient approaches \citep{curran2015using}. 

%In Figure \ref{fig:alert_pipeline}, dashed frame \textcircled{1}, a simple example is given, in which only two dimensions are considered: namely, the maximum posterior (in blue) and the predicted label (in orange), both extracted from the classifier. %\textcolor{magenta}{La phrase précédente ne m'est pas compréhensible, je vois pas de "verbe" , sujet... Et où dans la figure 2?}

Throughout this article, several sets of candidate components of the state space vector will be considered, mainly inspired by the most efficient state-of-the-art algorithms \citep{renault2024early}: (\textit{i}) the \textit{predicted class label} \citep{schafer2020teaser}: $\argmax_{k \in \mathcal{Y}} p(y=k|x_t)$; (\textit{ii}) the \textit{maximum posterior} \citep{mori2017early, bilski2023calimera}: $\max_{k \in \mathcal{Y}} p(y = k | x_t)$; (\textit{iii}) the \textit{margin} \citep{mori2017early, schafer2020teaser, bilski2023calimera} which is the difference between the two largest posterior probabilities; (\textit{iv}) the estimation of the \textit{level of confidence} \citep{achenchabe2021early, zafar2021early} in the prediction(s), which can be represented by the bin index within an equal-frequency discretization of maximum posteriors; (\textit{v}) the \textit{current time} $t$ \citep{mori2017early}, which, in the case of finite time series of length $T$, provides the proportion of the time series observed so far; (\textit{vi}) the \textit{time series} itself, sub-sampled or not.
% \begin{itemize}
%     \item The \textit{predicted class label} \cite{schafer2020teaser}: $\argmax_{k \in \mathcal{Y}} p(y=k|x_t)$.
%     \item The \textit{maximum posterior} \cite{mori2017early, bilski2023calimera}: $\max_{k \in \mathcal{Y}} p(y = k | x_t)$.
%     \item The \textit{margin} \cite{mori2017early, schafer2020teaser, bilski2023calimera} which is the difference between the two largest posterior probabilities.
%     \item The estimation of the \textit{level of confidence} \cite{achenchabe2021early, zafar2021early} in the prediction(s), which can be represented by the bin index within an equal-frequency discretization of maximum posteriors.
%     \item The \textit{current time} $t$ \cite{mori2017early}, which, in the case of finite time series of length $T$, provides the proportion of the time series observed so far.
%     \item The \textit{time series} itself, sub-sampled or not.
% \end{itemize}   

Several combinations of the mentioned candidates have been tested as state spaces (see Section \ref{sec:expe_state_space_design}).  All of the components of the state space have been normalized when needed, to keep them all in the $[0,1]$ range, as suggested by \cite{tarasov2024revisiting, fujimoto2021minimalist}. Particularly, the predicted class label has been one-hot encoded and the indexes of confidence levels have been MinMax scaled.

\subsection{Reward function}
\label{sec_reward_fct}

%In RL, candidate policies are evaluated with respect to the reward. The optimal one minimizes the average cost (Eq. \ref{eq_avgcost_weigth}). 

%\textcolor{pink!50!orange}{Here again,} 
There can be multiple choices, organized according to two main options %\textcolor{pink!50!orange}{as outlined in Section \ref{Adapting_RL_to_ECTS}}
: \textbf{(1)} provide the reward at the end of each episode only, or \textbf{(2)} provide rewards at every time step. The rest of this section describes these two types of rewards.

\medskip
\noindent
\textbf{(1)} Reward is provided at the \textit{end of episodes}. The most straightforward approach is to provide a reward directly related to the paid cost at the time the decision is made: 
\begin{equation}
    reward^{trigger} = -\mathrm{C}(\hat{y}_t, y, \hat{t}) = -\mathrm{C}_m(\hat{y}_t|y) - \mathrm{C}_d(\hat{t})
    \label{eq:r1_trigger}
\end{equation}
        
        % \item \textcolor{black}{TO DO: ne laisser que les eq. 1et4} Inspired by imitation learning \cite{hussein2017imitation, kumar2022should}, one could also leverage the optimal cost knowledge\footnote{In the remainder of the paper, ``optimal'' cost $\mathrm{C}^\star$ refers to the best achievable cost given a pre-trained classifier, within the separable ECTS context, and the knowledge of the complete time series $\mathbf{x}^T$.}. It is thus possible to define another reward function such as the gap to this optimal cost: 
        % \begin{equation}
        %     reward^{trigger} = -|\mathrm{C}(\hat{y}_t, y, \hat{t}) -\mathrm{C}^\star| 
        % \text{, ~~with } \mathrm{C}^\star = \mathrm{C}(\hat{y}_t, y, t^\star)    
        % \end{equation}

        %Note here that, even at the optimal decision time $t^\star$, the prediction $\hat{y}_{t^\star}$ may differ from the true class $y$. %\textcolor{magenta}{besoin de le dire ici et pas ailleurs ?} 
        %\ac{(Est-il nécessaire de le dire ? On se demande ce que cela change pour la récompense. À mon avis, cela pose plus de question pour le lecteur dans une section déjà difficile.)}
        
        % \item A third possibility does not use the costs, but only the difference between trigger time $t$ and the optimal one $t^\star$: 
        % \begin{equation}
        %     reward^{trigger} = -|\hat{t} - t^\star|
        % \text{ st. } t^\star = \argmin_{t \in \{1, T\}} \mathrm{C}(\hat{y}_t, y, t)    
        % \end{equation}
        
\noindent
\textbf{(2)} Rewards are provided at \textit{every time step}\footnote{Note that this formulation assumes that misclassification and delay costs can be decomposed additively.}. At each time step, the agent can receive information that waiting entails a delay cost:
\begin{equation}
    reward^{wait} = -\Delta \mathrm{C}_d(t) =  \mathrm{C}_d(t-1) - \mathrm{C}_d(t) \text{, ~~with }\mathrm{C}_d(0) = 0   
    \label{eq:r1_wait}
\end{equation}
        
        % \item Another possibility is to make the agent incur a negative reward $-\Delta C_d(t)$ only if it delays the decision time $\hat{t}$ after the optimal one $t^\star$ and no penalty before. 
    
In both latter cases, the final opposite classification error cost 
$- \mathrm{C}_m(\hat{y}_{\hat{t}}|y)$ is provided at the time the prediction $\hat{y}_{\hat{t}}$ is made. Alternative rewards functions, inspired by imitation learning \citep{hussein2017imitation, kumar2022should}, that depend on optimal trigger time $t^{\star}$ or associated cost $\mathrm{C}^{\star}$, are investigated in Appendix \ref{sec:expe_sensitivity_reward}.

\subsection{Interactions buffer}

% When framed within RL, the ECTS problem has the particularity that actions ($\mathcal{A} = \{wait,\:trigger \}$) do not modify the observed data, i.e. triggered predictions are final and therefore, no more measurements after triggering are observed. It is possible to simulate the outcome of deciding to trigger a prediction for each time step $t \in \{1, T\}$ on each training time series. For each $t$ on a given time series $\mathbf{x}_T$, the costs $\mathrm{C}_m(\hat{y}_t|y)$ and $\mathrm{C}_d(t)$ are indeed available. Having observed that, one can thus build an \textit{exhaustive} set of interactions based on the training time series, before learning any trigger function.

Given a complete training time series $\mathbf{x}_T$, it is possible to compute for each time step $t \in \{1, T\}$ the cost of making a decision: $\mathrm{C}_m(\hat{y}_t|y) + \mathrm{C}_d(t)$. Given $N$ training time series, this can potentially yield $N \times T$  training examples for learning the trigger function. However, as usually done in the literature \citep{mori2017early, schafer2020teaser, bilski2023calimera}, the time series are divided into a smaller, limited number of time step, rather than considering all time steps $t$. This significantly reduces the number of possible transitions and provides a way to standardize the number of time steps considered when datasets contain time series of different lengths. Typically, the number of  time steps considered is set to 20, each one representing $5\%$ of the time series. For each of these 20 time steps, both the ``\textit{wait}'' and the ``\textit{trigger}'' actions are simulated, and the associated rewards are computed, using pre-trained classifier's predictions.
%All of the possible transitions $(s_t, a_t, s_{t+1}, r_t)$ are gathered into a buffer, that remains static for the rest of the training procedure. \ac{(Pas clair)}

%\subsubsection{DQN action selection}
%\subsection{Model training}

\subsection{Offline DDQN}

Finally, a Double Deep-Q-Network (DDQN) model \citep{mnih2015human,van2016deep} is learned in an offline manner to act as our triggering agent. It is based on the classical \textit{Q-learning} \citep{dayan1992q} algorithm. 

\paragraph{Q-Learning:}

The algorithm consists of learning a policy by using a state-action value function, which, for each pair, maps $(s, a)$ the expected gain starting from $s$, taking action $a$ and following $\pi$ afterwards:
\begin{equation}
    \begin{split}
    q_\pi(s_t, a_t) \; &\doteq \; \mathbbm{E}_\pi[G_t \, | \, s_t, a_t] \\ 
    % &= \; \sum_{s_{t+1}, r} p(s_{t+1},r \, | \, s_t,a_t) \, 
    % \biggl[r(s_t, a, s_{t+1})\\ &\hspace{8mm}  \, + \, \gamma \, \sum_{a_{t+1}} \pi(a_{t+1}|s_{t+1}) q_\pi(s_{t+1}, a_{t+1}) \biggr]
    %\\
    % &= \; \sum_{s_{t+1}, r} p(s_{t+1},r \, | \, s_t,a) \, \bigl[r \, + \, \gamma \, v_\pi(s_{t+1}) \bigr] 
    \end{split}
    \label{eq_rl_v_function}
\end{equation}
where $\mathbbm{E}_\pi[\cdot]$ denotes the expected value of a random variable given that the agent follows the policy $\pi$. %and $G_t$ the discounted sum of future rewards. 
%\ac{($G_t$ n'a jamais été défini avant.)}

Optimal policies share the same optimal action-value functions:
\begin{equation}
    \forall (s_t, a_t) \in \mathcal{S} \times \mathcal{A}: ~~ q^\star(s_t, a_t) = \operatornamewithlimits{max}_\pi \, q_\pi(s_t,a_t)
\end{equation}

Thus, provided that an optimal state-action value function has been found, an optimal policy can be derived: $\pi^\star(s) = \argmax_{a} q^\star(s,a)$. Q-learning is a popular way of directly approximating $q^{\star}$ with updates:
\begin{equation}
    Q(s_t, a_t) \xleftarrow{}Q(s_t,a_t) + \delta \bigl[r_t + \gamma \max_a Q(s_{t+1}, a) - Q(s_t, a_t) \bigr]
    \label{eq_Q_learning}
\end{equation}
where $Q$ is the estimated $q$ function, $r_t$, the immediate reward and $\delta$ the learning rate.

%\textcolor{gray}{and \textcolor{red}{$R_{t+1}'$} is the measured return from state $s_t$ when choosing action $a_t$. $R_{t+1}'$ is typically either the immediate reward $R_t$ or a cumulated gain from the current state to the episode's end, if episodes are defined \textcolor{magenta}{($\alpha$?)}}.

When working with high-dimensional and/or continuous state-action spaces, one typically learns a parametrized value function $Q(s, a, \theta)$ \citep{mnih2015human} where $\theta$ should minimize the following: %\textcolor{blue}{(!!notation $r_t$ eq 4 et 5!!)} : 
\begin{equation}\label{eq:loss}
    \mathcal{L}(\theta|s_t, a_t, r_t, s_{t+1}) = \bigl(\overbrace{(r_t + \gamma \max_{a_{t+1}} Q_{\theta}(s_{t+1}, a_{t+1}))}^{Y^{DQN}} - Q_{\theta}(s_t, a_t) \bigr)^2
\end{equation}

%%%% DDQN algorithm

In practice, for stability reasons, the Q-learning target $Y^{DQN}$ in Deep Q-Network (DQN) is computed via a target network, with parameters $\theta^-$, equal to those of the main network, except that its parameters are governed by an exponential moving average of the main network's parameters, with a smoothness parameter $\tau$ \citep{lillicrap2015continuous}. \\
\indent
One well-known issue in DQN is the overestimation bias, which often occurs as the standard DQN update rule is based on a maximum value of a noisy and uncertain estimation of the Q-values, potentially accumulating error as training goes on. \textit{Double DQN} \citep{van2016deep} addresses it by de-correlating action selection from action evaluation. In practice, one can simply use the main online network to select the action, then evaluate it using the target network. The DDQN target thus becomes:
\begin{equation} \label{eq:ddqn}
    Y^{DDQN} = r_t + \gamma Q_{\theta^-}(s_{t+1}, \argmax_a Q_{\theta}(s_{t+1}, a))    
\end{equation}

\paragraph{Offline DDQN :}

\textsc{Alert} adopts an \textit{offline} training scheme \citep{levine2020offline, kostrikov2021offline}. This reduces the computational burden and allows training the triggering agents on many different datasets. In order to further compare \textsc{Alert} with state-of-the-art algorithms, a full batch training procedure has been chosen for all competitors, thus including \textsc{Alert}. Note that extending \textsc{Alert} to offline-to-online or fully online settings remains possible.
In addition, some disadvantages of offline RL, particularly the sensitivity to distribution shifts, appear somewhat mitigated when approaching ECTS as done in this paper. Indeed, as mentioned earlier, the static buffer is exhaustive and balanced in terms of action representation. Thus, out-of-distribution actions are not a matter here, reducing the risk of extrapolation error. 

Transitions are sampled uniformly, as done by \cite{cetin2024simple}, 
%\textcolor{blue}{Note that, by sampling transitions uniformly, we did not preserve temporal consistency. Furthermore, neither \textit{valuable} transitions are prioritized nor \textit{useless} ones are filtered out. Indeed, we found that neither strategy led to improvements compared to uniform sampling.} 
from the buffer $\beta$ and transmitted to the Q-estimator neural net in order to minimize the loss in Equation \ref{eq:loss}. We used Layer Normalization \citep{lei2016layer} in the Q-network since it further mitigates overestimation bias by bounding the output Q-values \citep{ball2023efficient, tarasov2024revisiting}. 
%\ac{(C'est assez bizarre de reparler de l'overestimation bias ici, alors qu'on laissait entendre que le problème était résolu quelques paragraphes avant.)}
Once trained, the Q-Network is used as our triggering module: based on an observed state $s_t$, action $a_t$ is selected such that its associated estimated Q-value is maximal: $a_t = \arg\max_{a\in\mathcal{A}} Q(s_t, a)$. In the case where the policy selects the ``\textit{wait}'' action after reaching the deadline $T$, the prediction is triggered regardless, and the associated reward is provided to the agent. 

\subsection{Model selection}\label{model_selection}

Regularization is ensured by a model selection strategy that limits the number of epochs and thus effectively combats overfitting. First, several policies are trained over different train/validation splits, as this greatly improves offline off-policy evaluation \citep{nie2022data}. Then, for each split, the policy under training is evaluated over the validation set at a given epoch frequency in an \textit{online} fashion, i.e. with iterative measurements over the time series, like at testing time. The \textit{AvgCost} is used as a metric at each validation stage, and the corresponding model is saved. Finally, after training, the epoch index for which the validation metric is lowest on average across all splits is selected. Among the models trained for this number of epochs, the best performing one across all splits is selected to be the final model.

\section{Evaluation protocol}
\label{sec_exp_protocol}

%\textcolor{red}{TODO : intro de section en renvoyant vers l'Annexe pour les parties retirées, nommer les compétiteurs ? .}

In the experiments conducted, the following state-of-the-art algorithms were tested: \textsc{Proba-Threshold}, \textsc{Stopping Rule}, \textsc{Economy}, \textsc{Calimera}, and \textsc{Earliest}. More details about the competitors and the experimental protocol, including datasets and implementation specifications, are reported in Appendix \ref{sec:protocol_app}.
%\paragraph{Cost setting and evaluation metrics}
%\label{sec_eval_metrics}
%\textcolor{black}{Alternative cost functions are evaluated in Appendix \ref{linear_appendix}.} 
Evaluation is conducted using the \textit{AvgCost} metric.
% \begin{align}
%     AvgCost_{\alpha} = \frac{1}{N} \sum_{i=1}^{N} \alpha \times C_m(\hat{y}_i|y_i) + (1 - \alpha) \times C_d(\hat{t}_i)
%     \label{eq_avgcost_weigth}
% \end{align}
\begin{align}
    AvgCost = \frac{1}{N} \sum_{i=1}^{N} C_m(\hat{y}_i|y_i) + C_d(\hat{t}_i)
    \label{eq_avgcost_weigth}
\end{align}

To meet the requirements of potential applications such as anomaly detection or decision-making in hospital emergency departments, we have chosen to use unbalanced misclassification costs and exponential delay costs, which are more consistent with real-world conditions, as done by \cite{renault2024early}. 
%In addition of being more realistic, this setting is also more demanding for (separable) ECTS systems since the classifiers on which they are based are usually not learned using these costs. 
%For our experiments, we mainly used the definition of the costs described:
\textcolor{black}{Furthermore, in order to assess how the methods adapt to various balances $\alpha$ between the misclassification and delay costs, the training and evaluation (including model selection and testing) are conducted using weighted cost functions, with values of $\alpha$ varying from $0$ to $1$, with a $0.1$ step: }
\begin{align}
    &C_d(t) = \textcolor{black}{(1-\alpha)} \times \exp( \frac{t}{T} \times \log100) \\
    &C_m(\hat{y}|y) = \left\{ 
    \begin{array}{ll}
          \textcolor{black}{\alpha} \times 100 \times \mathbbm{1}(\hat{y} \neq y) & \mbox{if $y \in$  minority class}   \\
         \textcolor{black}{\alpha} \times \mathbbm{1}(\hat{y} \neq y) & \mbox{otherwise}
    \end{array}
    \right.
\end{align}

\textcolor{black}{
The more standard \textit{linear} and \textit{balanced} cost functions are also tested further in the experiments: 
\begin{align}
    C_d(t) &= (1-\alpha) \times \frac{t}{T} \\
    C_m(\hat{y}|y) &= \alpha \times \mathbbm{1}(\hat{y}\neq y) 
\end{align}
}

\section{Experimental results}
\label{sec_exp_results}

The \textsc{Alert} framework presented in Section \ref{sec:proposed_methodology} is used in our experiments to learn various trigger functions that exploit different state spaces and reward functions. The experiments reported in this section aim to shed light on the two questions:

\begin{enumerate}
   \item First, \textit{given the same input information, how do we compare the performance of handcrafted trigger functions and their RL counterparts}? Section \ref{same_state_space} examines this question and the impact of state space design. Using this comparison, new state spaces can be tested and their performance measured. Section \ref{design_state} considers different trigger functions learned from several variations of the state space and concludes by highlighting a new RL-based system, called \textsc{Alert}$^+$.

   \item \textit{How does \textsc{Alert}$^+$ fare against state-of-the-art approaches}? (see Section \ref{sec:expe_SOTA_comparison})

   %\item \textcolor{gray}{What is the impact of the reward function? (see Section \ref{sec:expe_senitivity_reward})}%Does it change the qualitative findings about the respective values of the approaches?  (see Section \ref{sec:expe_senitivity_reward})
\end{enumerate}

In this section, the used reward function is Equation \ref{eq:r1_wait}, i.e. distilling temporal cost at each time step.
%alternative rewards are tested in Appendix \ref{sec:expe_sensitivity_reward}. %\textcolor{blue}{since it is among the best ones for learning trigger functions using RL} (see Section \ref{sec:expe_senitivity_reward}).

% \newcolumntype{R}[2]{%
%     >{\adjustbox{angle=#1,lap=\width-(#2)}\bgroup}%
%     l%
%     <{\egroup}%
% }
%\newcommand*\rot{\multicolumn{1}{R{45}{1em}}}% no optional argument here, please!
%\newcommand{\rot}[1]{\rotatebox{45}{\parbox{1cm}{\centering #1}}}
%\newcommand{\rot}[1]{\multicolumn{1}{c}{\rotatebox{45}{#1}}}
\newcommand{\rot}[1]{%
  \multicolumn{1}{l}{\rotatebox[origin=l]{45}{\makebox[0pt][l]{#1}}}%
}

\begin{table}[!htb]
\caption{State spaces of \textsc{Alert}'s variants.}
\centering
\vspace{2.3cm}
\resizebox{0.6\textwidth}{!}{
\begin{tabular}{rccccccc|l}
 & \rot{Max. posterior} & \rot{All posteriors} & \rot{Margin} & \rot{Pred.class} &  \rot{Lvl of conf. \scriptsize(num. bins = 10)} & \rot{Time} & \rot{Time Series \scriptsize(num. of points)} & \rot{\textbf{Total Dim $|\mathcal{S}|$}}
    \\ \hline
\multicolumn{1}{c|}{\textsc{Alert Pt}} & \cmark &&&&&&& 1 \\ 
\multicolumn{1}{c|}{\textsc{Alert Eco}} &&&&& \cmark &&& 1  \\
\multicolumn{1}{c|}{\textsc{Alert Sr}} & \cmark && \cmark &&& \cmark && 3 \\ 
\multicolumn{1}{c|}{\textsc{Alert Post.}} && \cmark &&&&&& $|\mathcal{Y}|$ \\
\multicolumn{1}{c|}{\textsc{Alert Cal}} & \cmark & \cmark & \cmark &&&&& $|\mathcal{Y}|+2$ \\
\multicolumn{1}{c|}{\textcolor{red}{\textsc{Alert}$^+$}} & \cmark & & \cmark & \cmark & \cmark & \cmark && $|\mathcal{Y}|+4$ \\
\hline
%\multicolumn{1}{c|}{\textcolor{red}{$\hookrightarrow$} \textsc{Random}} & \cmark & & \cmark & \cmark & \cmark & \cmark && $|\mathcal{Y}|+24$ \\
\multicolumn{1}{c|}{\textcolor{red}{$\hookrightarrow$} \textsc{Sub Series}\footnotemark[5]} & \cmark & & \cmark & \cmark & \cmark & \cmark & \cmark$ {(20)}$ & $|\mathcal{Y}|+24$ \\
\multicolumn{1}{c|}{\textcolor{red}{$\hookrightarrow$} \textsc{Full Series}} & \cmark & & \cmark & \cmark & \cmark & \cmark & \cmark$ {(T)}$ & $|\mathcal{Y}|+4+T$ \\
\hline
\end{tabular}%
}
\label{table_alert}
\end{table}

\subsection{State space design}
\label{sec:expe_state_space_design}

\footnotetext[5]{~Time series are averaged over windows, representing $5\%$ of the series, so that the resulting number of points is equal to 20.}

\subsubsection{Same state space: SOTA vs. \textsc{Alert}}
\label{same_state_space}

Here, we compare the performance of the handcrafted trigger functions of four representative algorithms: \textsc{Economy}, \textsc{Stopping Rule}, \textsc{Calimera} and \textsc{Proba-Threshold} for ECTS, and the corresponding RL-based trigger functions: \textsc{Alert Eco}, \textsc{Alert Sr}, \textsc{Alert Cal} and \textsc{Alert Pt} , which use the same input information (see Table \ref{table_alert}). %\textcolor{blue}{(RL counterpart pas vraiment expliqué + vocabulaire pas homogène)}
%Namely, (\textit{i}) $\mathcal{S} = \{ \textit{level of confidence} \}$ for \textsc{Economy}, (\textit{ii}) $\mathcal{S} = \{\textit{max posterior}, \textit{margin}, \textit{time} \}$ for \textsc{Stoppig Rule}, and (\textit{iii}) $\mathcal{S} = \{\textit{all posteriors}, \textit{max posterior}, \textit{margin} \}$ for \textsc{Calimera}. 

\begin{figure}[!htb]
    \centering
    \subfloat[\textsc{Economy}: $\mathcal{S} = \{ \textit{level of confidence} \}$]{\includegraphics[width=0.5\linewidth]{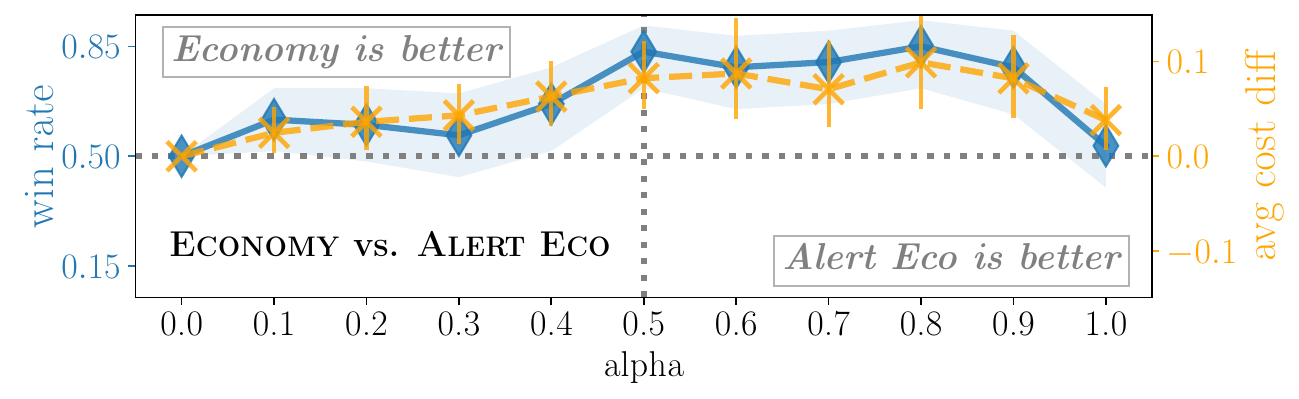}\label{pair_eco}}
    \subfloat[\textsc{Stopping Rule}: $\mathcal{S} = \{\textit{max posterior}, \textit{margin}, \textit{time} \}$]{\includegraphics[width=0.5\linewidth]{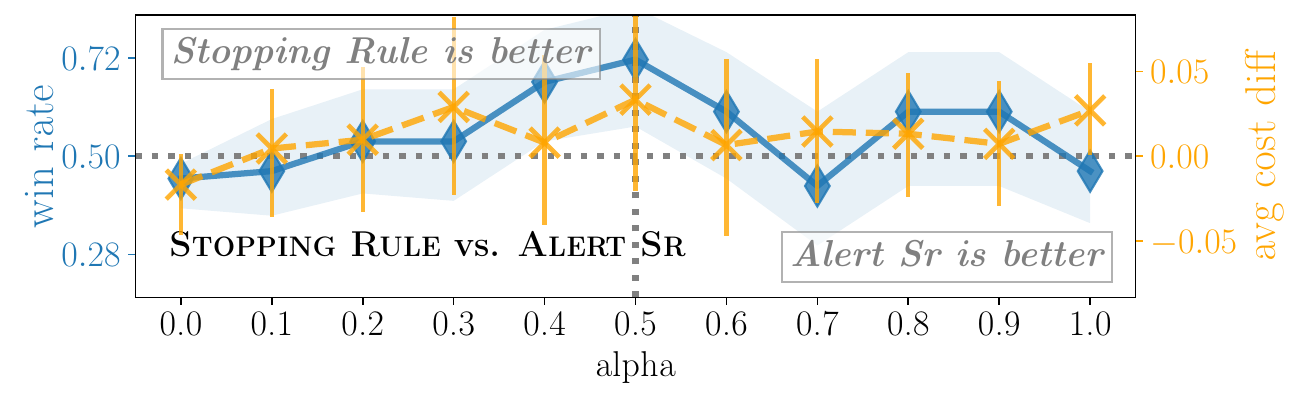}\label{pair_sr}} \\
    \subfloat[\textsc{Calimera}: $\mathcal{S} = \{\textit{all posteriors}, \textit{max posterior}, \textit{margin} \}$]{\includegraphics[width=0.5\linewidth]{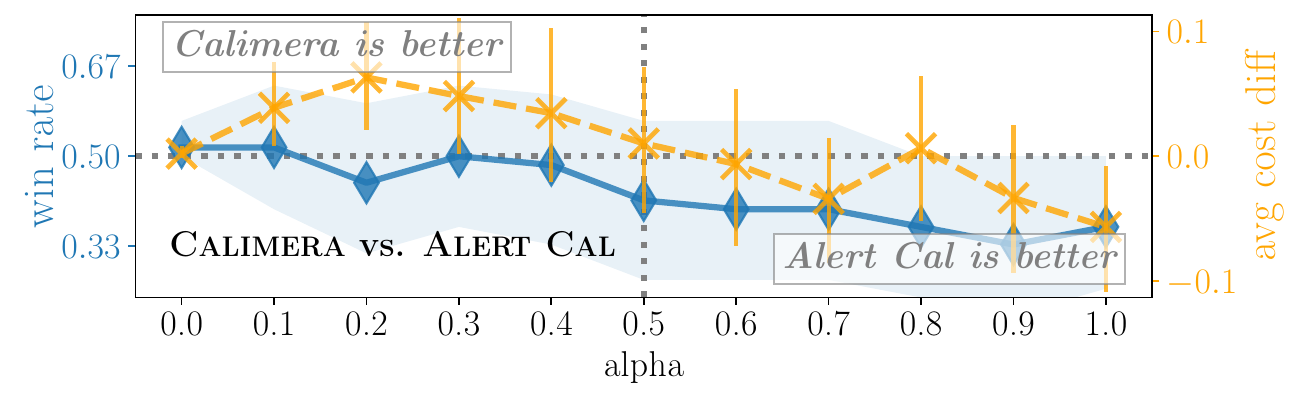}\label{pair_cal}}
    \subfloat[\textsc{Proba-Threshold}: $\mathcal{S} = \{\textit{max posterior} \}$]{\includegraphics[width=0.5\linewidth]{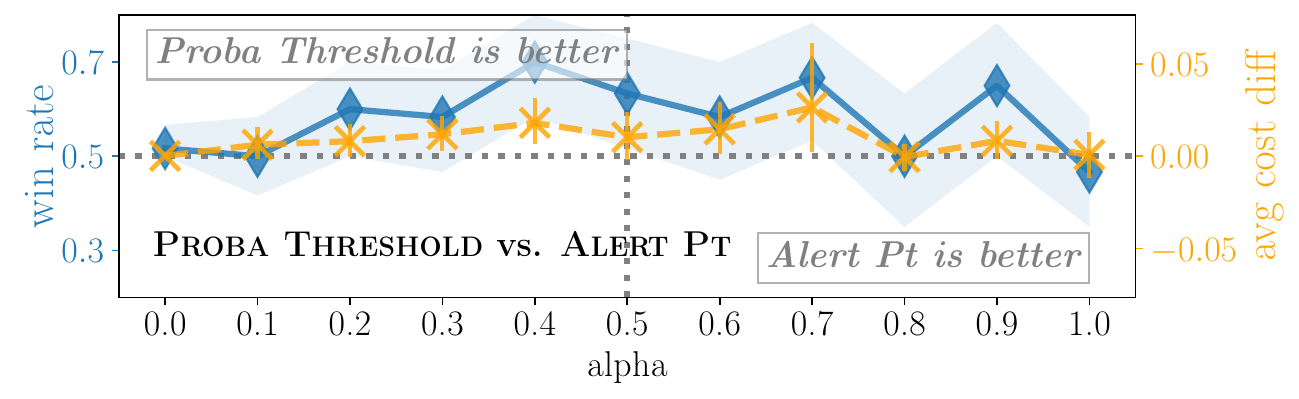}\label{pair_pt}}
    \caption{Pairwise comparison of SOTA methods versus their RL counterpart using same information as input. The \textcolor{blue_mpl}{black} curve, ranging from 0 to 1, represents the win rate of the original method. The \textcolor{orange_mpl}{orange} curve, represents the difference of \textit{AvgCost} between SOTA and RL counterparts, normalized by $\textit{AvgCost}^{\star}$, \textcolor{black}{computed as $\frac{AvgCost_{SOTA}-AvgCost_{RL}}{AvgCost^{\star}}$}. In both cases, points above the horizontal line indicate that the original method is better than its RL-based counterpart. Shaded areas/vertical bars correspond to 90\%  bootstrap confidence intervals.
    %\textcolor{orange}{(est-ce qu'il a été dit quelque part ce qu'est une ``base'' method ?)
    }
    \label{fig:pairwise}
\end{figure}

Figure \ref{fig:pairwise} provides an overall qualitative view of the performance for each handcrafted trigger function vs. RL-based one for values of $\alpha \in [0, 1]$. 
%Several trends are apparent. \ac{(Et one ne parle que d'une tendance !? C'est déstabilisant. Si une seule, ne pas évoquer l'existence de plusieurs.)} 
\textcolor{black}{The overall trend is that \textbf{as the state space gets enriched, RL-based algorithms tend to match their handcrafted counterparts} thus suggesting that RL is able to extract better trigger functions when given enough information. That being said, with restricted input, RL-based algorithms tend to be outperformed.} In effect, \textsc{Economy's} trigger function extracts from $\mathcal{S} = \{ \textit{level of confidence} \}$ what turns out to be useful for decision-making.
%\ac{(toute l'information ! qu'est-ce que ça veut dire ?)}. 
For instance, in Figure \ref{pair_eco}, when $\alpha = 0.8$, \textsc{Economy} wins over almost $85\%$ of the datasets, and is $10\%$ closer to the $AvgCost^{\star}$. This is clearly less the case when operating over larger state space $\mathcal{S} = \{\textit{all posteriors}, \textit{max posterior}, \textit{margin} \}$.

One related question naturally arises: is this enough to enlarge the state space to obtain better RL-based trigger functions? 
%Second, does the choice of the input features also play a role? 
A set of experiments using \textsc{Alert} has been designed to answer these questions.

%\textcolor{magenta}{dans la légende en gris en bas à droite mettre ALert XXX à chaque fois}

\subsubsection{Exploring the design of state spaces for RL-based trigger functions:}  
\label{design_state}

One advantage of \textsc{Alert} is that adding features to the state space has little impact on computation time and implementation. It is thus easy to try out several variations of the state space used. Specifically, we compare the performance obtained with \textsc{Alert} using the state spaces displayed in Table \ref{table_alert}, introducing \textsc{Alert}$^+$ which relies on the union of state spaces of the best SOTA approaches, namely: \textsc{Calimera}, \textsc{Economy}, \textsc{Stopping Rule} and \textsc{Teaser}.%\textcolor{blue}{(déplacer Table 1?)}

%\ac{\begin{itemize}
%   \item \textsc{Alert} \textsc{Pt} (corresponding to \textsc{Probability Threshold}) with state space equal to the maximum posterior probability: $\operatornamewithlimits{max}_{y \in \mathcal{Y}} Pr(y)$ at time $t$, of size 1.
%   \item \textsc{Alert} \textsc{Eco} (\textsc{Economy}) with $\mathcal{S} = \{ \textit{level of confidence} \}$ of size $|\mathcal{S}| = 1$
%   \item \textsc{Alert} \textsc{Sr} (\textsc{Stopping Rule} with $\mathcal{S} = \{\textit{max posterior}, \textit{margin}, \\
%    \textit{time} \}$ of size $|\mathcal{S}| = 3$.
%   \item \textsc{Alert} \textsc{Posteriors} where the state space is the vector of probabilities associated with all classes, thus of size $|\mathcal{Y}|$
%   \item \textsc{Alert} \textsc{Cal} (\textsc{Calimera}) with $\mathcal{S} = \{\textit{all posteriors}, \textit{max posterior}, \\
%   \textit{margin} \}$ of size $|\mathcal{S}| = |\mathcal{Y}| + 2 $.
%   \item \textsc{Alert}$^+$, where the state space is the union of the state spaces of \textsc{Economy}, \textsc{Stopping Rule}, \textsc{Calimera} and {\textsc{Teaser}}:\\  $\{\textit{max posterior},\: \textit{margin},$ $\textit{pred. class}$,\: $\textit{level of confidence},\: \textit{time} \}$, of size $|\mathcal{Y}| + 4$ since \textit{pred. class} is one-hot encoded.
%\end{itemize}}

\begin{figure}[!htb]
    \centering
    \includegraphics[width=0.65\linewidth]{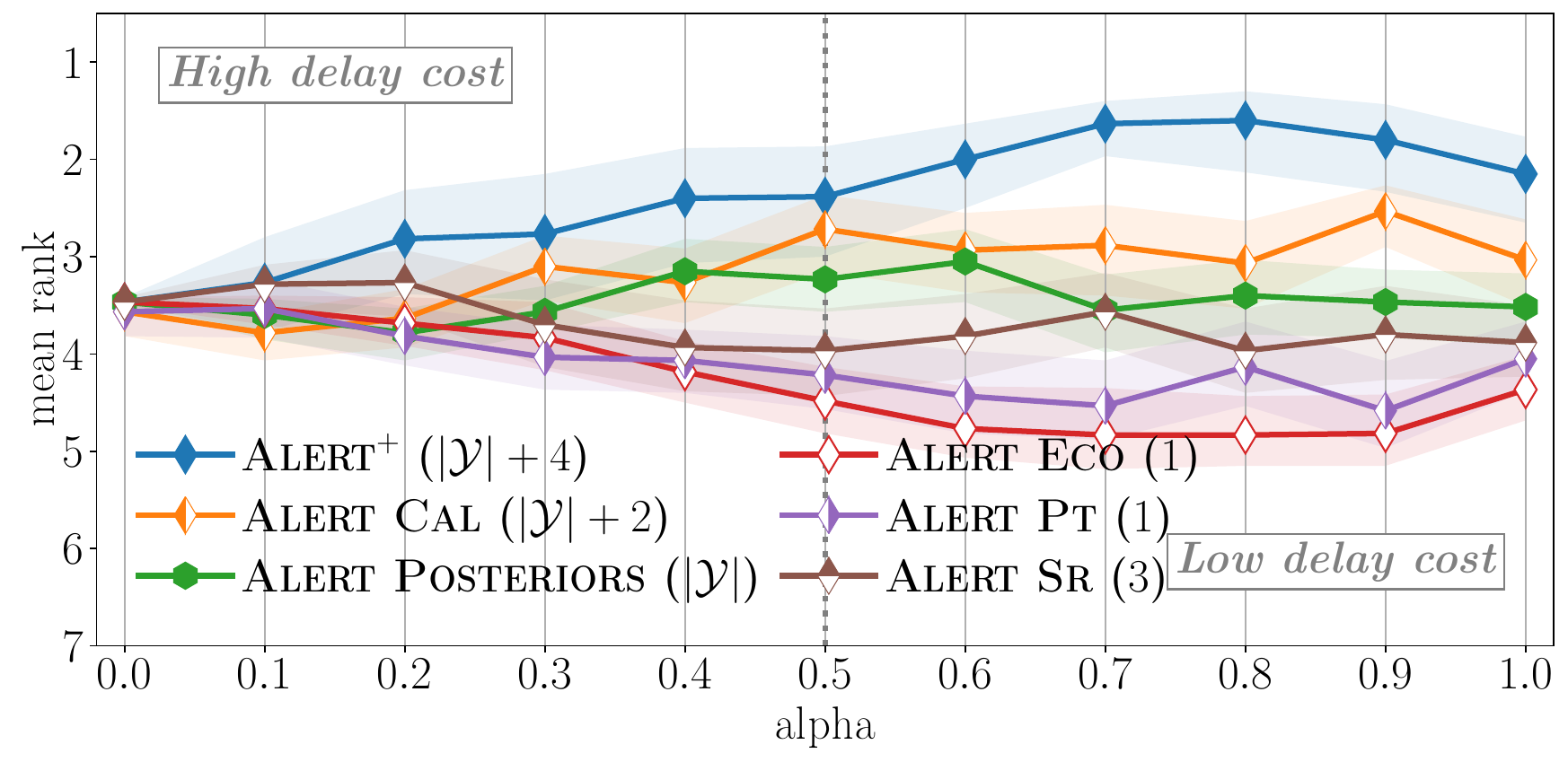}
    \caption{Evolution of the mean ranks, for every $\alpha$. %based on the \textit{AvgCost}. 
    \textsc{Alert$^{+}$} is combining SOTA state spaces.}
    \label{fig:bump_state1}
\end{figure}

Figure \ref{fig:bump_state1} reports the ranking of the resulting ECTS performance on the datasets while testing the versions of \textsc{Alert} in the upper part of Table \ref{table_alert}. A clear trend emerges, with a marked advantage when \textsc{Alert} uses larger state spaces. In addition, it is interesting to see that \textsc{Alert} \textsc{Posteriors} is outranked both by \textsc{Alert}$^+$ and by \textsc{Alert} \textsc{Cal}, while it is possible to derive the inputs of these two algorithms using the inputs of the former, emphasizing the impact of processing input information. \textcolor{black}{Additional results about each tested features' importance can be found in Appendix \ref{state_ablation}.}  

Instead of considering only the features used in handcrafted trigger functions from the literature, \textit{could it be that by further increasing the state space, better performance can still be reached?} To answer that question, we tried including even more information related to the time series in the state space and compared them with \textsc{Alert}$^+$ (lower part of Table \ref{table_alert}). 
This experiment shows that even though \textsc{Alert}$^+$ has the most limited state space in size in this comparison, it tops the versions that somehow include time series information (see Appendix \ref{app_state_space} for more details).

\subsection{\textsc{Alert}$^+$ vs. State-of-the-art}
\label{sec:expe_SOTA_comparison}

The previous section concluded that \textsc{Alert}$^+$ seems to strike a good balance between too small a state space and too a large one with no gain in useful information. But how does it fare against the state-of-the-art competitors described in Section \ref{sec_competing_approaches}? 

\begin{figure}[!htb]
    \centering
    \subfloat[Comparison of the mean ranks, for every $\alpha$, based on the \textit{AvgCost} metric. \centering Shaded areas correspond to 90\%  bootstrap confidence intervals.]{\includegraphics[width=0.65\linewidth]{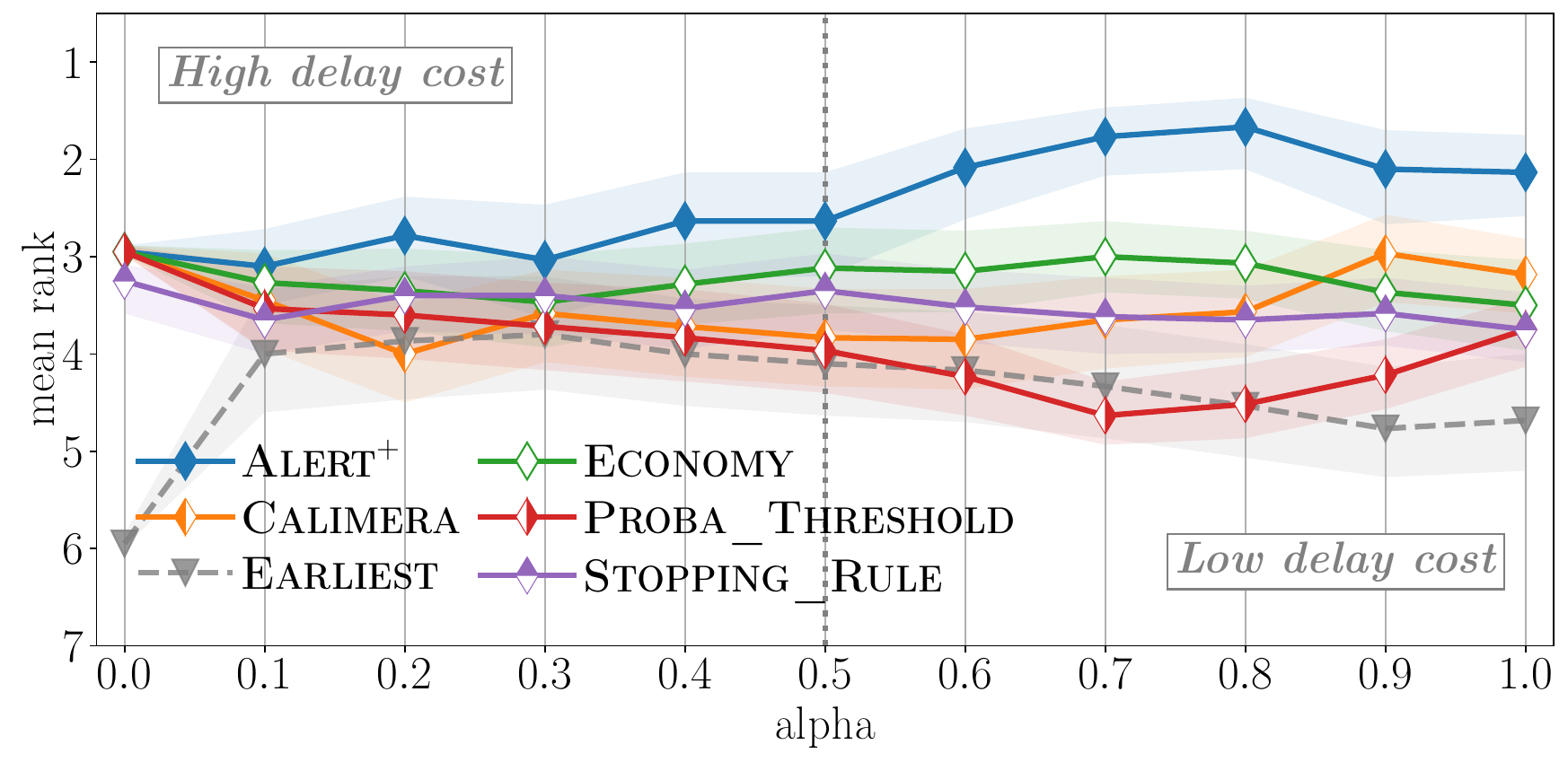}\label{bump_sota}} \\
    \subfloat[Wilcoxon signed-rank test labeled with mean \textit{AvgCost}, $\alpha = 0.8$.]{\includegraphics[width=0.7\linewidth]{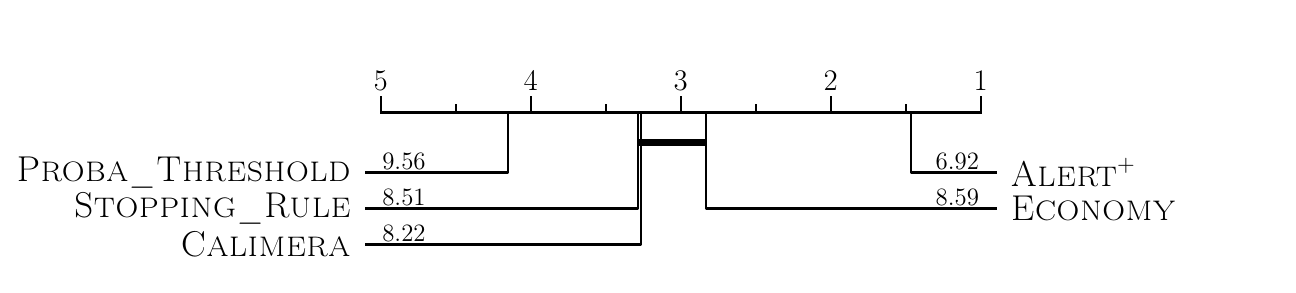}\label{cd_sota}}
    \caption{The ranking plot \protect\subref{bump_sota} shows that, across all $\alpha$, \textsc{Alert$^{+}$} dominates all competitors. This result is significant as supported by statistical tests as shown in plot \protect\subref{cd_sota} for $\alpha = 0.8$.}
    \label{fig:ranks_sota}
\end{figure}

\paragraph{Comparison in terms of ranks and \textit{AvgCost}:}
The results of the experiments carried out are summarized in Figure \ref{fig:ranks_sota}. \textcolor{black}{First of all, the new system \textsc{Alert}$^+$, having the richest state space, outranks all competitors for most values of $\alpha$ on average across the datasets.} Second, looking at \textit{AvgCost}, what is paid actually when using the algorithms, \textsc{Alert}$^+$ significantly outperforms its competitors in the low delay cost regime; for example, when $\alpha = 0.8$ as shown with the Wilcoxon signed-rank-test.
%(and for other values of $\alpha$ as well \textcolor{magenta}{affirmation gratuite où est la preuve dans l'article?}), 
For high delay costs, such as exponential ones, the superiority is less pronounced, as the optimal policy is to decide early on, which all algorithms tend to do. 

\begin{figure}[!htb]
    \centering
    \includegraphics[width=0.5\linewidth]{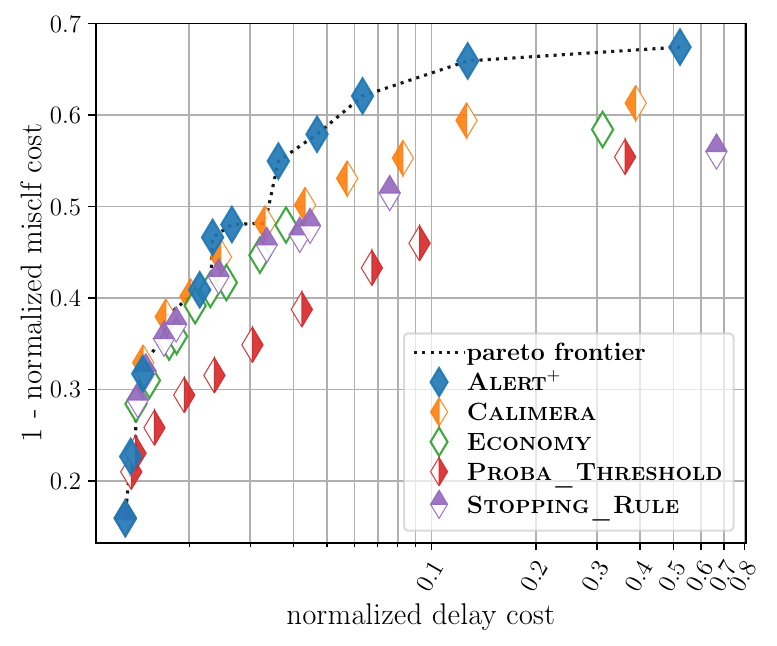}
    \caption{Pareto front, displaying for each $\alpha$, the normalized version of the \textit{AvgCost}, decomposed over delay and one minus misclassification cost on $x$-axis and $y$-axis respectively. Best approaches are located on the top left corner. High $\alpha$ values are located on the right, low ones on the left. Due to the exponential shape of the delay cost, the $x$-axis is on log scale. \textcolor{black}{Both axis are normalized based on the worst cost value possible, i.e. $C_d(T)$ and $\sum_{i=1}^N \max_{\hat{y}_i} C_m(\hat{y}_i|y_i)$ for the $x$-axis and $y$-axis respectively.}}
    \label{fig:pareto_sota}
\end{figure}

\paragraph{Comparison using Pareto front:}
Considering the minimization of misclassification and delay costs as two conflicting objectives, one can draw the Pareto front for each method: the set of points for which no other point dominates with respect to both objectives. 
Figure \ref{fig:pareto_sota} shows the result for $\alpha \in \{0, 0.1, 0.2, \ldots, 1.0\}$. What stands out first is the clear domination of \textsc{Alert}$^+$. A closer examination reveals that \textsc{Alert}$^+$ generally makes its decision later than its competitors, risking higher delay costs, but this extra cost is generally more than offset by lower misclassification costs. 

%This is confirmed in Figure \ref{fig:scatter_rmse} with the marginals over the trigger moments. 

% \bigskip
% \textcolor{gray}{Figure \ref{fig:pareto_sota} offers another view on the experiments results, \ab{by considering the minimization of misclassification and delay costs as two conflicting objectives. The \textit{Pareto front} is the set of points for which no other point dominates with respect to both objectives. It is drawn here when varying $\alpha \in \{0, 0.1, 0.2, \ldots, 1.0\}$.} 
% %showing the Pareto front of the two types of costs, i.e. misclassification and delay, while vary $\alpha$. 
% Once again, \textsc{Alert$^{\star}$} clearly dominates other approaches, 
% %. It is noticeable that \textsc{Alert$^{\star}$} 
% \ab{and} usually pays a higher delay cost than competitors for %similar 
% \ab{a fixed} $\alpha$. Yet, it outperforms state-of-the-art methods by reducing the \ab{misclassification} cost. }
% %paid due to errors in predictions. %The resulting Pareto front clearly demonstrates the effectiveness of \textsc{Alert$^{\star}$}.

\begin{figure*}[!htb]
    \centering
    % \subfloat[\textsc{Alert$^{+}$}]{\includegraphics[width=0.45\linewidth]{plots/qualitative/scatter_rmse/tex/scatter_rmse_alert.pdf}\label{fig:scatter_alert}} 
    % \subfloat[\textsc{Economy}]{\includegraphics[width=0.45\linewidth]{plots/qualitative/scatter_rmse/tex/scatter_rmse_economy_innerplot.pdf}}\label{fig:scatter_eco} \\
    % \subfloat[\textsc{Stopping Rule}]{\includegraphics[width=0.45\linewidth]{plots/qualitative/scatter_rmse/tex/scatter_rmse_stopping_rule.pdf}}\label{fig:scatter_sr}
    % \subfloat[\textsc{Calimera}]{\includegraphics[width=0.45\linewidth]{plots/qualitative/scatter_rmse/tex/scatter_rmse_calimera_innerplot.pdf}}\label{fig:scatter_cal}
    \subfloat[\textsc{Alert$^{+}$}]{\includegraphics[width=0.33\linewidth]{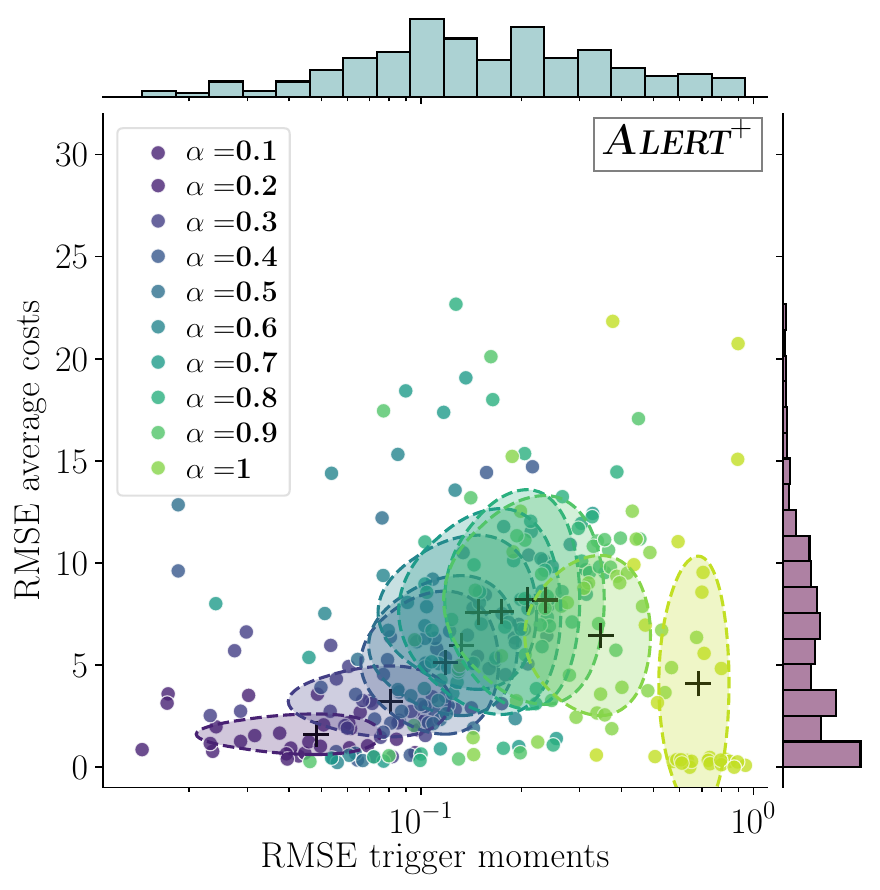}\label{fig:scatter_alert}}
    \subfloat[\textsc{Economy}]{\includegraphics[width=0.33\linewidth]{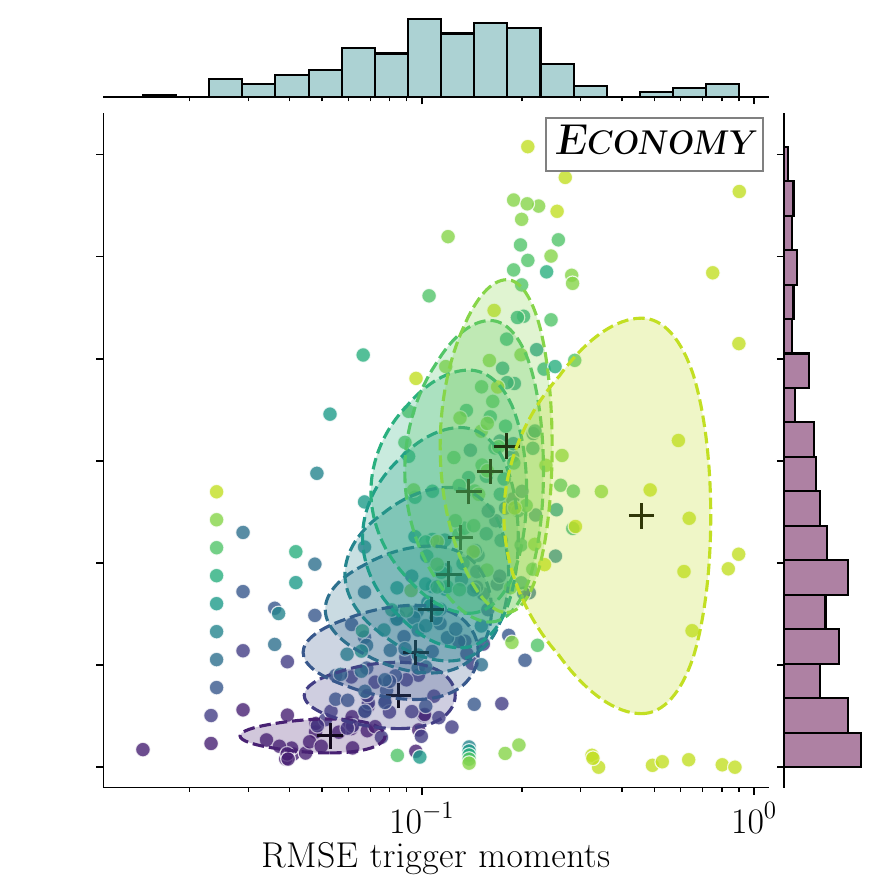}\label{fig:scatter_eco}}
    \subfloat[\textsc{Calimera}]{\includegraphics[width=0.33\linewidth]{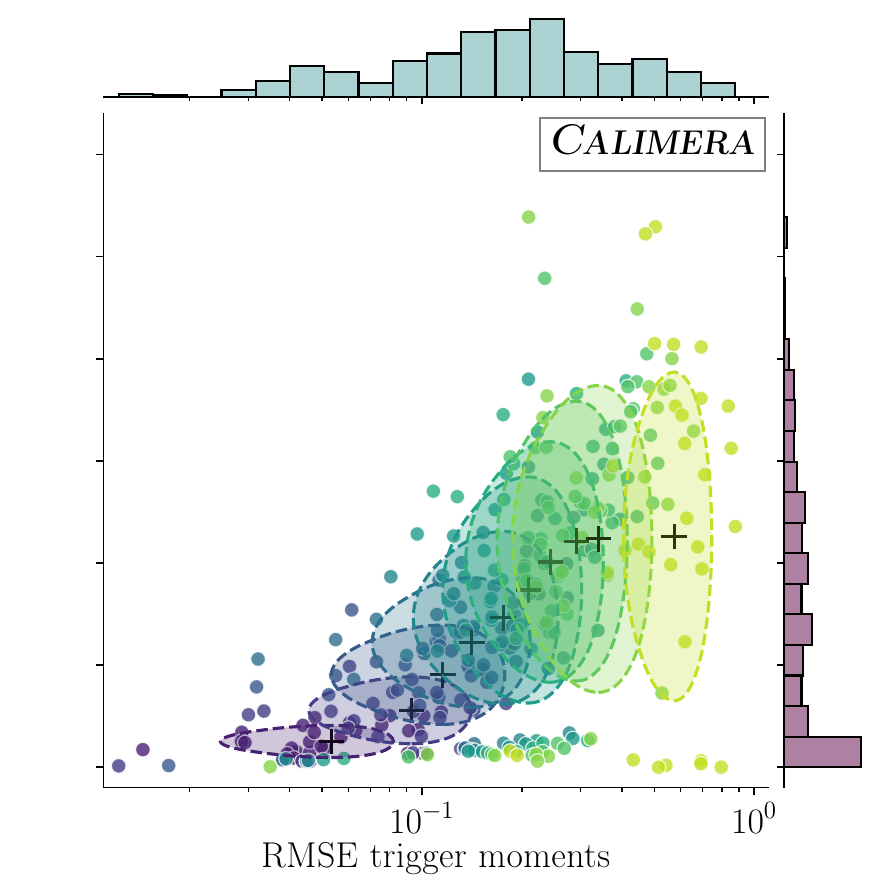}}\label{fig:scatter_cal}
    \caption{The $x$-axis reports how far is the normalized triggering time from the best \textit{a posteriori} one, on log scale: left is better. The $y$-axis reports the difference between the \textit{AvgCost} incurred by the algorithm compared to the best \textit{a posteriori} one, \textit{AvgCost}$^{\star}$: lower is better. The black crosses report the average performance for each value of $\alpha$ (greater values correspond to lower relative importance of the delay cost). Ellipses display $2 \times$ the standard deviation over both axis, computed for each $\alpha$ value.}
    %Scatter plot representing the RMSE computed over trigger moments, normalized by $T$, \textcolor{orange}{(??)} in $x$-axis on log scale and RMSE computed over \textit{AvgCost} in $y$-axis. One point represents a dataset with a corresponding $\alpha$ value. Best reachable policies are located on the bottom left corner. Ellipses display $2 \times$ the standard deviation over both axis, computed for each $\alpha$ values.} 
    %The marginals are plotted using $\sqrt{n\_datasets \times n\_alpha} \approx 18$ bins. \textcolor{orange}{(Je ne comprends pas.)}}
    \label{fig:scatter_rmse}
\end{figure*}
%\clearpage

\paragraph{Detailed 
%\textcolor{red}{(Qualitative?) (AC: utiliser ``qualitative'' est dangereux. Ça fait trop \textit{soft}. On a quand même des mesures précises et détaillant plusieurs aspects de ces algorithmes.)} 
comparison of \textsc{Alert}$^+$, \textsc{Economy} and \textsc{Calimera}:}
Figure \ref{fig:scatter_rmse} conveys a wealth of information \textcolor{black}{(the rest of the competing approaches can be found in Appendix \ref{app_scatter}).} The $x$-axis reports how far the normalized trigger time is from the best \textit{a posteriori} one%\textcolor{red}{(given the pre-trained classifier?)}
, on log scale, the smaller the better.  The $y$-axis reports the difference between the \textit{AvgCost} incurred by the algorithm compared to the best \textit{a posteriori} one, \textit{AvgCost}$^{\star}$, again lower is better. The colored dots correspond to experimental results for the datasets and for values of $\alpha \in \{0, \ldots, 1\}$. The marginals for the $x$-axis and the $y$-axis are reported resp. on the top and on the right of the graph.

It is readily apparent that \textsc{Alert}$^+$ tends to yield a lower average cost, especially for intermediate values of $\alpha$, confirming the findings reported above. \textsc{Alert}$^{+}$ is also more robust with respect to the variety of datasets, displaying smaller ellipses and thus less variance than its competitors. Finally, \textsc{Alert}$^{+}$ spreads its decision times more than its competitors. For low values of $\alpha$ (high delay costs), the ellipses of \textsc{Alert}$^+$ are more extended along the $x$-axis, with some values of RMSE trigger moments being half of those of \textsc{Economy} and \textsc{Calimera}. For high values of $\alpha$ (low delay costs), \textsc{Alert}$^+$ tends to spread its decision time (high RMSE trigger moments) even more than \textsc{Economy} and significantly more than \textsc{Calimera}. It can indeed afford it in this regime. Overall, \textsc{Alert}$^+$ seems to better handle the decision time in favor of improved performance in terms of \textit{AvgCost}. \textcolor{black}{Toward the better understanding of the \textsc{Alert}$^+$ good performances, qualitative results are examined in Appendix \ref{qualitative}.} %Additional results, included runtime analysis (Appendix \ref{runtime}), sensitivity studies (Appendix \ref{sensitive}), qualitative results (Appendix \ref{qualitative}) are made available in the Appendix.

\subsection{Linear/Balanced cost setting}
\label{linear_appendix}

\begin{figure}[H]
    \centering
    \includegraphics[width=0.65\linewidth]{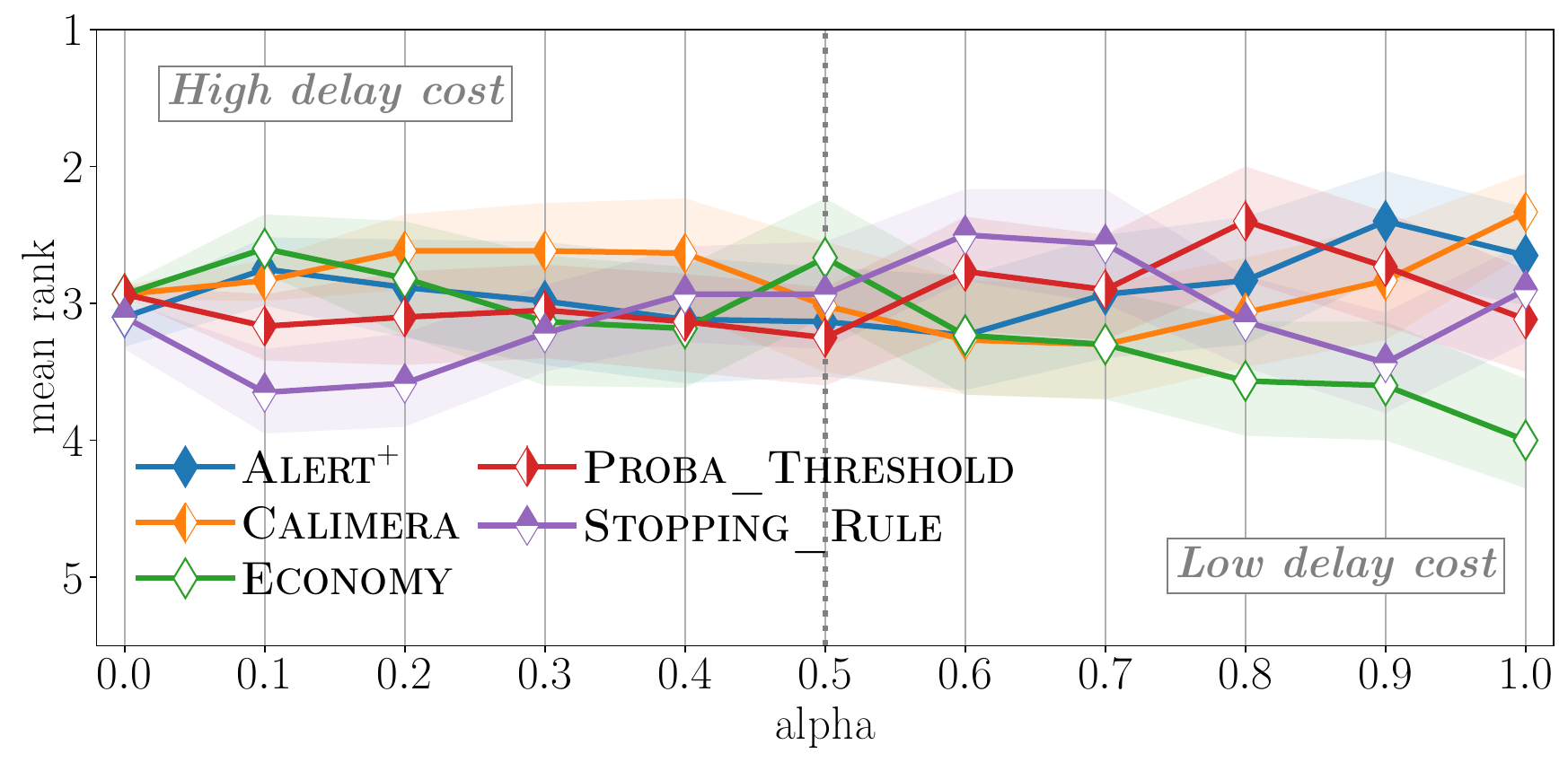}
    \caption{Evolution of the mean ranks, for every $\alpha$, based on the \textit{AvgCost}. Linear delay cost and balanced, binary misclassification cost.}
    \label{fig:bump_linear}
\end{figure}

\textcolor{black}{\textsc{Alert} is further compared with state-of-the-art methods under an alternative cost configuration characterized by a \textit{linear} delay cost and a \textit{balanced} misclassification cost.
%, defined as $C_m(\hat{y}\mid y)= \alpha \times \mathbbm{1}(\hat{y}\neq y)$ and $C_d(t)= (1-\alpha) \times \frac{t}{T}$. 
As illustrated in Figure \ref{fig:bump_linear}, no significant performance differences are observed among the evaluated trigger models in this setting. These findings are broadly consistent with those reported by \cite{renault2024early}, and highlight the importance of investigating alternative cost configurations as well as to find more relevant datasets to better discriminate between competing approaches.}

\section{Conclusion}
\label{sec:conclusion}

This work investigated whether Reinforcement Learning (RL) can provide tangible benefits in Early Classification of Time Series (ECTS). To address this question in a principled way, we focused on separable architectures and explicitly decoupled the time-series classifier from the triggering policy. We introduced \textsc{Alert}, a generic Deep-RL framework for separable, classifier-agnostic ECTS that learns trigger functions on top of any pre-trained classifier, from user-defined state representations. This design enables, to the best of our knowledge, the \emph{first} controlled comparison of RL-based and handcrafted separable triggers under identical information.

Using 30 public datasets, we conducted a systematic study of state space design for RL-based triggers. \textcolor{black}{Our experiments indicate that, given the same input information, RL agents are often competitive with strong handcrafted trigger functions from the ECTS literature.}
By focusing on the impact of the information used by the decision-making function, this paper shows that an informed choice of descriptors can lead to significant performance gains. This choice itself can be considered a hyperparameter to be optimized according to the type of application and the data available, a perspective that is new to the ECTS field and open to easy investigation using the \textsc{Alert} framework. Building on these insights, we derived \textsc{Alert}$^+$, a simple instantiation that combines a small set of widely used ECTS features. \textcolor{black}{\textsc{Alert}$^+$ achieves the best average rank across the considered misclassification--delay trade-offs and often compares favorably with the strongest separable methods in terms of empirical risk, with statistically significant improvements for several cost trade-offs. It also tends to exhibit favorable Pareto fronts over misclassification and delay costs. While our results support RL as a promising approach for trigger learning in ECTS, they do not isolate the effect of RL per se, as part of the observed performance differences may also reflect differences in learning objectives and model parameterization.}

%\textcolor{red}{\textsc{Alert}$^+$ consistently attains the best average rank across a wide range of misclassification--delay trade-offs and, compared to the strongest separable methods, yields statistically significant reductions in empirical risk, while also achieving favorable Pareto fronts over misclassification and delay costs.}

%Beyond raw performance, \textsc{Alert} serves as a unified, modular testbed for early decision-making on time series. 
%By explicitly separating the classifier, state representation, reward design, and RL algorithm, it allows users to attribute performance gains to specific components and systematically explore design choices under different cost settings. \ac{(J'ai l'impression que ce début de paragraphe répète ce qui a déjà été dit plus haut. Je serais pour le retirer).}\textcolor{orange}{(c'était l'idée qu'on offre un 'labo' mais c'est peut-être pas necessaire de le garder ;) ou mal dit ? - Alexis)}
Our published open source implementation is intended both as a reproducible research tool and as a practical asset for time-critical applications, such as industrial monitoring, predictive maintenance, and medical triage.

\textcolor{black}{The reliance on a pre-engineered state space remains a limitation of the current approach. A natural extension would be to study whether jointly learning time-series representations and the decision policy can further improve performance.}
Finally, one of the challenges facing ECTS is early classification in non stationary environments, either when covariate shift occurs (e.g. the prevalence of anomalies changes over time) and/or when there is concept shift (e.g. the same time series may point to a different type of fraud). Furthermore, in the context of ECTS, the prediction itself may alter the future of the time series. (e.g. when braking is triggered on an autonomous vehicle when a collision is predicted). In all these cases, the adaptive nature of RL to non-stationary environments seems to be an interesting avenue to explore.  

\bibliographystyle{tmlr}
\bibliography{biblio.bib}

%\newpage
%\onecolumn
%\clearpage
\appendix

%\tableofcontents
%\section{Deep Q Network background}

\section*{Appendix}
% \addcontentsline{toc}{section}{Appendix}
% \etocsetnexttocdepth{subsection}
% \localtableofcontents

%\mtcaddchapter[Appendix]
%\section{Reproducibility details}
%\subsection{Datasets}
%The datasets we use are the ones introduced in \cite{renault2024early}. This set contains 31 datasets that are not z-normalized and can all be found at \url{https://urlz.fr/qRqu}. This can indeed, cause information leaking as pointed out in \cite{wu2021early}, when not normalizing properly. The code to run the experiments is based on \texttt{PyTorch} \cite{paszke2019pytorch} for automatic differentiation and on \texttt{ml\_edm} \cite{renault2024ml_edm} for general ECTS evaluation functions and interface. The code is made publicly available at \url{https://github.com}

% \subsection{Implementation specifications}

% For all datasets, we use $70\%$ training, $30\%$ testing. Within the training set, $50\%$ is used for classifiers' and $50\%$ for training the trigger model. The classifier module is a collection of \textsc{MiniROCKET} \cite{dempster2021minirocket} estimators, learned over every $5\%$ of the time series. A calibration step is added as in \cite{bilski2023calimera, renault2024early}. For \textsc{Alert}, $30\%$ of the training data is used for validation. We use a single-layer Neural Network with a hidden dimension of 32 to be our Q-estimator. The model is optimized using Adam \cite{kingma2015adam} with a learning rate of $1e^{-4}$. The target network is updated based on parameter $\tau$ equal to $3e^{-3}$.

\section{Experimental protocol}
\label{sec:protocol_app}

\subsection{Datasets}
\label{sec:expe_data_and_costs}
Extensive experiments have been carried out on 30, both synthetic and real-life datasets: 19 from the UCR archive \citep{dau2019ucr} (both binary and multi-class) and 11\footnote{And not 15 as done by \cite{renault2024early}, as in the anomaly detection setting, some of the problems become too hard for the considered classifiers to operate, i.e. there is no performance gain when increasing the number of observations in the time series.} from the Monash time series extrinsic regression archive \citep{tan2021time}, transformed into binary classification tasks. These are the same used as \cite{renault2024early} and can be downloaded from these links\footnote{\url{http://timeseriesclassification.com/dataset.php}, \url{https://urlz.fr/qRqu}}. We selected datasets that are not z-normalized, to avoid possible information leakage \citep{wu2021early}.

\subsection{Competing Approaches}
\label{sec_competing_approaches}

Four competing separable approaches have been selected from the top performers benchmarked by \cite{renault2024early}. In addition, the end-to-end approach \textsc{Earliest}, although not directly comparable since it does not use the same classifier, is considered in our experiments because it is the main RL-based competitor from the literature.
%is still considered the main RL-based competitor. \textcolor{magenta}{Dernière phrase soit mal tournée, soit à déplacer après les 4 bullets ci-dessous}

\begin{itemize}
    \item \textsc{Calimera} \citep{bilski2023calimera} triggers a decision when the value predicted by the regressor models becomes positive. %(see Section \ref{man_tailored}). 
    
    \item \textsc{Economy-$\gamma$-Max} \citep{achenchabe2021early, zafar2021early} triggers a decision if the predicted cost expectation is the lowest at time $t$ when compared with the expected cost for all future time steps. %\ac{(dire un mot sur le cas de tâche de classification binaire ?)}
    
    \item \textsc{Stopping Rule} \citep{mori2017early} uses a linear combination of two confidence levels and a delay measure.
    
    \item \textsc{Proba-Threshold} triggers a prediction if the maximum posterior exceeds some threshold, found by grid search.
    
    \item \textsc{Earliest} \citep{hartvigsen2019adaptive} is the only end-to-end Deep-RL method that has been adapted, in our experiments, to different cost settings by cross-validating the $\lambda$ hyperparameter, measuring the importance of the earliness. 
    %and choose the value minimizing  $AvgCost$ measured on the validation set, computed for the considered cost setting.
\end{itemize}

\subsection{Implementation specifications}
For all datasets, we use $70\%$ training, $30\%$ testing. Within the training set, $50\%$ is used for classifiers' and $50\%$ for training the trigger model. The classifier module is a time-indexed set 
%\textcolor{magenta}{(le mot "collection" pourrait être mal interprété , genre "ensemble" de minrocket, peut être tourner la phrase pour dire un minirocket par taille croissante de la TS par pas de  5\%?)}
of \textsc{MiniROCKET} \citep{dempster2021minirocket} estimators, learned over every $5\%$ of the time series. A calibration step is added as done by \cite{bilski2023calimera, renault2024early}. For \textsc{Alert} models, $30\%$ of the trigger training data is used for validation and model selection. %\textcolor{magenta}{(et le reste pourquoi ? à uniformiser en rédaction avec le 50\%, 50\%)} .
We use a single-layer Neural Network with a hidden dimension of 32 as our Q-estimator. \textcolor{black}{As prediction horizon remains small using time step discretization, we set the discount factor $\gamma=1$}. The model is optimized using Adam \citep{kingma2015adam} with a learning rate of $1e^{-4}$. The target network uses soft updates based on parameter $\tau$ \citep{lillicrap2015continuous} equal to $3e^{-3}$. The code to run the experiments is available on \url{https://github.com/aurelien-renault/ALERT}. It is based on \texttt{PyTorch} \citep{paszke2019pytorch} for automatic differentiation and on \texttt{ml\_edm} \citep{renault2024ml_edm} for general ECTS evaluation functions and interface.

\section{Supplementary Results}
\label{sec:sup_results}

% \subsection{Pairwise comparison: other competitor}\label{app_pairwise}

% \begin{figure}[h]
%     \centering
%     \subfloat[\textsc{Proba threshold}: $\mathcal{S} = \{\textit{max posterior} \}$]{\includegraphics[width=0.9\linewidth]{plots/pairwise_sota/tex/alert_pt.pdf}\label{pair_pt}}
%     \caption{Pairwise comparison of \textsc{Proba Threshold} vs. RL counterpart using same information as input. Points above the horizontal line indicates that the man-tailored method is better than its RL-based counterpart}
%     \label{fig:enter-label}
% \end{figure}

\subsection{Pairwise comparison: statistical tests}
\label{wilco_pair}

\begin{table}[h]
    \centering
    \caption{Wilcoxon tests p-values, comparing pairwise man-tailored and RL-based algorithms. \textbf{Bold}, resp. \textit{Italic} values indicate a value below the significance level equal to 0.05, in favor of handcrafted method, resp. RL-based method. \textcolor{black}{When all \textit{AvgCost}'s are identical, i.e. there is only ties, p-value computation is undefined.}}
    \resizebox{\linewidth}{!}{
    \begin{tabular}{cc|c|c|c|c|c|c|c|c|c|c|c|}
    \cline{3-13}
    & $\alpha$ $\rightarrow$ & \multirow{2}{*}{\textbf{0  }} & \multirow{2}{*}{\textbf{0.1}} & \multirow{2}{*}{\textbf{0.2}} & \multirow{2}{*}{\textbf{0.3}} & \multirow{2}{*}{\textbf{0.4}} & \multirow{2}{*}{\textbf{0.5}} & \multirow{2}{*}{\textbf{0.6}} & \multirow{2}{*}{\textbf{0.7}} & \multirow{2}{*}{\textbf{0.8}} & \multirow{2}{*}{\textbf{0.9}} & \multirow{2}{*}{\textbf{1}} \\
    \textbf{method} $\downarrow$ & & & & & & & & & & & & \\
    \hline 
    \multicolumn{2}{|c|}{\textsc{Economy}} & NaN & 0.062 & 0.091 & 0.082 & \cellcolor{lightgray}\textbf{0.003} & \cellcolor{lightgray}\textbf{<1e-3} & \cellcolor{lightgray}\textbf{<1e-3} & \cellcolor{lightgray}\textbf{<1e-3} & \cellcolor{lightgray}\textbf{<1e-3} & \cellcolor{lightgray}\textbf{0.001} & 0.421 \\
    \multicolumn{2}{|c|}{\textsc{Stopping Rule}} & \textit{0.046} & 0.672 & 0.701 & 0.635 & 0.15 & \cellcolor{lightgray}\textbf{0.01} & 0.134 & 0.931 & 0.111 & 0.141 & 0.673 \\
    \multicolumn{2}{|c|}{\textsc{Calimera}} & 0.317 & 0.464 & 0.522 & 0.644 & 0.799 & 0.961 & 0.604 & 0.474 & 0.161 & 0.069 & 0.086 \\
    \multicolumn{2}{|c|}{\textsc{Proba-threshold}} & 0.317 & 0.803 & 0.066 & 0.097 & \cellcolor{lightgray}\textbf{0.005} & 0.074 & 0.112 & \cellcolor{lightgray}\textbf{0.036} & 0.953 & \cellcolor{lightgray}\textbf{0.018} & 0.386 \\
    \hline
    \end{tabular}}
    \label{tab:wilco_pair}
\end{table}

\subsection{State space: adding time series}
\label{app_state_space}

As a baseline, we enlarge \textsc{Alert}$^+$ state space with twenty randomly drawn points (\textsc{Alert}$^+$ \textsc{Random}).

\begin{figure}[H]
    \centering
    \includegraphics[width=0.65\linewidth]{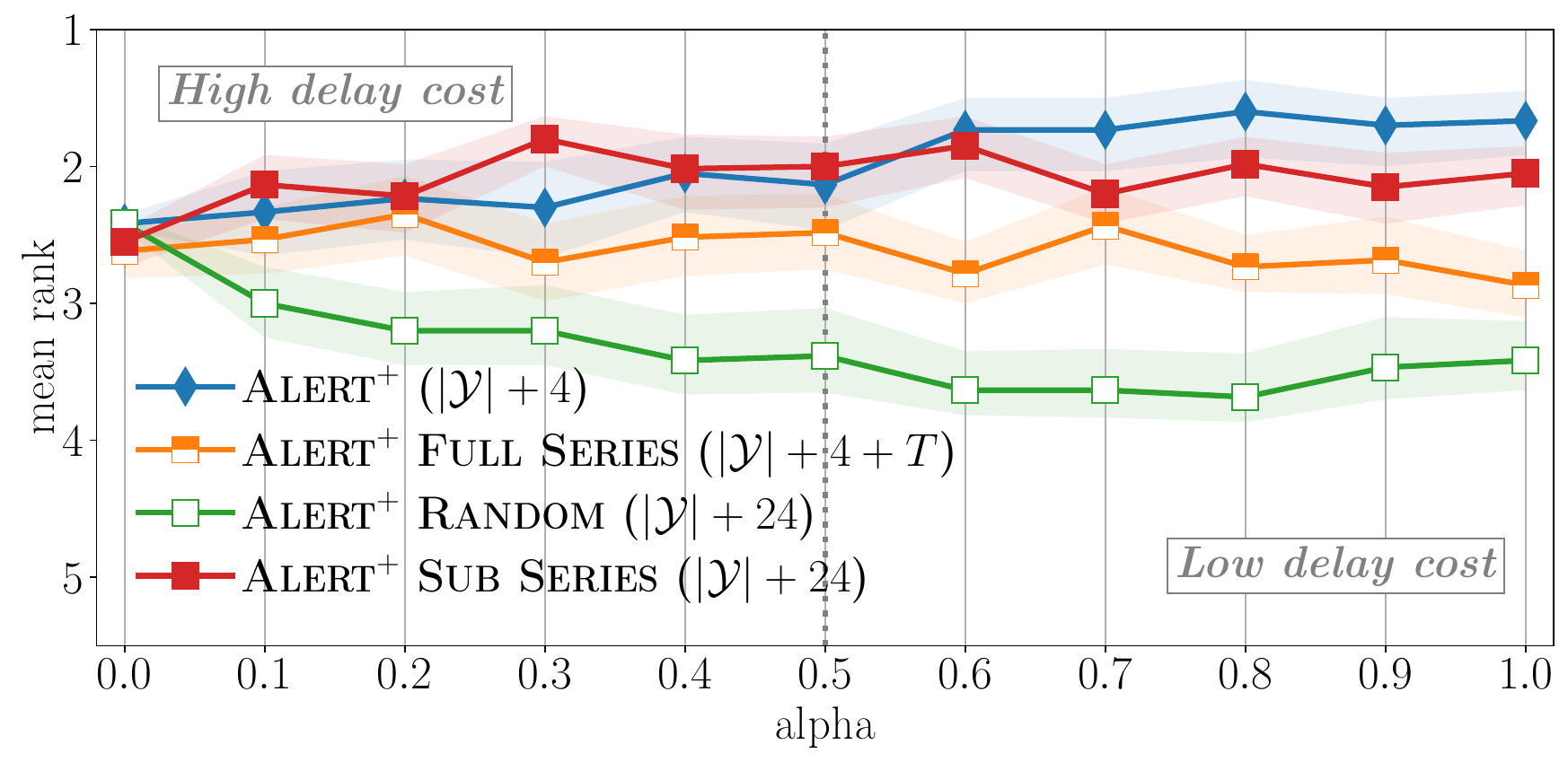}
    \caption{Evolution of the mean ranks, for every $\alpha$, based on the \textit{AvgCost}. Adding time series on \textsc{Alert$^{+}$}' state space.}
    \label{fig:bump_ablation}
\end{figure}

Figure \ref{fig:bump_ablation} shows that even though \textsc{Alert}$^+$ has the most limited state space in size in this comparison, it tops both \textsc{Alert}$^+$ \textsc{Full Series}  and \textsc{Alert}$^+$ \textsc{Random} for all values of $\alpha$, and is better than \textsc{Alert}$^+$ \textsc{Sub Series} for low delay cost ($\alpha \geq 0.6$) and equal or second for $\alpha \leq 0.5$. Furthermore, the size of the state space itself is not a predictor of performance, since \textsc{Alert}$^+$ \textsc{Random} is the last for all values of $\alpha$. The choice of input features plays a role in performance, as evidenced by the fact, with the same state space size, \textsc{Alert}$^+$ \textsc{Series} clearly outperforms \textsc{Alert}$^+$ \textsc{Random}.

\subsection{\textsc{Alert$^{+}$} vs. SOTA: statistical tests}
\label{wilco_alert}

\begin{table}[H]
    \centering
    \caption{Wilcoxon tests p-values, comparing \textsc{Alert$^+$} to state-of-the-art algorithms. \textbf{Bold} values indicate a value below the significance level equal to 0.05, in favor of \textsc{Alert$^+$}. \underline{Underline} values indicate p-values below original significance level, but not below the Holm's \citep{holm1979simple} corrected value, that depends on the number of tested hypothesis.}
    \resizebox{\linewidth}{!}{
    \begin{tabular}{cc|c|c|c|c|c|c|c|c|c|c|c|}
    \cline{3-13}
    & $\alpha$ $\rightarrow$ & \multirow{2}{*}{\textbf{0  }} & \multirow{2}{*}{\textbf{0.1}} & \multirow{2}{*}{\textbf{0.2}} & \multirow{2}{*}{\textbf{0.3}} & \multirow{2}{*}{\textbf{0.4}} & \multirow{2}{*}{\textbf{0.5}} & \multirow{2}{*}{\textbf{0.6}} & \multirow{2}{*}{\textbf{0.7}} & \multirow{2}{*}{\textbf{0.8}} & \multirow{2}{*}{\textbf{0.9}} & \multirow{2}{*}{\textbf{1}} \\
    \textbf{\textsc{Alert$^+$} vs.} $\downarrow$ & & & & & & & & & & & & \\
    \hline 
    \multicolumn{2}{|c|}{\textsc{Economy}} & 1.0 & 0.759 & \underline{0.034} & 0.083 & \underline{0.024} & \underline{0.049} & \underline{0.016} & \cellcolor{lightgray}\textbf{<1e-3} & \cellcolor{lightgray}\textbf{<1e-3} & \cellcolor{lightgray}\textbf{<1e-3} & \cellcolor{lightgray}\textbf{0.002} \\
    \multicolumn{2}{|c|}{\textsc{Calimera}} & 1.0 & 0.58 & \underline{0.041} & 0.342 & \underline{0.006} & \underline{0.029} & \underline{0.009} & \cellcolor{lightgray}\textbf{0.004} & \cellcolor{lightgray}\textbf{<1e-3} & \cellcolor{lightgray}\textbf{0.002} & \cellcolor{lightgray}\textbf{0.001} \\
    \multicolumn{2}{|c|}{\textsc{Stopping Rule}} & 0.208 & 0.129 & 0.086 & 0.41 & \cellcolor{lightgray}\textbf{0.015} & 0.05 & \cellcolor{lightgray}\textbf{0.002} & \cellcolor{lightgray}\textbf{<1e-3} & \cellcolor{lightgray}\textbf{<1e-3} & \cellcolor{lightgray}\textbf{<1e-3} & \cellcolor{lightgray}\textbf{<1e-3} \\
    \multicolumn{2}{|c|}{\textsc{Proba-threshold}} & 1.0 & 0.223 & 0.078 & \cellcolor{lightgray}\textbf{0.037} & \cellcolor{lightgray}\textbf{0.002} & \cellcolor{lightgray}\textbf{0.002} & \cellcolor{lightgray}\textbf{<1e-3} & \cellcolor{lightgray}\textbf{<1e-3} & \cellcolor{lightgray}\textbf{<1e-3} & \cellcolor{lightgray}\textbf{<1e-3} & \cellcolor{lightgray}\textbf{<1e-3} \\
    \multicolumn{2}{|c|}{\textsc{Earliest}} & \cellcolor{lightgray}\textbf{<1e-3} & 0.347 & 0.176 & 0.327 & \cellcolor{lightgray}\textbf{0.014} & \cellcolor{lightgray}\textbf{0.006} & \cellcolor{lightgray}\textbf{<1e-3} & \cellcolor{lightgray}\textbf{<1e-3} & \cellcolor{lightgray}\textbf{<1e-3} & \cellcolor{lightgray}\textbf{<1e-3} & \cellcolor{lightgray}\textbf{<1e-3} \\
    \hline
    \end{tabular}}
    \label{tab:wilco_alert}
\end{table}

\subsection{\textsc{Calimera}$^+$: \textsc{Calimera} with \textsc{Alert}$^+$' state space}
\label{supervised_appendix}

\begin{figure}[H]
    \centering
    \includegraphics[width=0.65\linewidth]{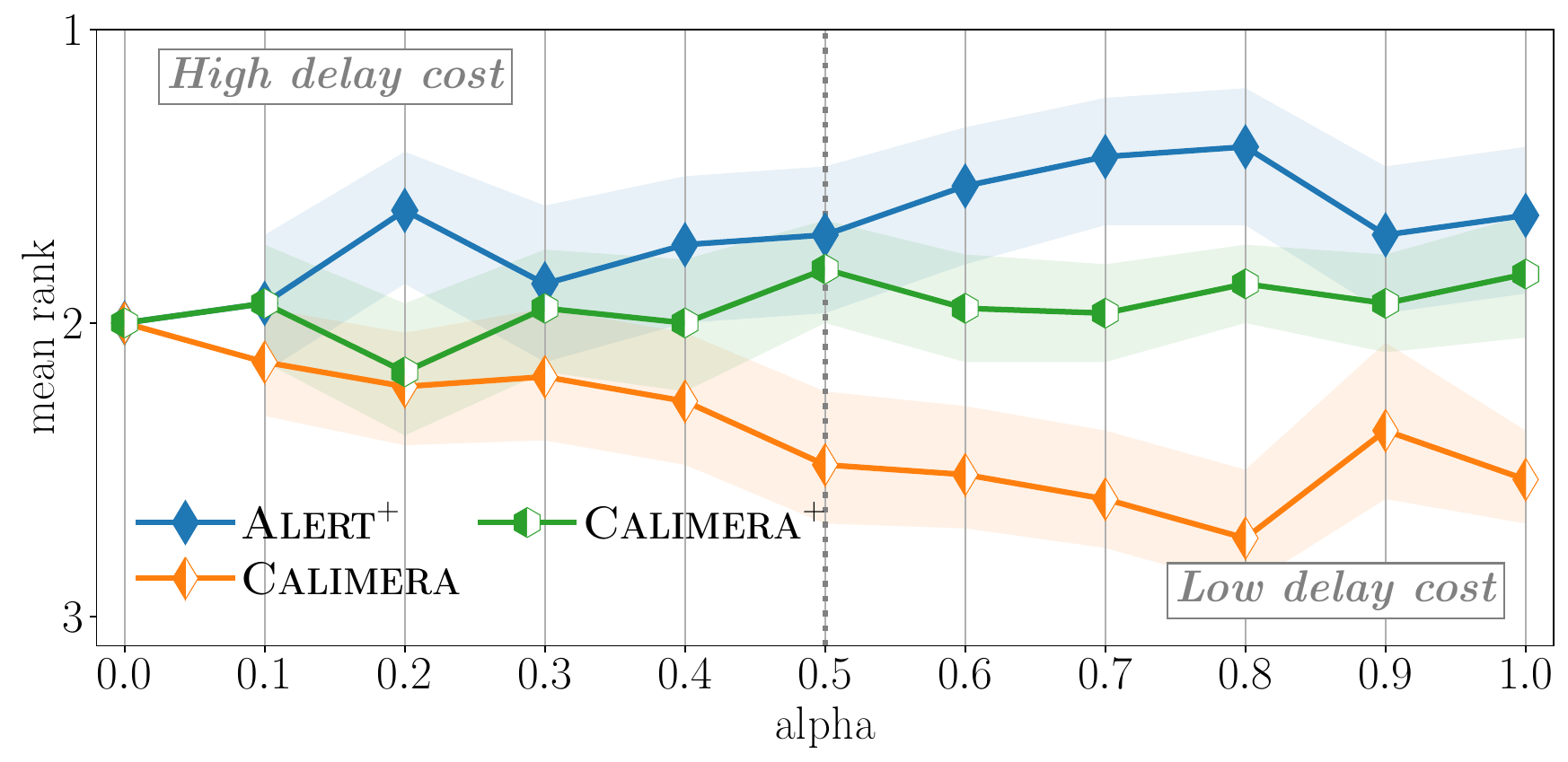}
    \caption{Evolution of the mean ranks, for every $\alpha$, based on the \textit{AvgCost}. Providing \textsc{Calimera} with \textsc{Alert$^{+}$}' state space.}
    \label{fig:bump_cal_plus}
\end{figure}

\textcolor{black}{To further examine the extent to which the observed performance gains are attributable to the richer state representation, we also evaluate a variant of \textsc{Calimera} equipped with the same state space as \textsc{Alert}$^+$, referred to as \textsc{Calimera}$^+$. Since \textsc{Calimera} formulates triggering as a non-myopic cost regression problem, it can be viewed as a supervised counterpart to \textsc{Alert}. As shown in Figure~\ref{fig:bump_cal_plus}, enriching \textsc{Calimera} with the \textsc{Alert}$^+$ features improves its average ranking. However, its performance generally remains below that of \textsc{Alert}$^+$, with statistically significant differences for $\alpha \in [0.6,0.8]$. These results suggest that the state representation likely contributes substantially to the strong performance of \textsc{Alert}, while also indicating that it does not fully account for the observed gap.}

\subsection{Scatter plots: other competitors}
\label{app_scatter}

\begin{figure}[H]
    \centering
    \subfloat[\textsc{Stopping Rule}]{\includegraphics[width=0.33\linewidth]{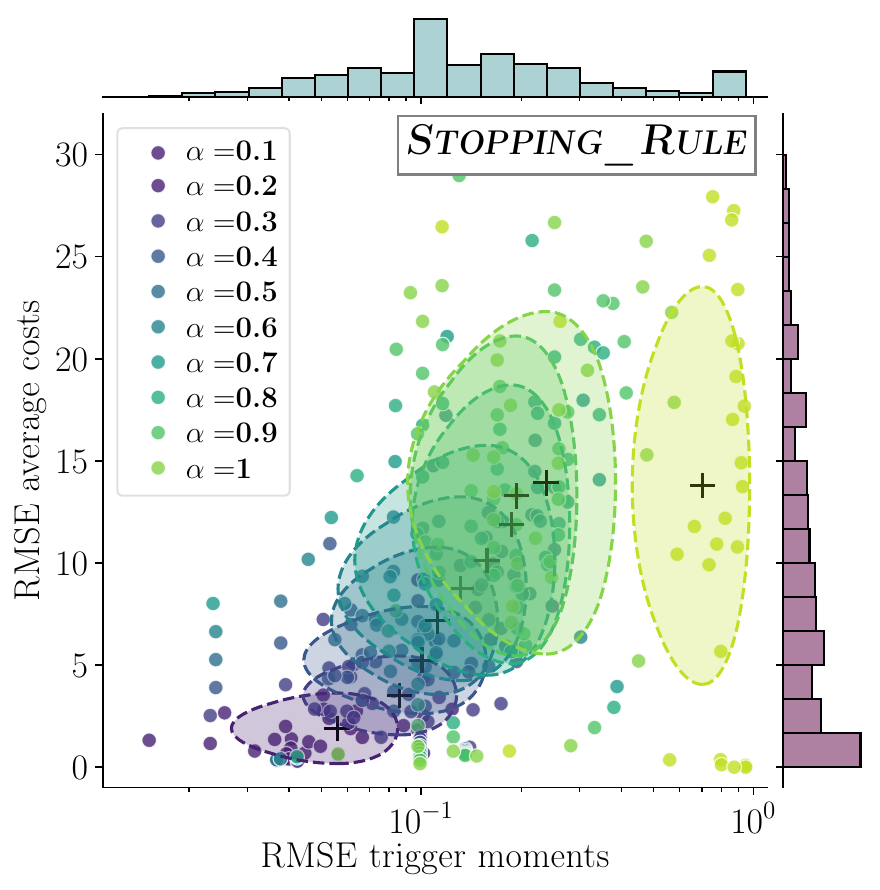}} 
    \subfloat[\textsc{Proba-threshold}]{\includegraphics[width=0.33\linewidth]{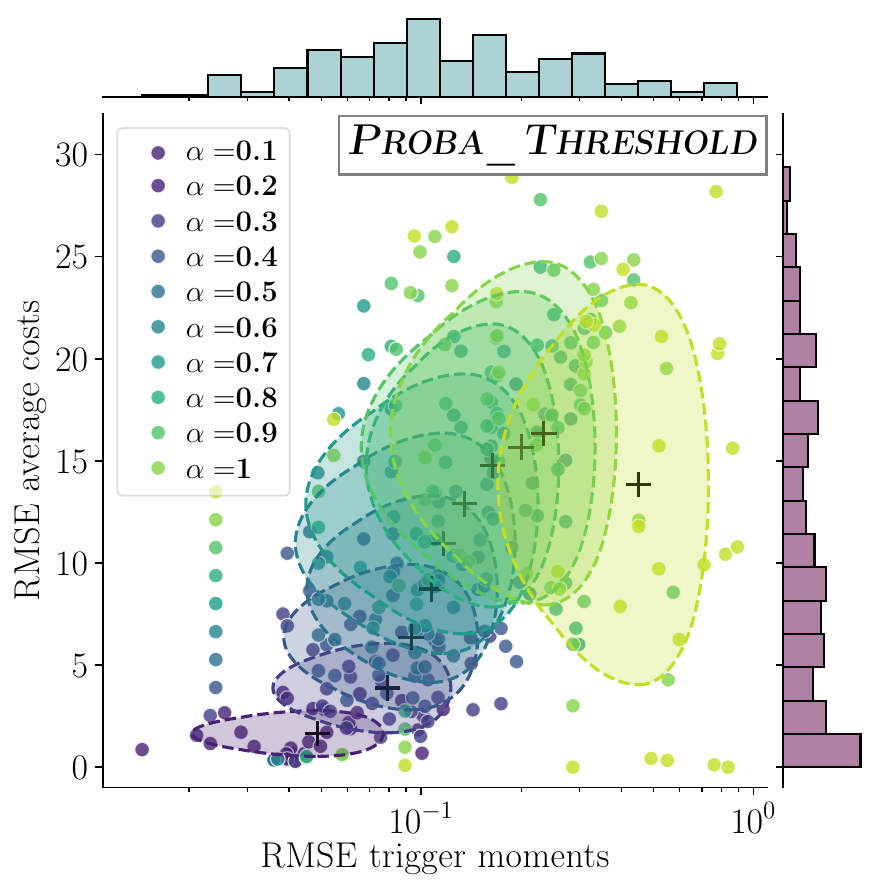}}
    \caption{The $x$-axis reports how far is the triggering time from the best a posteriori one: left is better. The $y$-axis reports the difference between the \textit{AvgCost} incurred by the algorithm compared to the best a posteriori one, \textit{AvgCost}$^{\star}$ : lower is better. The black crosses report the average performance for each value of $\alpha$ (greater values correspond to higher relative importance of the delay cost). Ellipses display $2 \times$ the standard deviation over both axis, computed for each $\alpha$ value.}
    \label{fig:enter-label}
\end{figure}

% \subsection{Linear/balanced cost setting}
% \label{linear_appendix}

% \textcolor{black}{\textsc{Alert} is further compared with state-of-the-art methods under an alternative cost configuration characterized by a \textit{linear} delay cost and a \textit{balanced} misclassification cost, defined as $C_m(\hat{y}\mid y)=\mathbbm{1}(\hat{y}\neq y)$ and $C_d(t)=\frac{t}{T}$. As illustrated in Figure \ref{fig:bump_linear}, no significant performance differences are observed among the evaluated trigger models in this setting. These findings are broadly consistent with those reported by \cite{renault2024early}, and highlight the importance of investigating alternative cost configurations to better discriminate between competing approaches.}

% \begin{figure}[H]
%     \centering
%     \includegraphics[width=0.65\linewidth]{plots/complete_sota/ranks/tex/bump_linear.pdf}
%     \caption{Evolution of the mean ranks, for every $\alpha$, based on the \textit{AvgCost}. Linear delay cost and balanced, binary misclassification cost.}
%     \label{fig:bump_linear}
% \end{figure}

% \subsection{No RL ablation}
% \label{rl_ablation}

% \begin{figure}[!htb]
%     \centering
%     \includegraphics[width=0.75\linewidth]{plots/complete_sota/ranks/tex/bump_calimeralert.pdf}
%     \caption{Evolution of the mean ranks, for every $\alpha$, based on the \textit{AvgCost}.}
%     \label{fig:bump_calimeralert}
% \end{figure}

\section{Runtime analysis}
\label{runtime}

\textcolor{black}{The runtime analysis presented in Table \ref{tab:times} shows that \textsc{Alert}$^+$ requires substantially longer training times than competing methods. This additional computational cost arises from the regularization mechanisms introduced to stabilize the RL process, which inevitably increase training overhead. However, once training is complete, \textsc{Alert}$^+$ achieves inference speeds comparable to those of other approaches, ensuring that its deployment in real-world applications is not compromised.}

\textcolor{black}{This substantial training time overhead can be worth paying in applications where reducing the \textit{AvgCost} is critical and/or where the trigger function is expected to be non-linear. Its flexibility, especially in incorporating additional information into the state space, is also an argument for favoring \textsc{Alert} over cheaper models.}

\begin{table}[h]
    \centering
    \caption{Computation times (in \textit{sec.}) for $\alpha=0.8$, five threads.}
    \begin{tabular}{c|cc|c}
         Trigger & Median Fit & Median Inf. & Total \\
         \hline
         \textsc{Alert}$^+$ & 123.057 & 0.011 & 9628.169 \\
         \textsc{Calimera} & 5.459 & 0.026 & 177.848 \\
         \textsc{Economy} & 1.039 & 0.025 & 154.717 \\
         \textsc{Proba-Threshold} & 0.052 & 0.024 & 5.305 \\
         \textsc{Stopping Rule} & 2.954 & 0.01 & 292.061 \\
    \end{tabular}
    \label{tab:times}
\end{table}

\begin{figure}[H]
    \centering
    \includegraphics[width=0.85\linewidth]{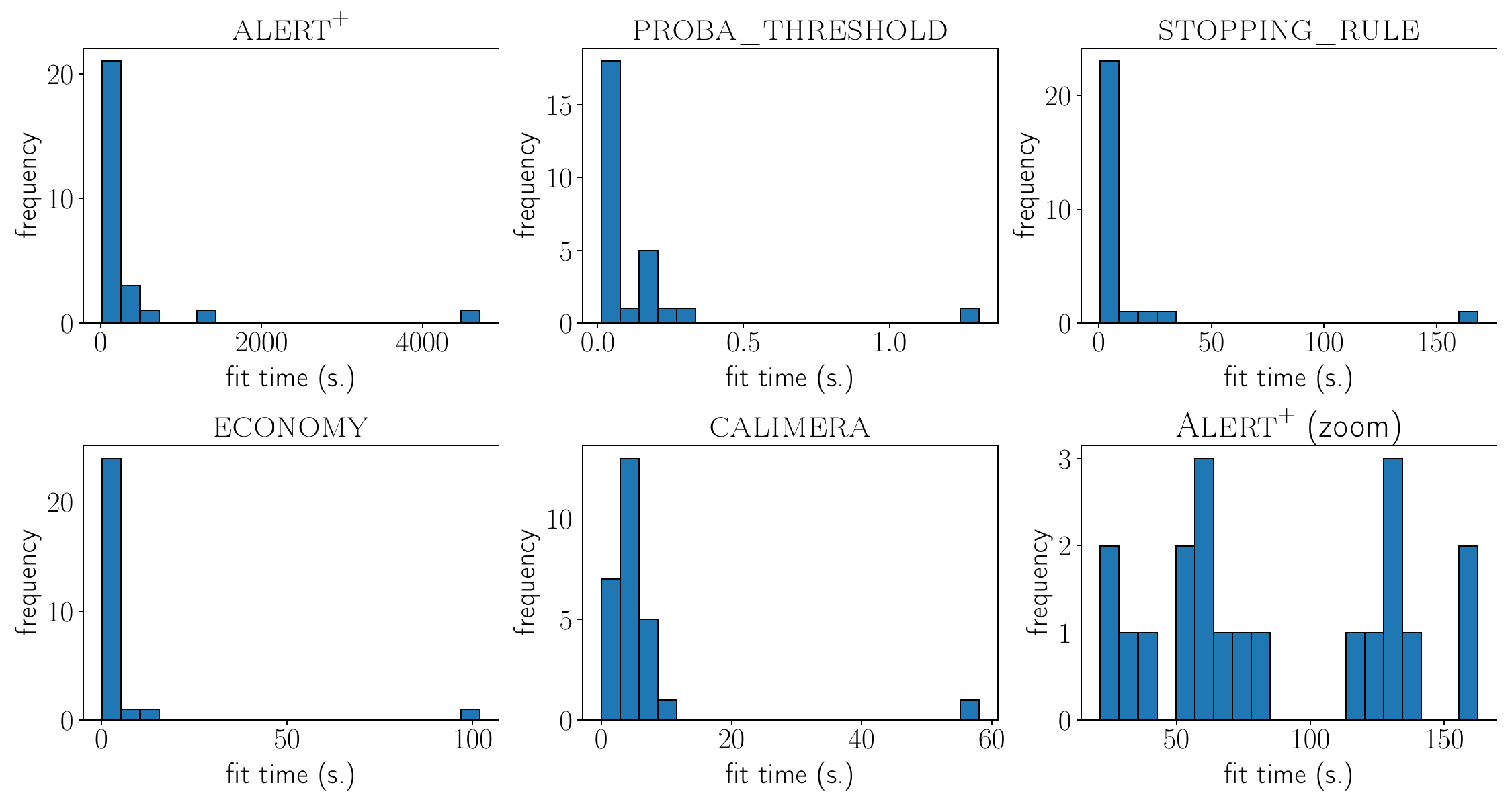}
    \caption{Histograms of fit times.}
    \label{fig:hist_times}
\end{figure}

\section{Sensitivity studies}
\label{sensitive}

\subsection{State space ablation}
\label{state_ablation}

% Each of the features that composed the \textsc{Alert}$^+$ state space is systematically removed, in order to assess the relative importance of each of them. For $\alpha=0.8$, Figure \ref{fig:cd_state_ablation} shows that the \textit{time} information is what degrade the most the performance when being removed, followed by the \textit{maximum posterior}. In third position, the \textit{level of confidence}, which brings information about the distribution of maximum posteriors over the training dataset. Finally, both the \textit{margin} and the \textit{prediction} itself don't seem to bring much performance within this setting.

\begin{figure}[H]
    \centering
\includegraphics[width=0.7\linewidth]{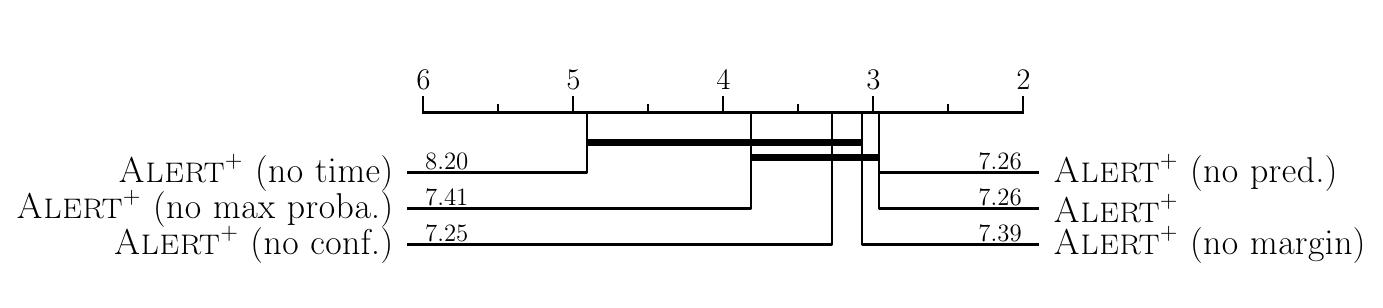}
    \caption{Wilcoxon signed-rank test labeled with mean \textit{AvgCost}, $\alpha = 0.8$.}
    \label{fig:cd_state_ablation}
\end{figure}

\textcolor{black}{
Each feature composing the \textsc{Alert}$^+$ state space is systematically removed to assess its relative contribution. For $\alpha = 0.8$, Figure \ref{fig:cd_state_ablation} indicates that removing the \textit{time} information leads to the largest performance degradation, followed by the removal of the \textit{maximum posterior}. The third most influential component is the \textit{level of confidence}, which captures information about the distribution of maximum posterior probabilities over the training dataset. In contrast, the \textit{margin} and the \textit{predicted label} appear to contribute only marginally to performance in this setting.}

\subsection{Reward function}
\label{sec:expe_sensitivity_reward}

% \paragraph{Delayed rewards}
% \label{app_end}

% The delayed reward function is tested here, i.e. giving fully paid cost once the trigger action has been chosen. Figure \ref{fig:bump_end} shows that even without having intermediate time rewards, \textsc{Alert$^{\star}$} still manages to outperform state-of-the-art algorithms. Thus, knowing how to decompose both misclassification and delay cost is not a strong requirement for \textsc{Alert$^{\star}$} to perform.

% \begin{figure}[!htb]
%     \centering
%     \includegraphics[width=0.75\linewidth]{plots/complete_sota/ranks/tex/bump_reward_end.pdf}
%     \caption{Evolution of the mean ranks, for every $\alpha$, based on the \textit{AvgCost} metric. Delayed reward function.}
%     \label{fig:bump_end}
% \end{figure}

As outlined in Section \ref{sec_reward_fct}, there exist multiple choices for the reward function. Since ECTS is a new problem for RL, it is important to assess whether some choices lead to significantly better results or if the reward function has little impact.  One debate is whether it is better to postpone the delay cost signal to the end of an episode at the triggering time or if it should be distilled at each time step. Another consideration is which delay cost signal should be provided during learning. Specifically, should it incorporate information about the \textit{a posteriori} best decision time $t^\star$?

To answer these questions, the following reward functions have been tested in our experiments using the same datasets as previously stated:

%Throughout this section, we'll study the impact of using different formulations for the reward function. The \textsc{Alert}$^+$ state space is kept constant in the following. In particular, the following definitions will be tested : 

\begin{align}
    r(t) &= \left\{
    \begin{array}{ll}
         -\Delta \mathrm{C}_d(t) & \mbox{if } a_t=\text{``wait''} \\  
         -C_m(\hat{y}_t|y) - \Delta \mathrm{C}_d(t) & \mbox{if } a_t=\text{``trigger''}
    \end{array}
    \right.
    \tag{R1} \label{eq:r1}
\end{align}

\begin{align}
    r(t) &= \left\{
    \begin{array}{ll}
         -\mathbbm{1}(t > t^{\star}) \Delta \mathrm{C}_d(t) & \mbox{if } a_t=\text{``wait''} \\  
         -|\mathrm{C}_m(\hat{y}_t|y) -\mathrm{C}_m(\hat{y}_{t^\star}|y)|  & \multirow{2}{*}{\mbox{if } $a_t=\text{``trigger''}$} \\ 
         \hspace{4mm} -\mathbbm{1}(t > t^{\star}) \Delta \mathrm{C}_d(t)
    \end{array}
    \right.
    \tag{R2} \label{eq:r2}
\end{align}
\begin{align}
    r(t) &= \left\{
    \begin{array}{ll}
         0 & \mbox{if } a_t=\text{``wait''}  \\  
         -|t - t^\star|  & \mbox{if } a_t=\text{``trigger''}
    \end{array}
    \right.
    \tag{R3} \label{eq:r3}
\end{align}

For both reward functions \ref{eq:r1} and \ref{eq:r2}, alternate versions in which all costs are given only at trigger time (i.e. end of episode), denoted with the ``\textsc{End}'' suffix, are tested. For all the experiments carried out, the \textsc{Alert}$^+$ state space is kept constant and equal to: $\mathcal{S} = \{\textit{max posterior},\: \textit{margin},$ $\textit{pred. class}$,\: $\textit{level of confidence},\: \textit{time} \}$.

\begin{figure}[H]
    \centering
    \includegraphics[width=0.65\linewidth]{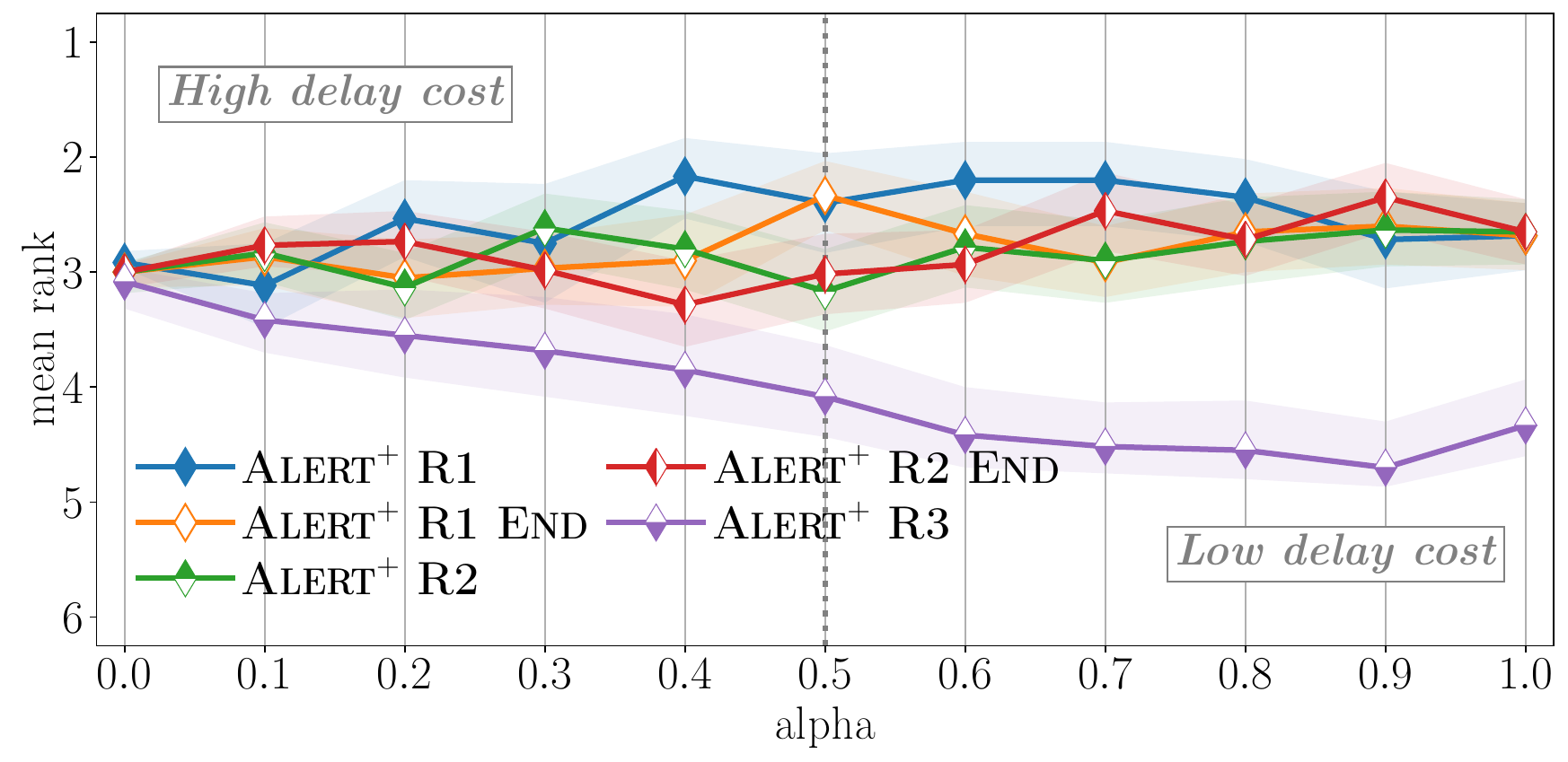}
    \caption{Robustness study on \textsc{Alert$^{+}$} reward's function.}
    \label{fig:bump_ablation_reward}
\end{figure}

Figure \ref{fig:bump_ablation_reward} synthesizes the results by comparing the ranking of five versions of \textsc{Alert}$^+$ using the alternative reward functions. The first conclusion is that not providing the misclassification cost (\textsc{Alert}$^+$ R3) leads to uniformly bad performance. The second one is that no other reward function can be singled out as the best one. In particular, no difference exists between providing the reward signal only at the end of each learning episode, or distilling the delay cost at each time step, even though \textsc{Alert}$^+$ R1 seems a little bit better for intermediate values of $\alpha$. Similarly, information about $t^\star$ does not provide an advantage. This is consistent with the findings of figure \ref{fig:scatter_alert} that shows that \textsc{Alert}$^+$, the best performing algorithm, tends to move away from $t^\star$ to achieve better \textit{AvgCost}s. 
%\textcolor{red}{in order to minimize the cost paid(?) (AC : à mon avis pas la peine d'insister sur ce que cherche l'algorithme)}.

It thus appears that using the reward function \ref{eq:r1} is a good choice.

\subsection{Discount factor $\gamma$}

\begin{figure}[H]
    \centering
    \includegraphics[width=0.5\linewidth]{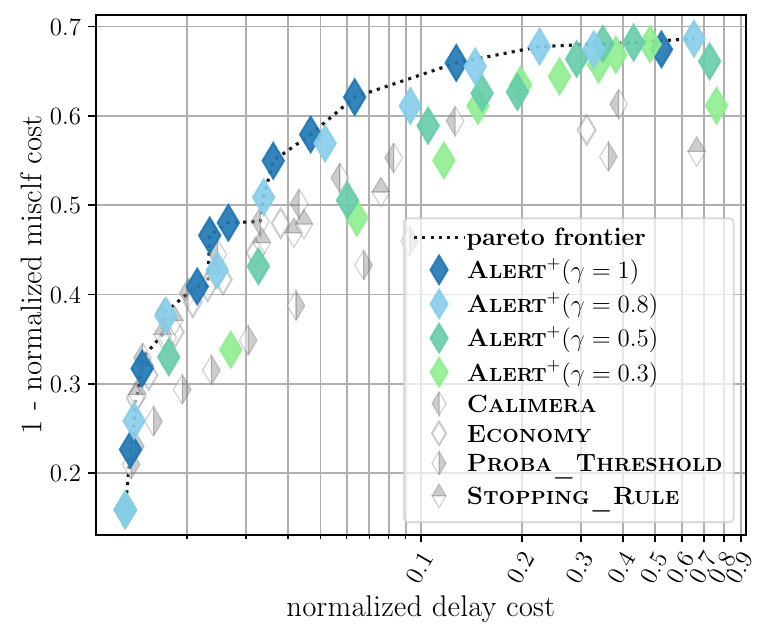}
    \caption{Pareto front, displaying for each $\alpha$, the normalized version of the \textit{AvgCost}, decomposed over delay and one minus misclassification cost on $x$-axis and $y$-axis respectively. Best approaches are located on the top left corner. High $\alpha$ values are located on the right, low ones on the left. Due to the exponential shape of the delay cost, the $x$-axis is on log scale.}
    \label{fig:pareto_myopic}
\end{figure}

\textcolor{black}{When varying the discount factor $\gamma$, \textsc{Alert}$^+$ generally exhibits a tendency to postpone its decisions. This behavioral shift leads to degraded performance for low values of $\alpha$, corresponding to settings with high temporal costs. In contrast, when the temporal cost is less pronounced (top-right region of Figure \ref{fig:pareto_myopic}), \textsc{Alert}$^+$ remains close to the Pareto front even for low values of $\gamma$. Notably, for $\alpha = 1$, the configuration with $\gamma = 0.8$ outperforms the original \textsc{Alert}$^+$, highlighting the potential regularization effect of $\gamma$ in regimes characterized by low temporal pressure.}

\subsection{DQN size}

\begin{figure}[h]
    \centering
\includegraphics[width=0.7\linewidth]{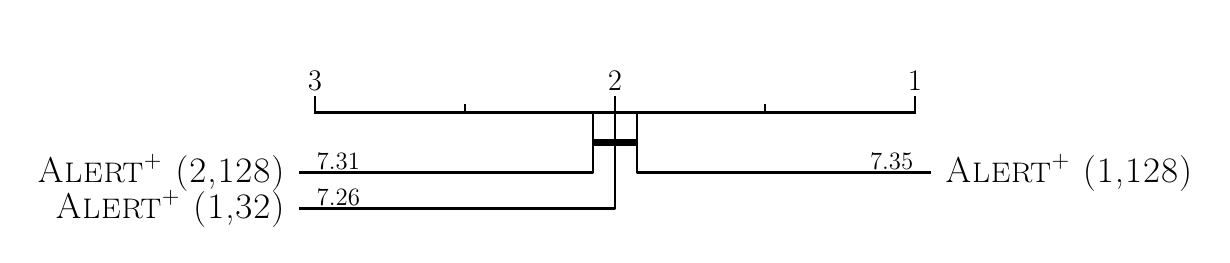}
    \caption{Wilcoxon signed-rank test labeled with mean \textit{AvgCost}, $\alpha = 0.8$. DQN architecure is  denoted as $(\text{num. of layers}, \text{hidden dim.})$.}
    \label{fig:cd_nn_size_08}
\end{figure}

\textcolor{black}{In this section, we evaluate several DQN architectures for \textsc{Alert}$^+$. To limit computational overhead, experiments are conducted solely with $\alpha = 0.8$. As shown in Figure \ref{fig:cd_nn_size_08}, increasing the number of layers and/or the hidden dimension does not yield significant performance improvements. Notably, the lowest \textit{AvgCost} is achieved by the simplest architecture, consisting of a single hidden layer with 32 units, i.e. base choice for \textsc{Alert}$^+$.}

% In this section, several DQN architectures are evaluated for \textsc{Alert}. To reduce computation time, only $\alpha=0.8$ is examined. Figure \ref{fig:cd_nn_size_08} shows that there's no significant differences when growing the number of layers and/or the hidden dimension. Specifically, the lowest \textit{AvgCost} is obtained with the simplest architecture, i.e. one hidden layer of size 32. 

\subsection{Sampling ratio}
\label{app_sampling}

\begin{figure}[H]
    \centering
    \subfloat[Sampling ratio equal to $2\%$]{
    \includegraphics[width=0.7\linewidth]{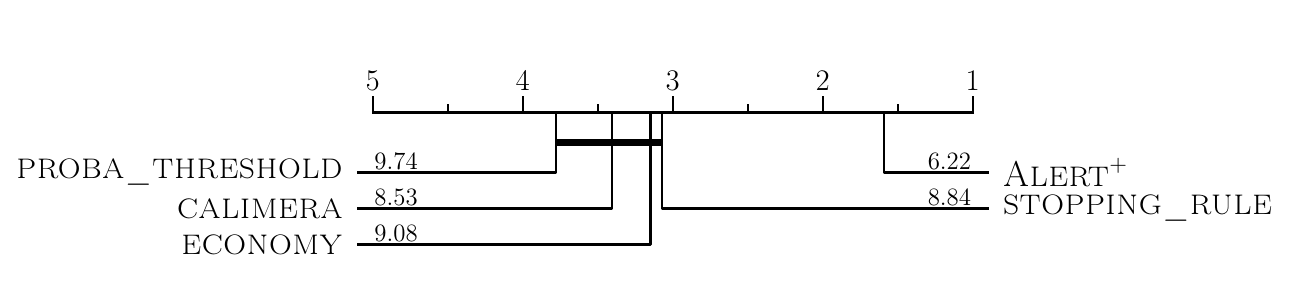}
    } \\
    \subfloat[Sampling ratio equal to $1\%$]{
    \includegraphics[width=0.7\linewidth]{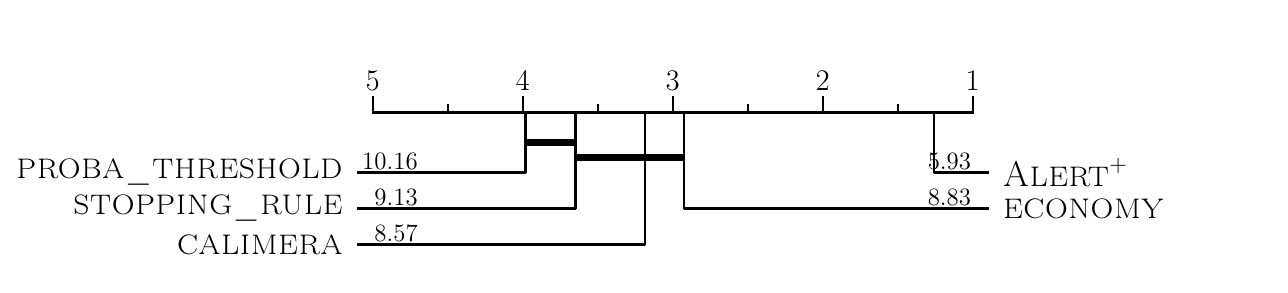}
    }
    \caption{Wilcoxon signed-rank test labeled with mean \textit{AvgCost}, $\alpha = 0.8$.}
    \label{fig:cd_sampling}
\end{figure}

We further evaluate the robustness of the results with respect to the sampling frequency of the time series. In this experiment, finer-grained sampling resolutions, specifically every $1\%$ and $2\%$ of the time series, are considered for $\alpha = 0.8$. As shown in Figure \ref{fig:cd_sampling}, \textsc{Alert}$^+$ continues to significantly outperform state-of-the-art methods, and the overall ranking of the approaches remains globally consistent across different sampling resolutions. The only notable change is that \textsc{Proba-Threshold}, although still ranked last, is no longer significantly different from the other state-of-the-art methods.

\subsection{Encoding}

\begin{figure}[H]
    \centering
    \subfloat[SAX features]{
    \includegraphics[width=0.7\linewidth]{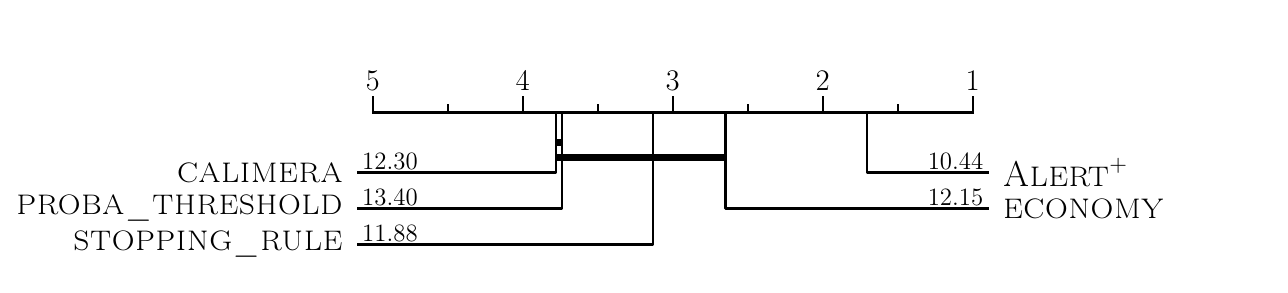}
    } \\
    \subfloat[Catch22 features]{
    \includegraphics[width=0.7\linewidth]{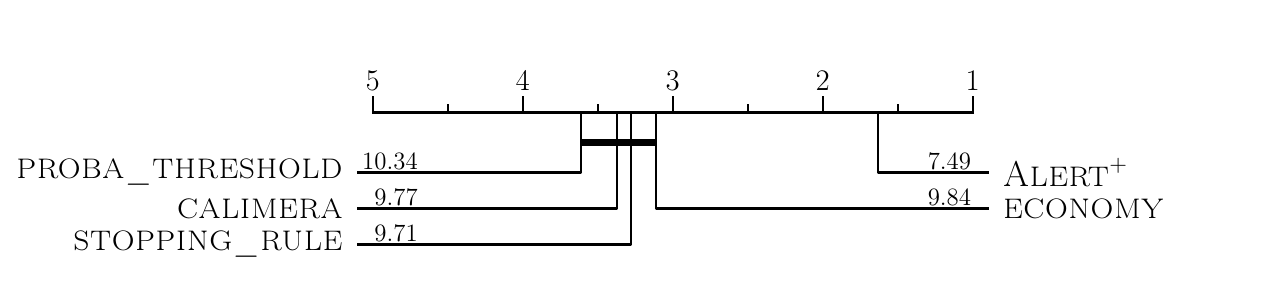}
    }
    \caption{Wilcoxon signed-rank test labeled with mean \textit{AvgCost}, $\alpha = 0.8$.}
    \label{fig:cd_clfs}
\end{figure}

As an alternative to MiniROCKET for feature generation, Catch22 \cite{lubba2019catch22}  and SAX \cite{lin2007experiencing} are evaluated. Figure \ref{fig:cd_clfs} shows that the overall ranking of the methods is only marginally affected by the choice of feature extractor, with \textsc{Alert}$^+$ continuing to outperform state-of-the-art competitors.

\subsection{Classifier}

\begin{figure}[H]
    \centering
    \includegraphics[width=0.65\linewidth]{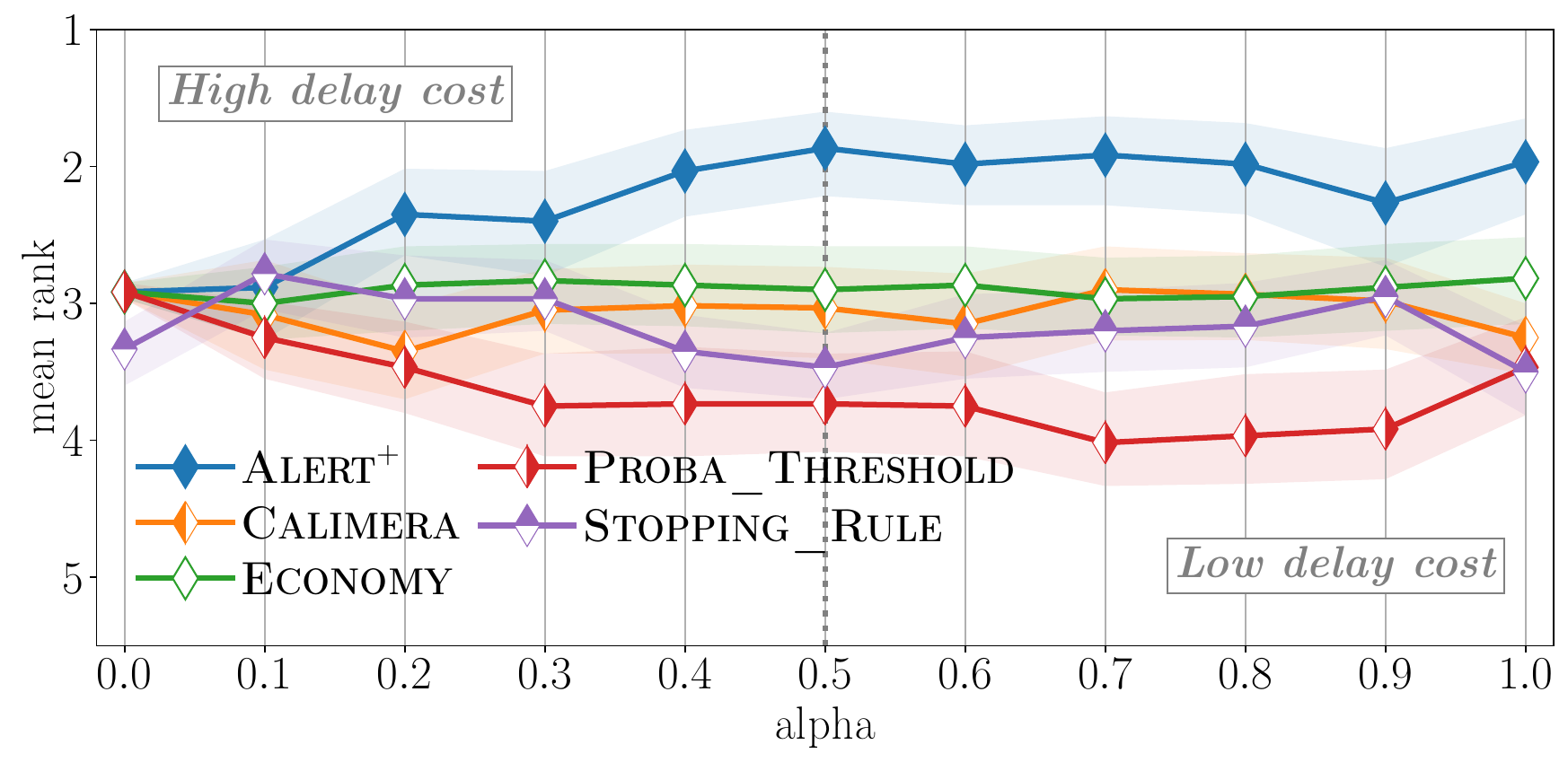}
    \caption{Evolution of the mean ranks, for every $\alpha$, based on the \textit{AvgCost}. Backbone classifier is a GBM.}
    \label{fig:bump_gbm}
\end{figure}

\begin{figure}[H]
    \centering
    \includegraphics[width=0.65\linewidth]{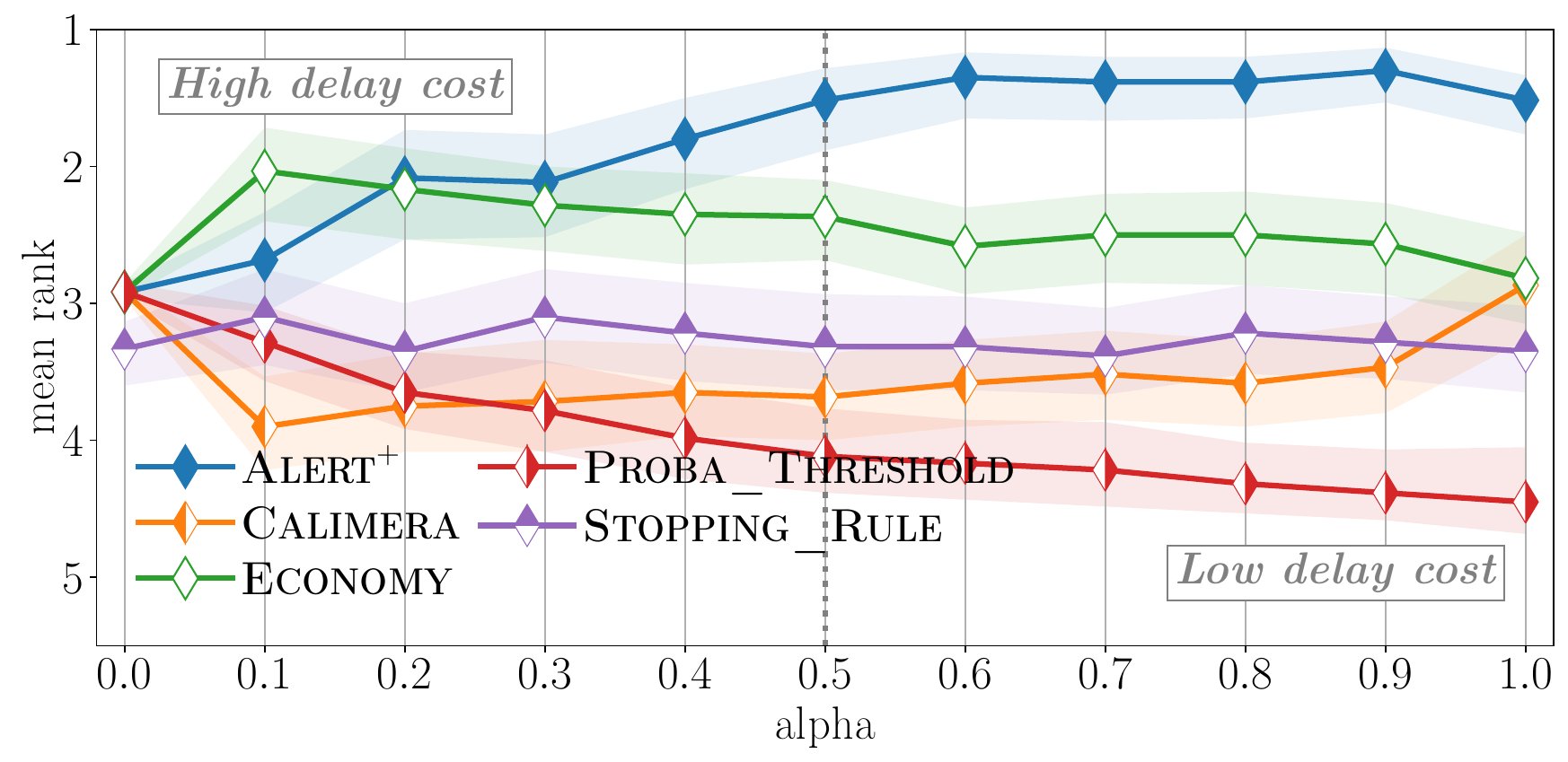}
    \caption{Evolution of the mean ranks, for every $\alpha$, based on the \textit{AvgCost}. Backbone classifier is InceptionTime.}
    \label{fig:bump_inception}
\end{figure}

\textcolor{black}{We also test alternative backbone classifiers, learning a Gradient Boosting Machine (GBM) classifier as well as the deep-learning-based InceptionTime classifier  \citep{ismail2020inceptiontime} on top of the raw time series. The GBM hyperparameters have been kept the same as the default ones used in \texttt{scikit-learn}. To reduce computation time, the number of epochs has been lowered down to 200. Others hyperparameters have been selected such that they offer the best trade-off between performances and efficiency according to the sensitivity studies realized by \cite{ismail2020inceptiontime}. The ranking displayed in both Figure \ref{fig:bump_gbm} and \ref{fig:bump_inception} is generally consistent with one observed using the MiniROCKET classifier. Both \textsc{Alert}$^+$ and \textsc{Economy} rank themselves higher when operating with the InceptionTime classifier. Otherwise, the main trends remain similar, displaying the robustness of \textsc{Alert}$^+$ with respect to the time series classifier at hand.
}

\section{Qualitative results}
\label{qualitative}

\subsection{Visualizing trigger scores}

\begin{figure}[H]
    \centering
    \subfloat[]{\includegraphics[width=0.33\linewidth]{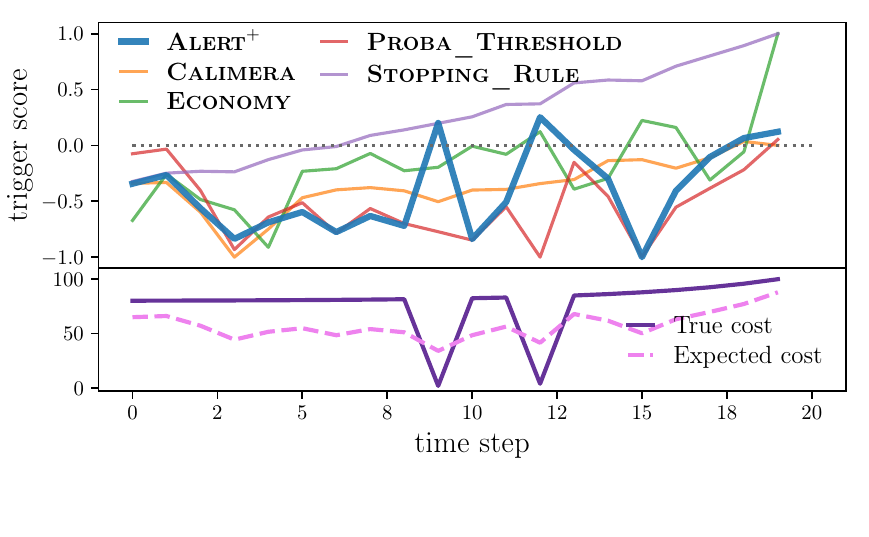}}
    %\vspace{-1.55cm}
    \subfloat[]{\includegraphics[width=0.33\linewidth]{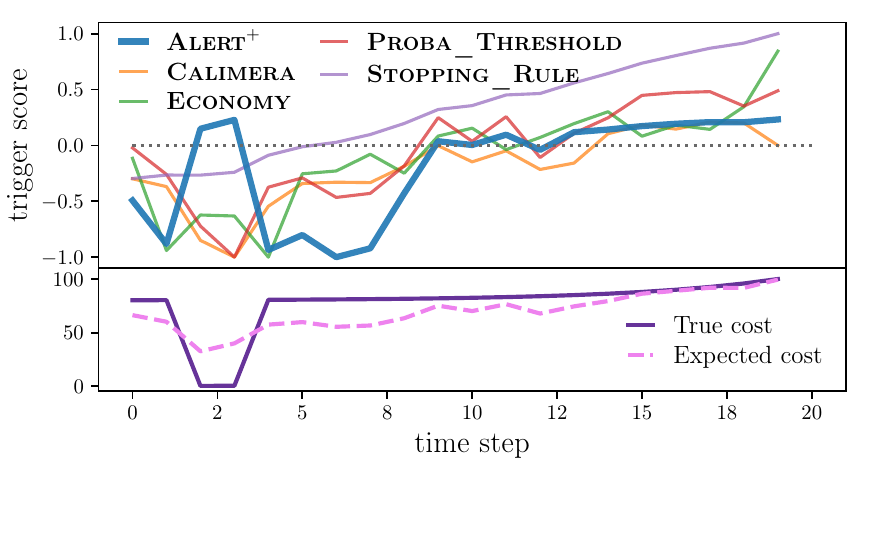}} 
    %\vspace{-1.55cm}
    \subfloat[]{\includegraphics[width=0.33\linewidth]{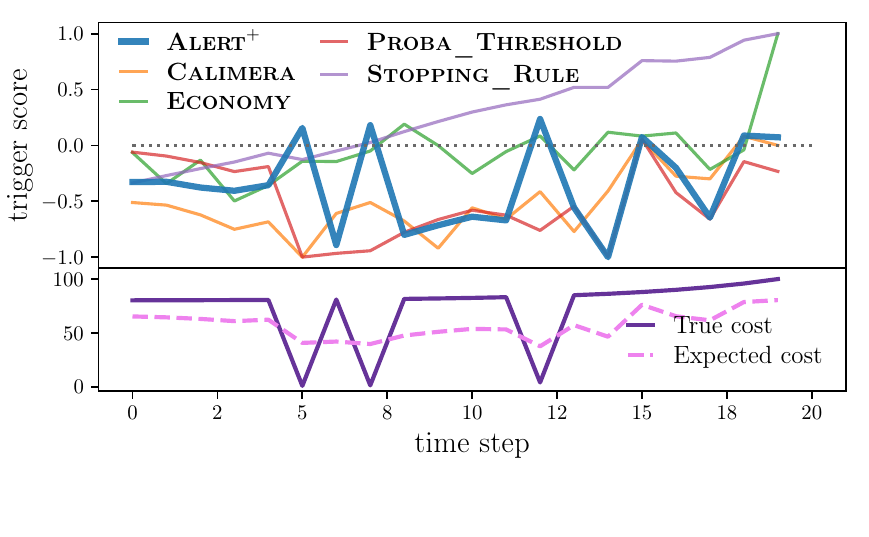}}
    \vspace{-1cm}
    \caption{AluminiumConcentration}
    \label{fig:alu}
\end{figure}

\begin{figure}[H]
    \centering
    \subfloat[]{\includegraphics[width=0.33\linewidth]{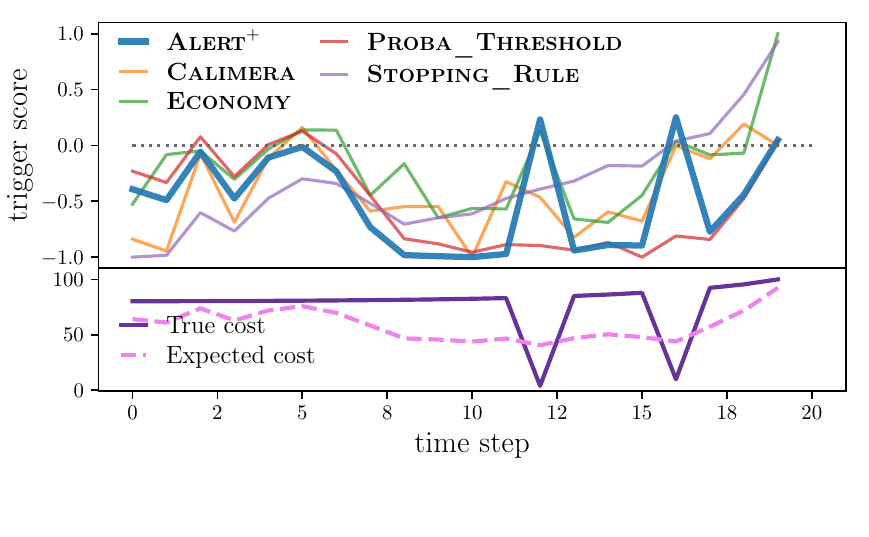}} 
    %\vspace{-1.55cm}
    \subfloat[]{\includegraphics[width=0.33\linewidth]{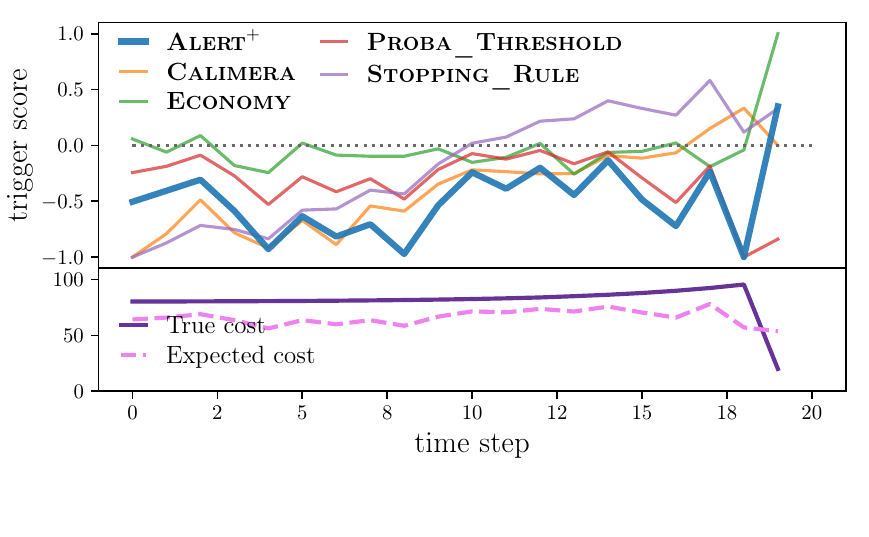}} 
    %\vspace{-1.55cm}
    \subfloat[]{\includegraphics[width=0.33\linewidth]{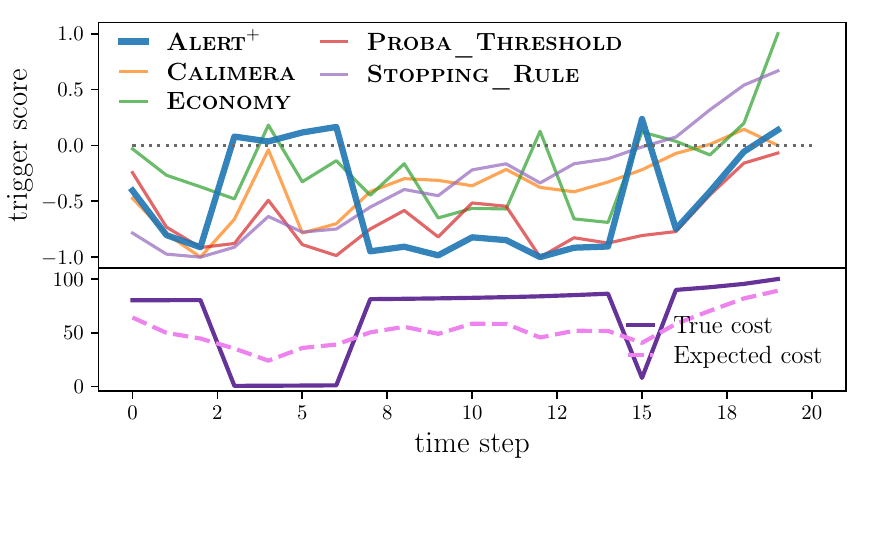}}
    \vspace{-1cm}
    \caption{WindTurbinePower}
    \label{fig:wind}
\end{figure}

\textcolor{black}{
In this section, we provide a visual interpretation of the strong performance achieved by \textsc{Alert}$^+$. To this end, the trigger scores $\{score_t\}_{(1 \leq t \leq T)}$ produced by several trigger models are visualized alongside the true and expected costs, the latter reflecting the confidence of the classifier. For each approach, the trigger score is normalized such that a decision is triggered whenever the score becomes strictly positive. In particular, the scores are computed as follows:
\begin{itemize}
\item \textsc{Alert}$^+$: $score_t = Q(s_t, a_t=1) - Q(s_t, a_t=0)$, corresponding to the difference between the two $Q$-values.
\item \textsc{Calimera}: $score_t = f_t(X_t)$, corresponding to the output of the regressor specialized for time step $t$.
\item \textsc{Economy}: $score_t = \min_{_{(t+1 \leq t' \leq T)}} {\hat{C}(t')} - \hat{C}(t)$, defined as the difference between the minimum forecasted cost and the estimated cost incurred at the current time step.
\item \textsc{Proba-Threshold}: $score_t = \max_{y \in \mathcal{Y}} p(y| \mathbf{x}_t) - \hat{\delta}$, representing the difference between the maximum posterior probability and the learned threshold.
\item \textsc{Stopping Rule}: $score_t = \hat{\gamma}_1 p_1 + \hat{\gamma}_2 p_2 + \hat{\gamma}_3 \frac{t}{T}$, corresponding to the \textsc{Stopping Rule} criterion as is.
\end{itemize}
Then, all of these scores are divided by the absolute maximum to get all of them in the $[-1, 1]$ range. For eight of the benchmark datasets, on which \textsc{Alert}$^+$ outperforms its competitors, we plot the top three test instances exhibiting the largest difference in incurred cost between \textsc{Alert}$^+$ and the average performance of competing methods. We set $\alpha = 0.8$, corresponding to the trade-off value for which \textsc{Alert}$^+$ achieves its best performance. Figures \ref{fig:alu}--\ref{fig:wind} generally illustrate scenarios in which the cost-optimal decision points are narrow and therefore difficult to identify. In these cases, \textsc{Alert}$^+$ demonstrates a stronger ability to capture complex patterns in classifier uncertainties and errors, which helps explain its superior performance relative to state-of-the-art approaches.}

\begin{figure}[H]
    \centering
    \subfloat[]{\includegraphics[width=0.33\linewidth]{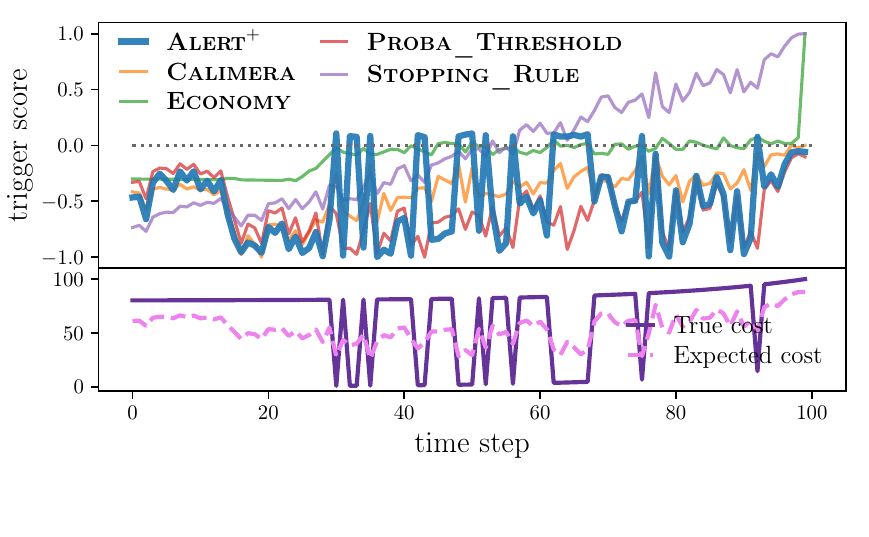}}
    %\vspace{-1.55cm}
    \subfloat[]{\includegraphics[width=0.33\linewidth]{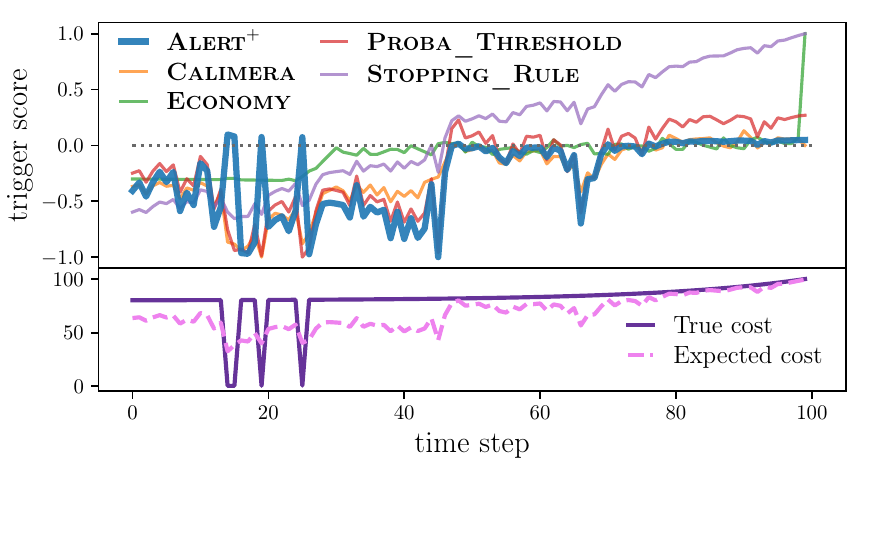}}
    %\vspace{-1.55cm}
    \subfloat[]{\includegraphics[width=0.33\linewidth]{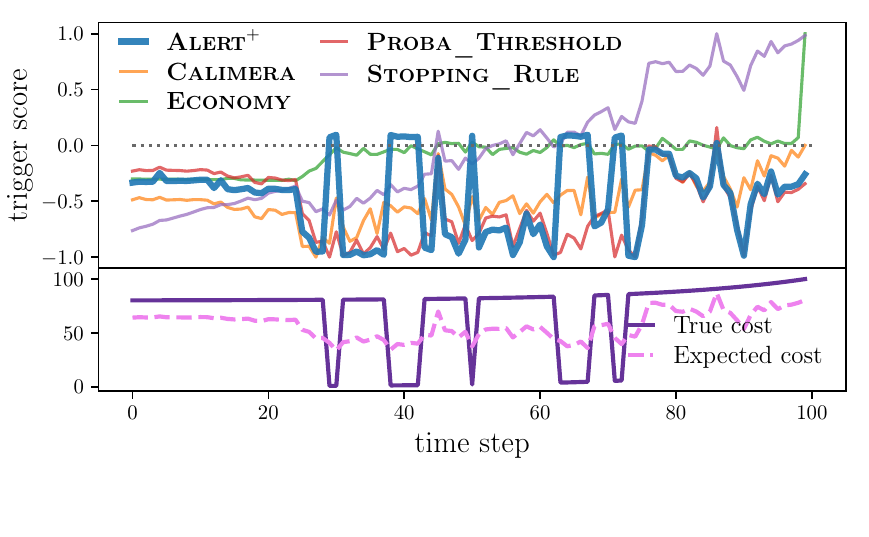}}
    \vspace{-1cm}
    \caption{AluminiumConcentration}
    \label{fig:alu_100}
\end{figure}

\begin{figure}[H]
    \centering
    \subfloat[]{\includegraphics[width=0.33\linewidth]{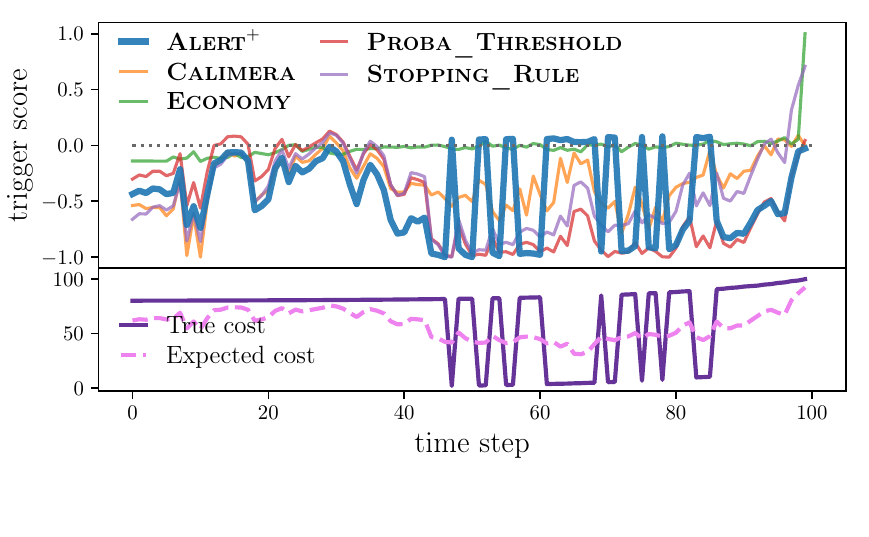}} 
    %\vspace{-1.55cm}
    \subfloat[]{\includegraphics[width=0.33\linewidth]{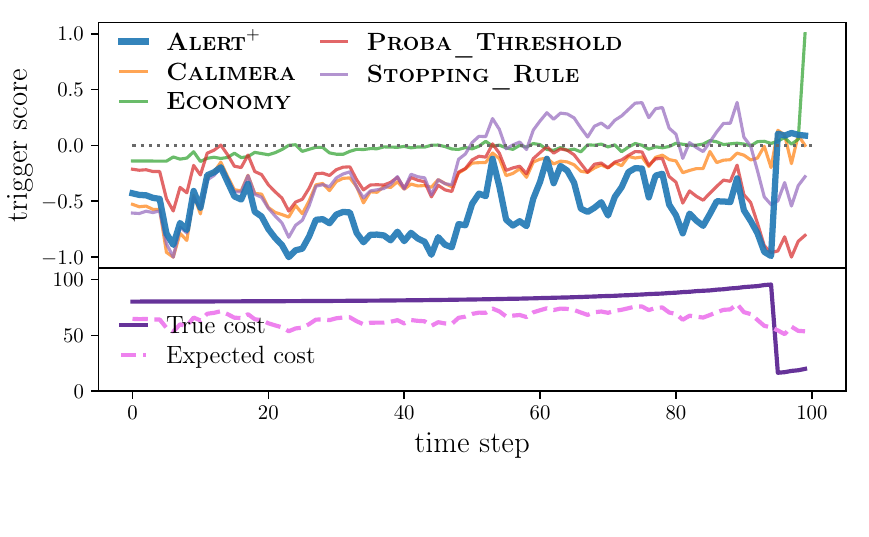}} 
    %\vspace{-1.55cm}
    \subfloat[]{\includegraphics[width=0.33\linewidth]{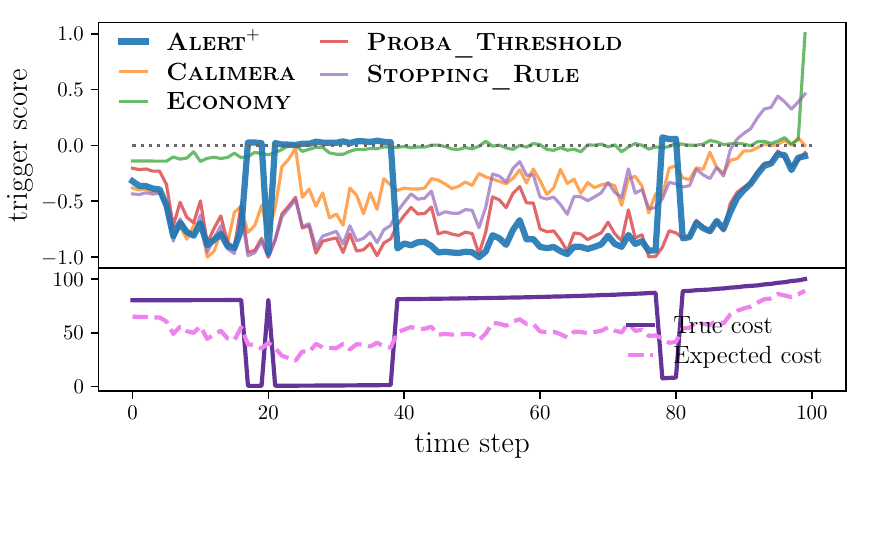}}
    \vspace{-1cm}
    \caption{WindTurbinePower}
    \label{fig:wind_100}
\end{figure}

\vspace{3cm}
\textcolor{black}{In Section \ref{examples_freq}, the same examples are revisited using a higher sampling frequency, with a classifier trained at each $1\%$ increment of the time series. This setting ensures that the original sampling rate of $5\%$ is not overly coarse, which could otherwise lead to overly punctual cost optima, thus artificially favoring \textsc{Alert}$^+$ against competing approaches. As shown in Figures \ref{fig:alu_100}--\ref{fig:wind_100}, the superior capabilities of \textsc{Alert}$^+$ are confirmed, and the initial analysis is further strengthened. Indeed, several examples exhibit substantially more complex patterns under this finer temporal discretization, yet \textsc{Alert}$^+$ continues to handle these scenarios effectively. These examples are thus consistent with results displayed in Section \ref{app_sampling}.
}
% In Section \ref{examples_freq}, the same examples are examined with a higher sampling frequency, learning a classifier each of the $1\%$ of the time series. This way, we ensure that the original sampling ratio of $5\%$ isn't too rough, which would eventually make cost optimum punctual. In Figures \ref{fig:alu_100}--\ref{fig:wind_100}, the higher capabilities of \textsc{Alert}$^+$ are confirmed, original analysis are even strengthen: some of the examples displaying way more complex patterns to learn with this more fine-grained time discretization, and \textsc{Alert}$^+$ still succeeding in handling those. 

\captionsetup[subfloat]{labelformat=empty}
% \begin{figure}[h]
%     \centering
%     \subfloat[]{\includegraphics[width=0.5\linewidth]{plots/qualitative/trigger_score/HouseholdPowerConsumption1/281.pdf}}
%     \subfloat[]{\includegraphics[width=0.5\linewidth]{plots/qualitative/trigger_score/HouseholdPowerConsumption1/283.pdf}} \\
%     \vspace{-1cm}
%     \subfloat[]{\includegraphics[width=0.5\linewidth]{plots/qualitative/trigger_score/HouseholdPowerConsumption1/330.pdf}} 
%     \subfloat[]{\includegraphics[width=0.5\linewidth]{plots/qualitative/trigger_score/HouseholdPowerConsumption1/401.pdf}}
%     \vspace{-1cm}
%     \caption{HouseholdPowerConsumption1}
%     \label{fig:placeholder}
% \end{figure}

\begin{figure}[H]
    \centering
    \subfloat[]{\includegraphics[width=0.33\linewidth]{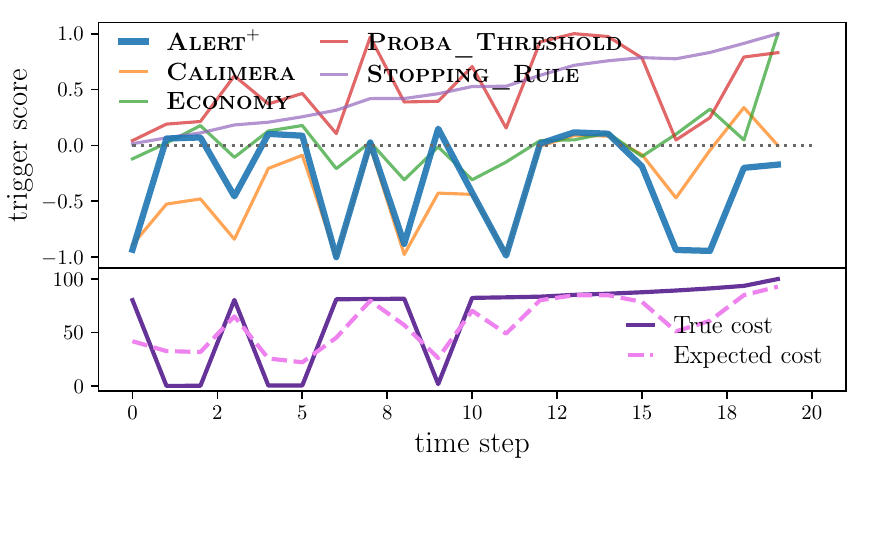}} 
    %\vspace{-1.55cm}
    \subfloat[]{\includegraphics[width=0.33\linewidth]{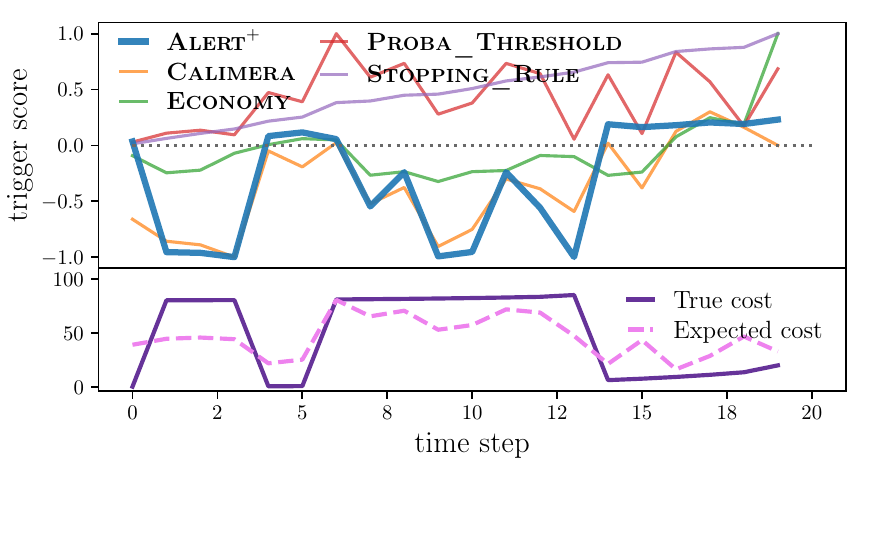}} 
    %\vspace{-1.55cm}
    \subfloat[]{\includegraphics[width=0.33\linewidth]{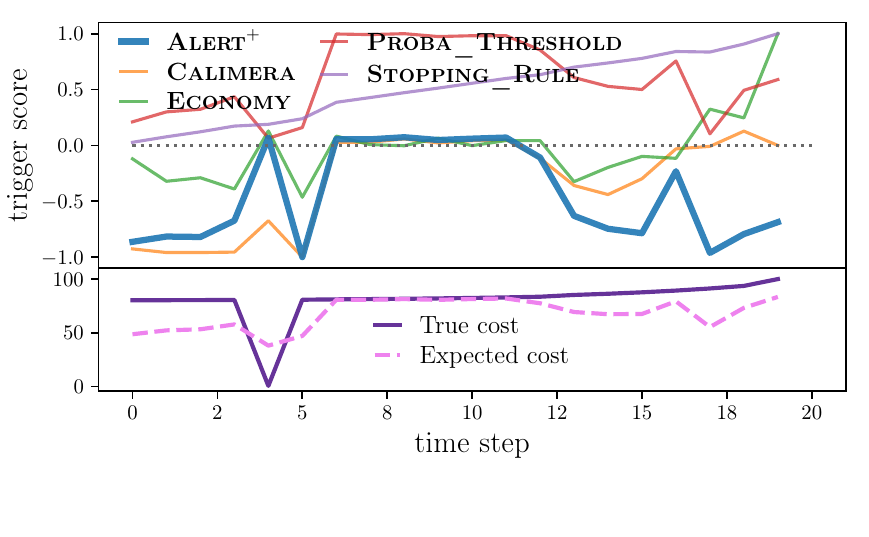}}
    \vspace{-1cm}
    \caption{DhakaHourlyAirQuality}
    \label{fig:placeholder}
\end{figure}

\begin{figure}[H]
    \centering
    \subfloat[]{\includegraphics[width=0.33\linewidth]{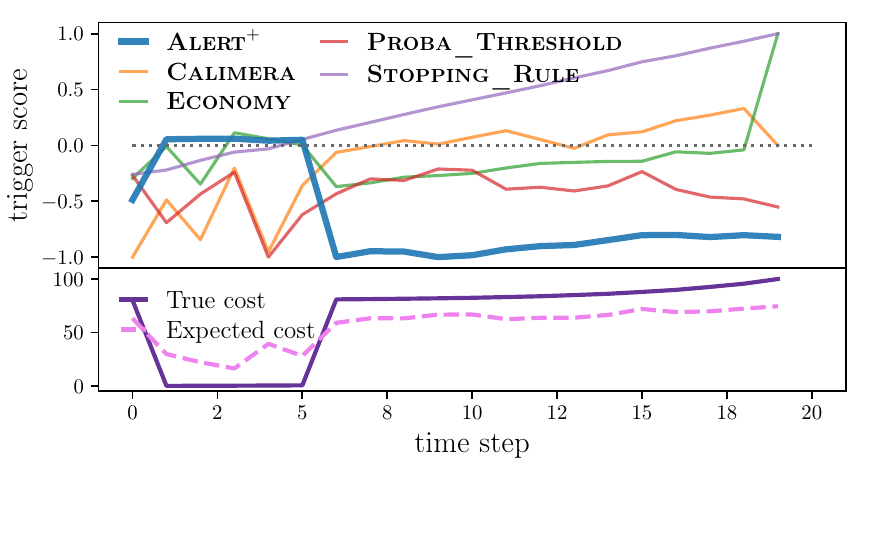}} 
    %\vspace{-1.55cm}
    \subfloat[]{\includegraphics[width=0.33\linewidth]{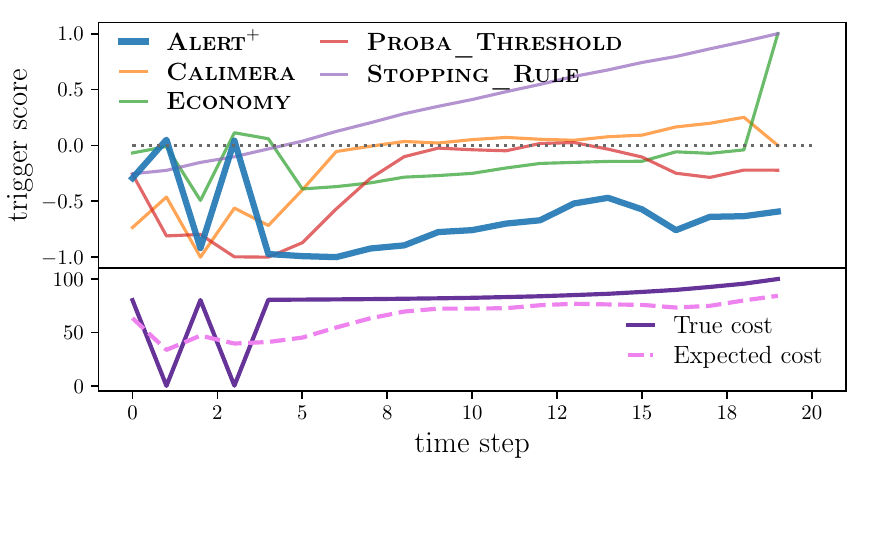}} 
    %\vspace{-1.55cm}
    \subfloat[]{\includegraphics[width=0.33\linewidth]{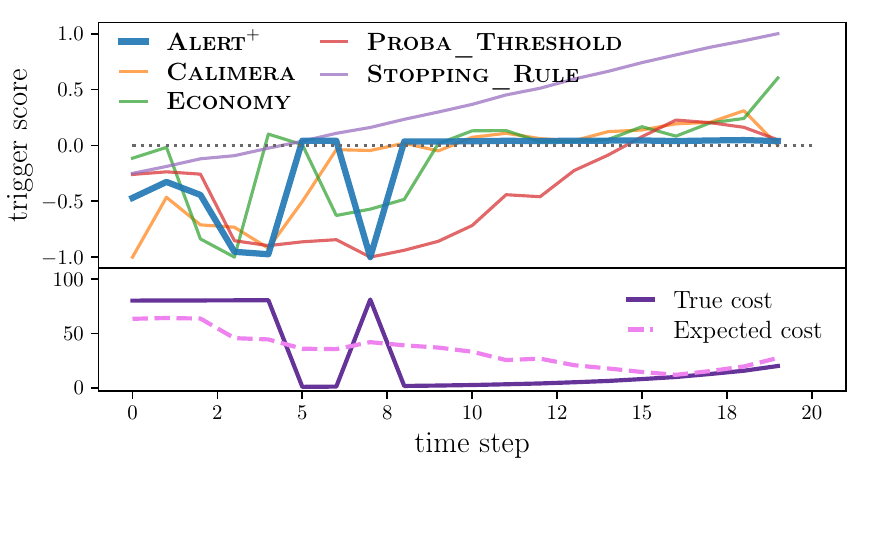}}
    \vspace{-1cm}
    \caption{GunPointAgeSpan}
    \label{fig:placeholder}
\end{figure}

\begin{figure}[H]
    \centering
    \subfloat[]{\includegraphics[width=0.33\linewidth]{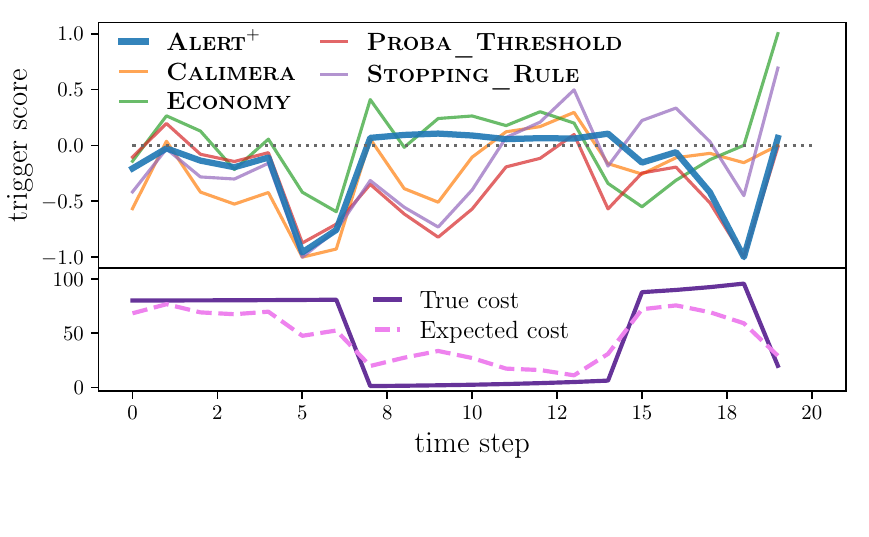}}
    %\vspace{-1.55cm}
    \subfloat[]{\includegraphics[width=0.33\linewidth]{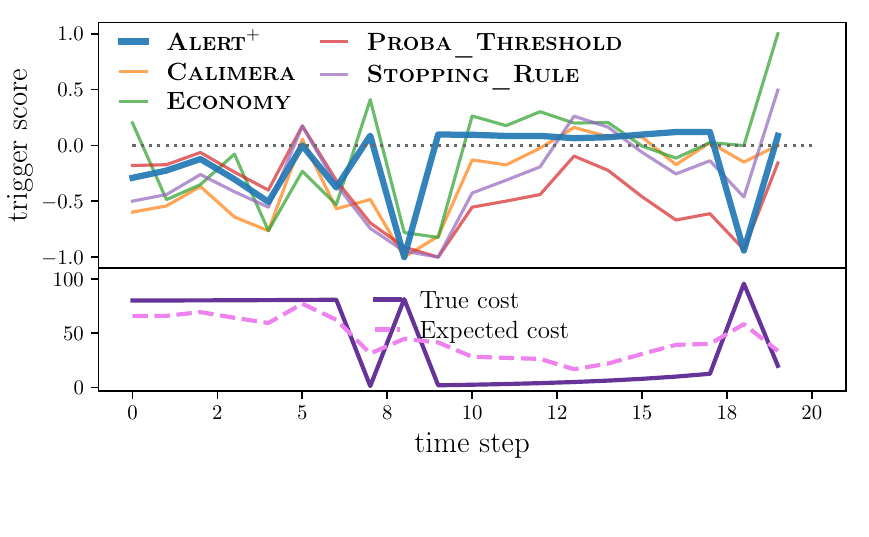}}
    %\vspace{-1.55cm}
    \subfloat[]{\includegraphics[width=0.33\linewidth]{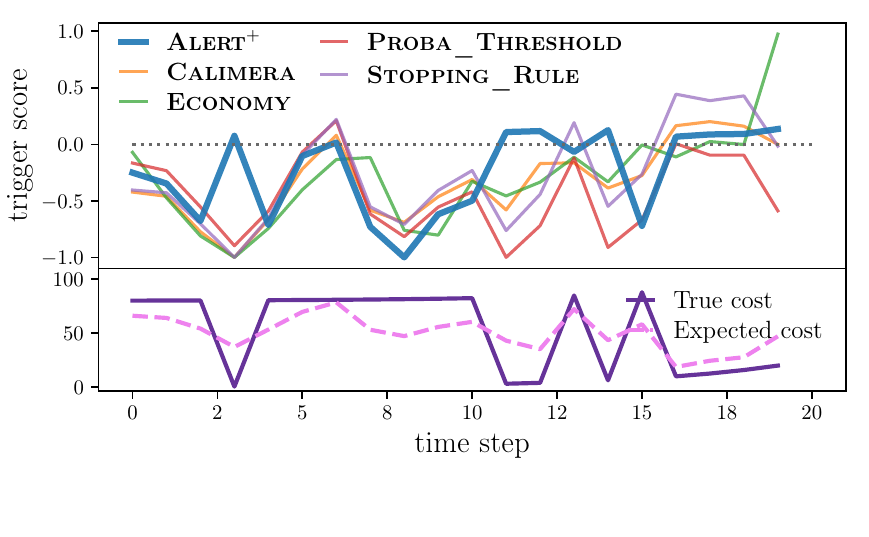}}
    \vspace{-1cm}
    \caption{HouseholdPowerConsumption1}
    \label{fig:placeholder}
\end{figure}

\begin{figure}[H]
    \centering
    \subfloat[]{\includegraphics[width=0.33\linewidth]{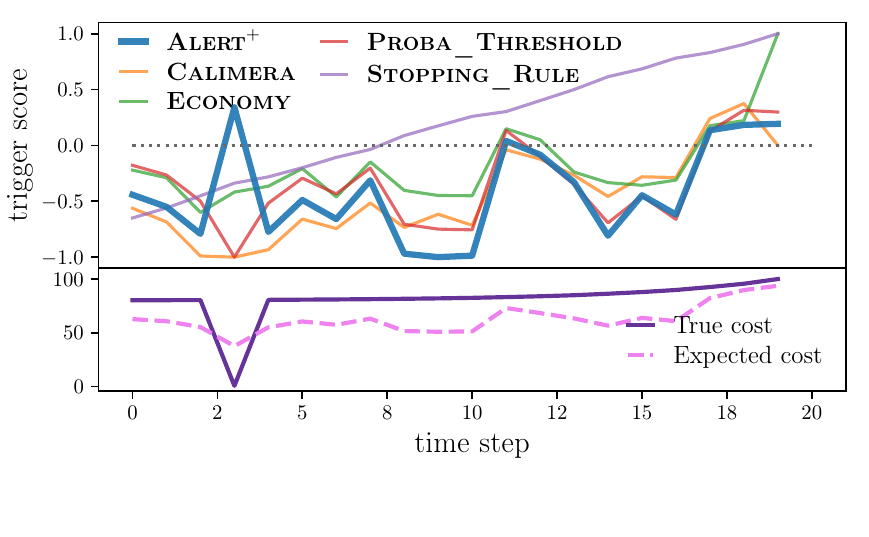}} 
    %\vspace{-1.55cm}
    \subfloat[]{\includegraphics[width=0.33\linewidth]{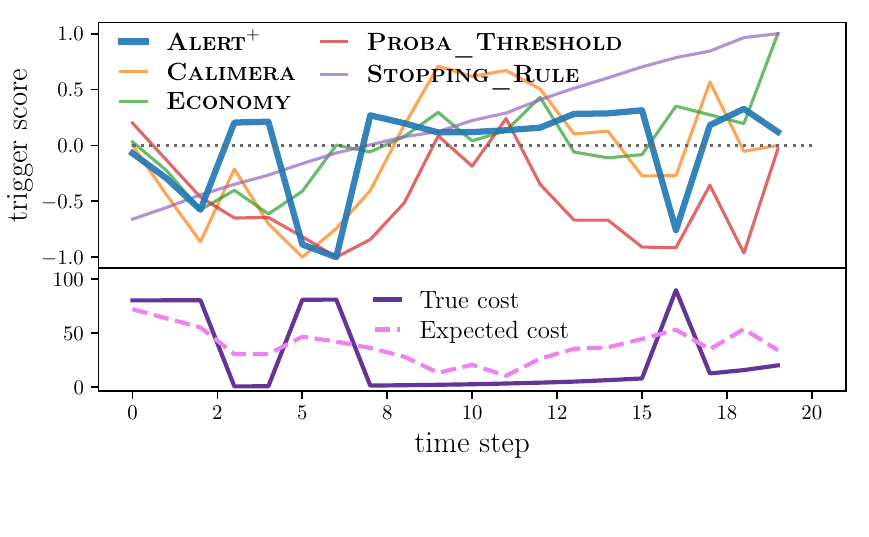}} 
    %\vspace{-1.55cm}
    \subfloat[]{\includegraphics[width=0.33\linewidth]{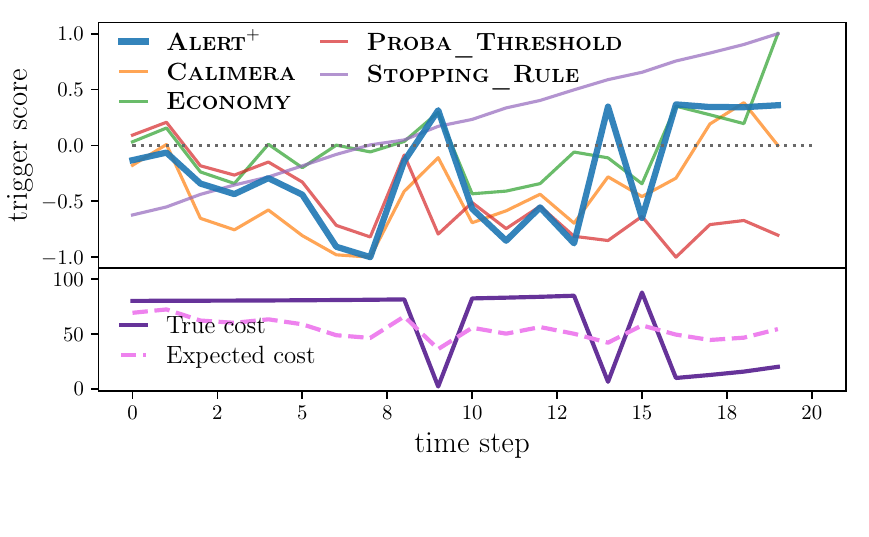}}
    \vspace{-1cm}
    \caption{MadridPM10Quality}
    \label{fig:placeholder}
\end{figure}

\begin{figure}[H]
    \centering
    \subfloat[]{\includegraphics[width=0.33\linewidth]{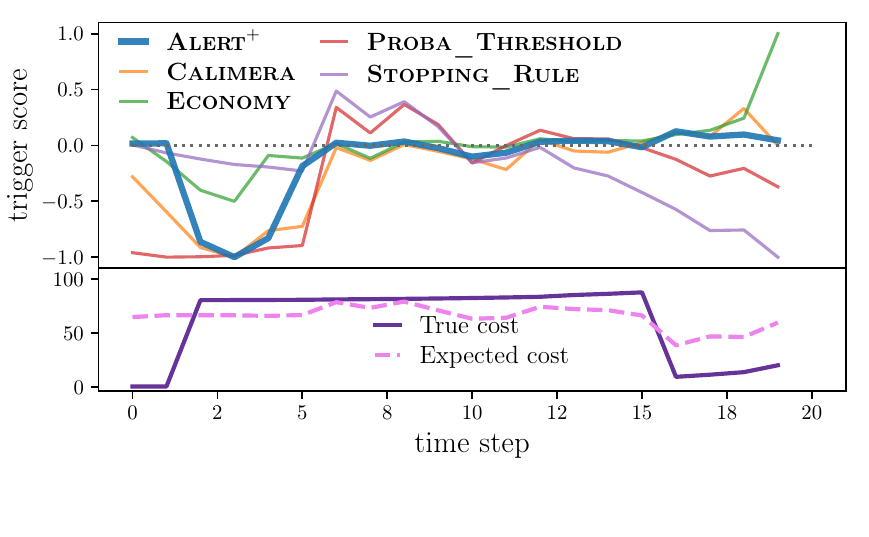}} 
    %\vspace{-1.55cm}
    \subfloat[]{\includegraphics[width=0.33\linewidth]{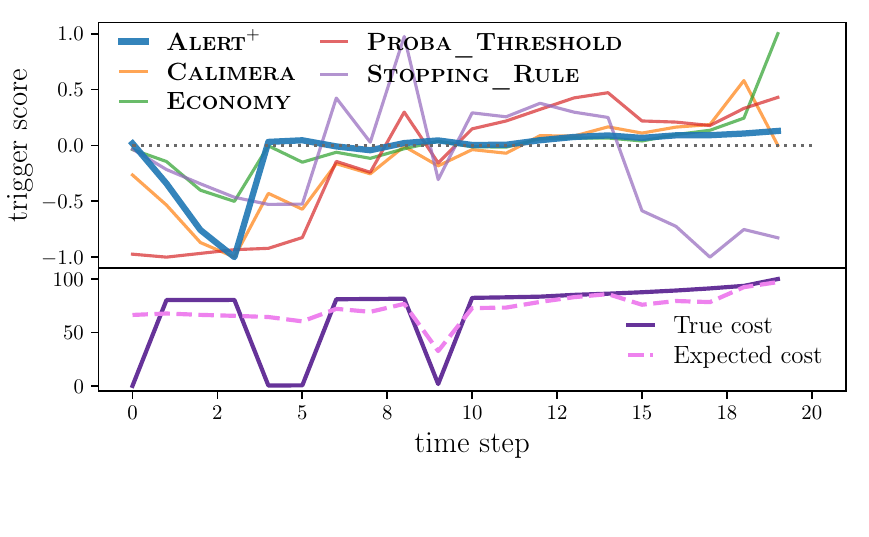}} 
    %\vspace{-1.55cm}
    \subfloat[]{\includegraphics[width=0.33\linewidth]{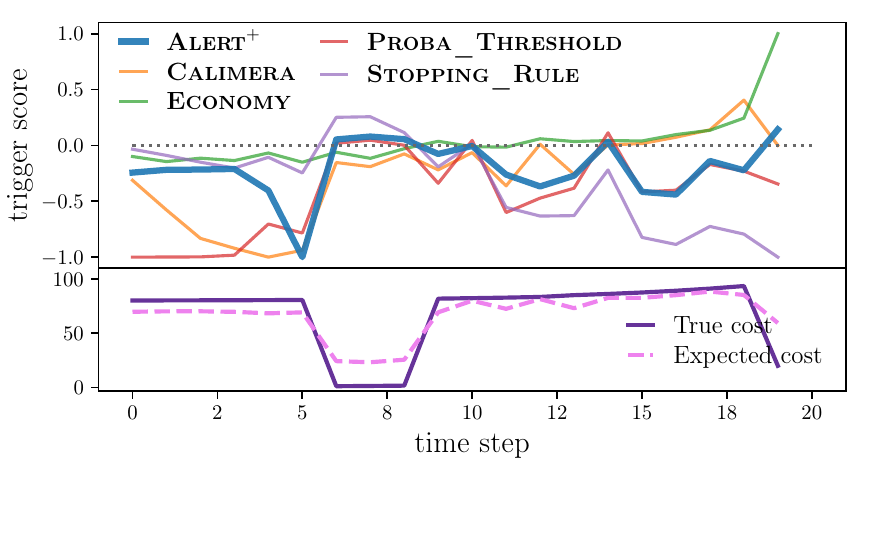}}
    \vspace{-1cm}
    \caption{MelbournePedestrian}
    \label{fig:placeholder}
\end{figure}

\begin{figure}[H]
    \centering
    \subfloat[]{\includegraphics[width=0.33\linewidth]{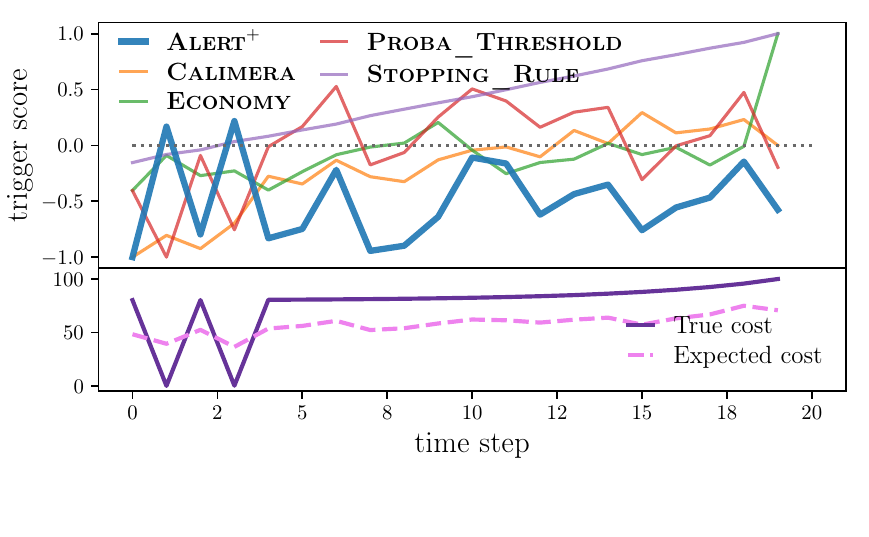}} 
    %\vspace{-1.55cm}
    \subfloat[]{\includegraphics[width=0.33\linewidth]{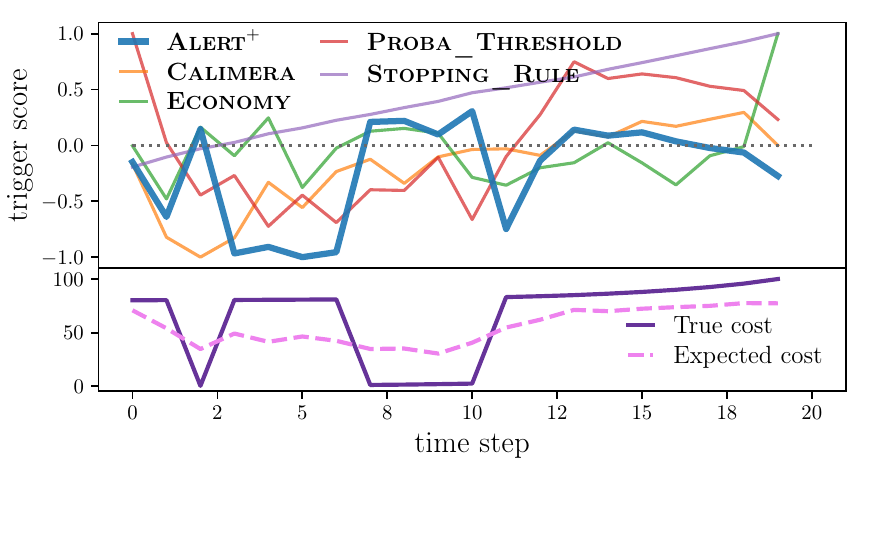}} 
    %\vspace{-1.55cm}
    \subfloat[]{\includegraphics[width=0.33\linewidth]{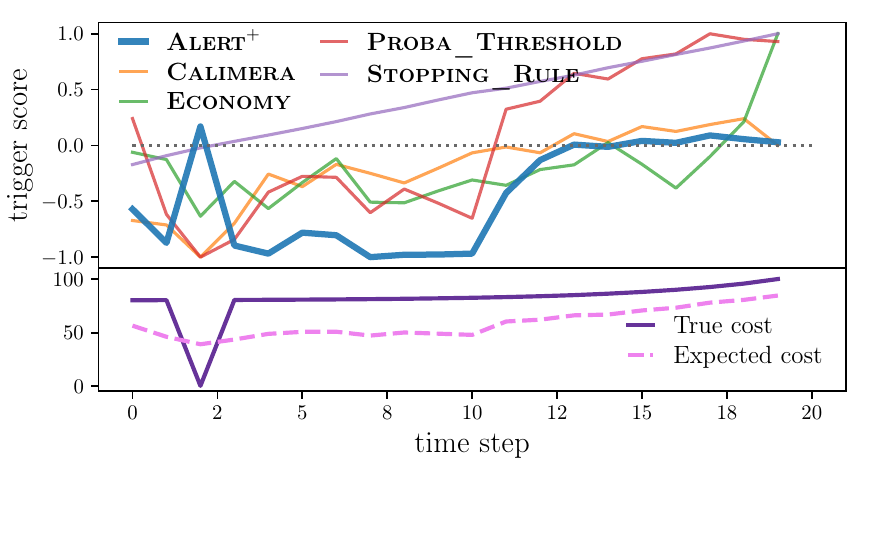}}
    \vspace{-1cm}
    \caption{PrecipitationAndalusia}
    \label{fig:placeholder}
\end{figure}

\subsection{Robustness to sampling frequency}
\label{examples_freq}

\begin{figure}[H]
    \centering
    \subfloat[]{\includegraphics[width=0.33\linewidth]{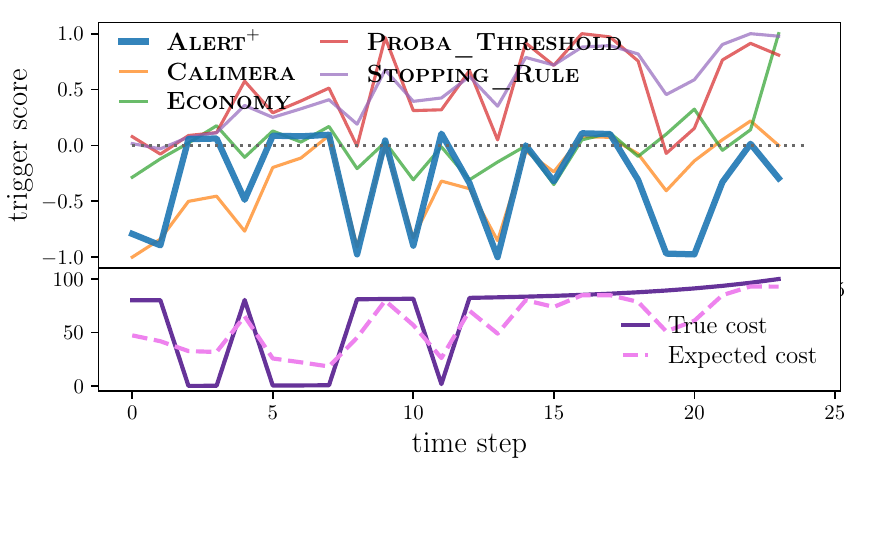}}
    %\vspace{-1.55cm}
    \subfloat[]{\includegraphics[width=0.33\linewidth]{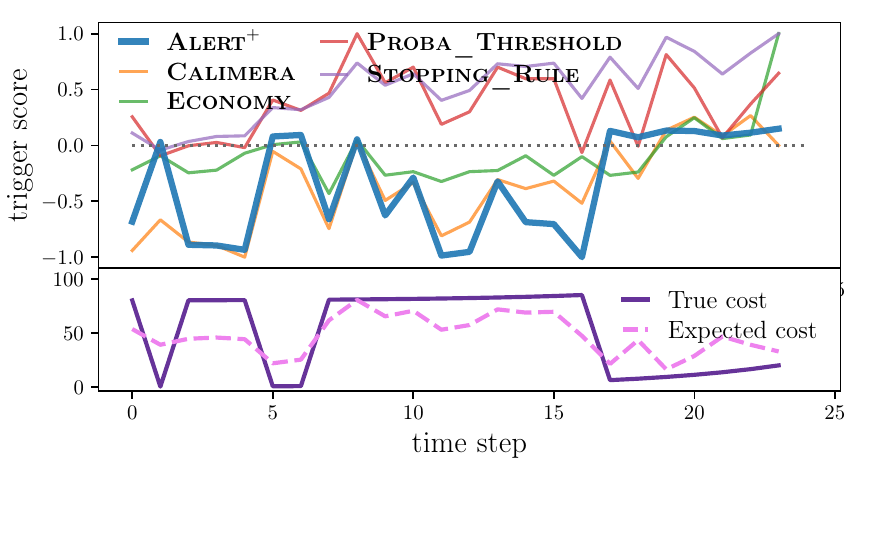}}
    %\vspace{-1.55cm}
    \subfloat[]{\includegraphics[width=0.33\linewidth]{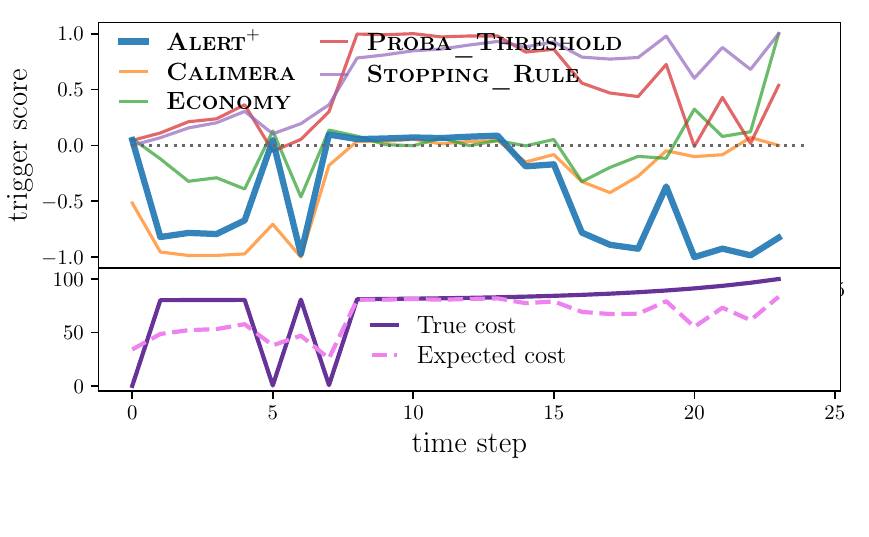}}
    \vspace{-1cm}
    \caption{DhakaHourlyAirQuality}
    \label{fig:placeholder}
\end{figure}

\begin{figure}[H]
    \centering
    \subfloat[]{\includegraphics[width=0.33\linewidth]{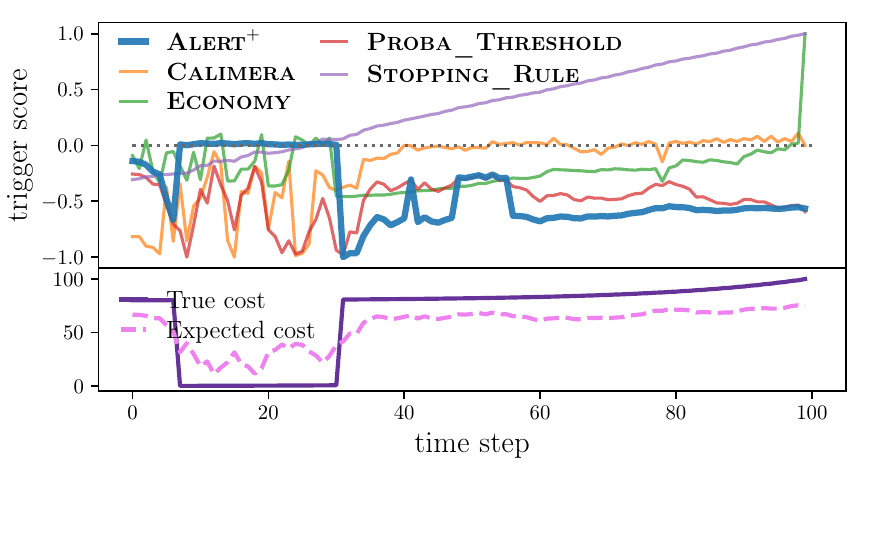}} 
    %\vspace{-1.55cm}
    \subfloat[]{\includegraphics[width=0.33\linewidth]{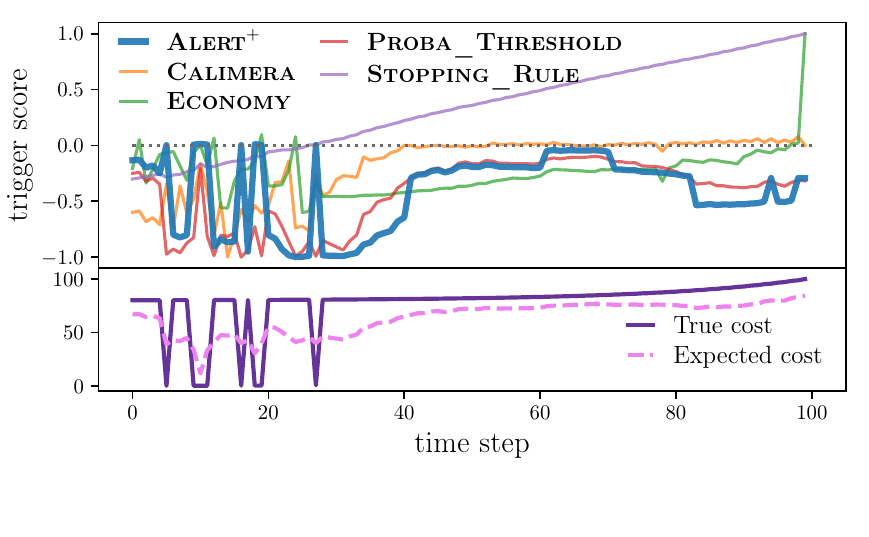}} 
    %\vspace{-1.55cm}
    \subfloat[]{\includegraphics[width=0.33\linewidth]{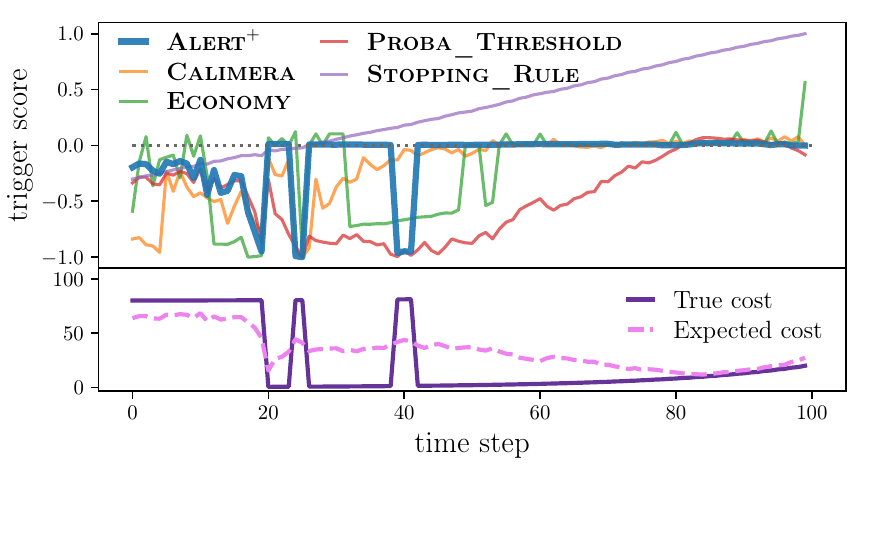}}
    \vspace{-1cm}
    \caption{GunPointAgeSpan}
    \label{fig:placeholder}
\end{figure}

\begin{figure}[H]
    \centering
    \subfloat[]{\includegraphics[width=0.33\linewidth]{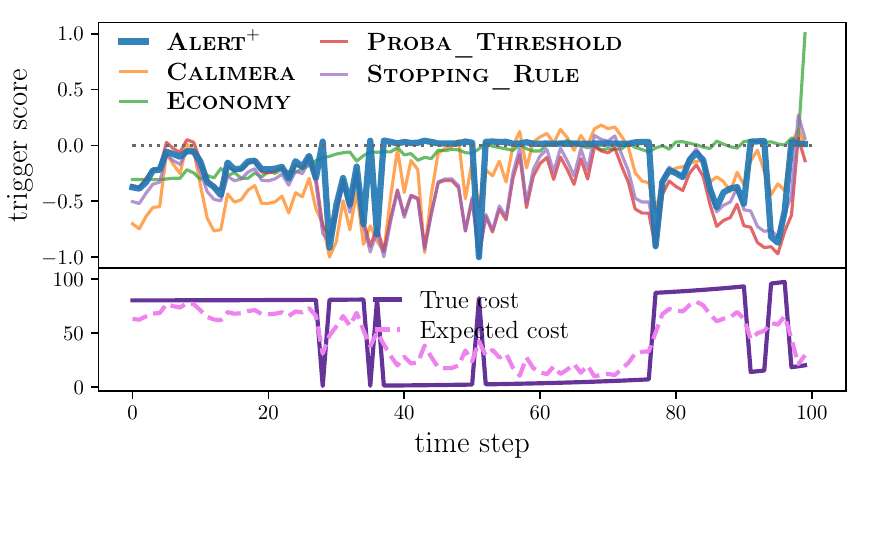}}
    %\vspace{-1.55cm}
    \subfloat[]{\includegraphics[width=0.33\linewidth]{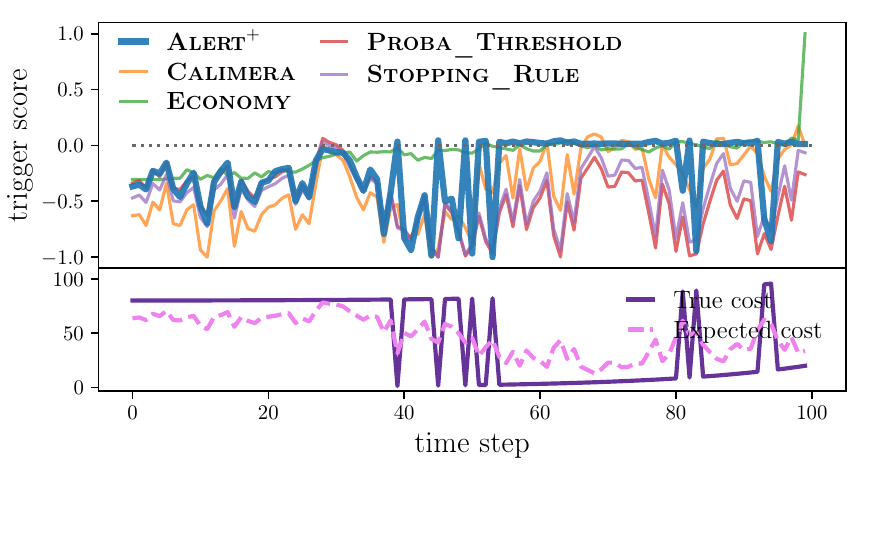}} 
    %\vspace{-1.55cm}
    \subfloat[]{\includegraphics[width=0.33\linewidth]{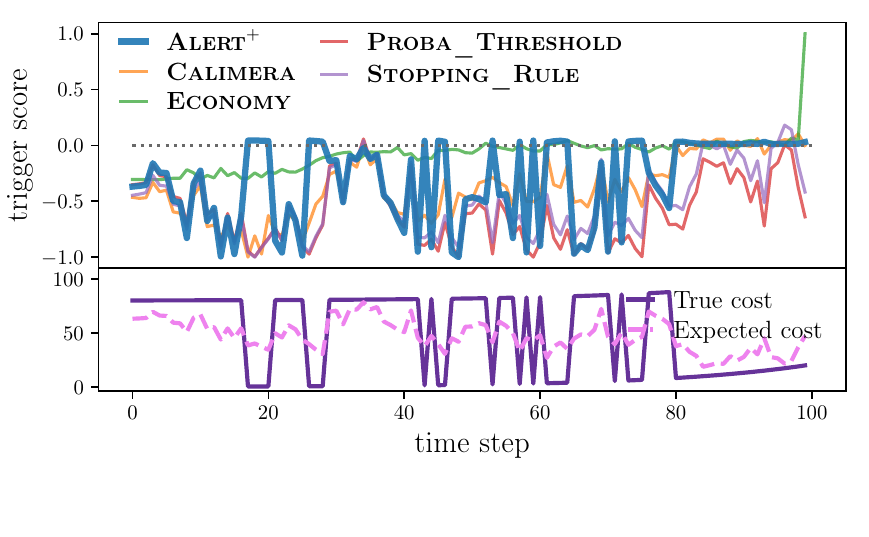}}
    \vspace{-1cm}
    \caption{HouseholdPowerConsumption1}
    \label{fig:placeholder}
\end{figure}

\begin{figure}[H]
    \centering
    \subfloat[]{\includegraphics[width=0.33\linewidth]{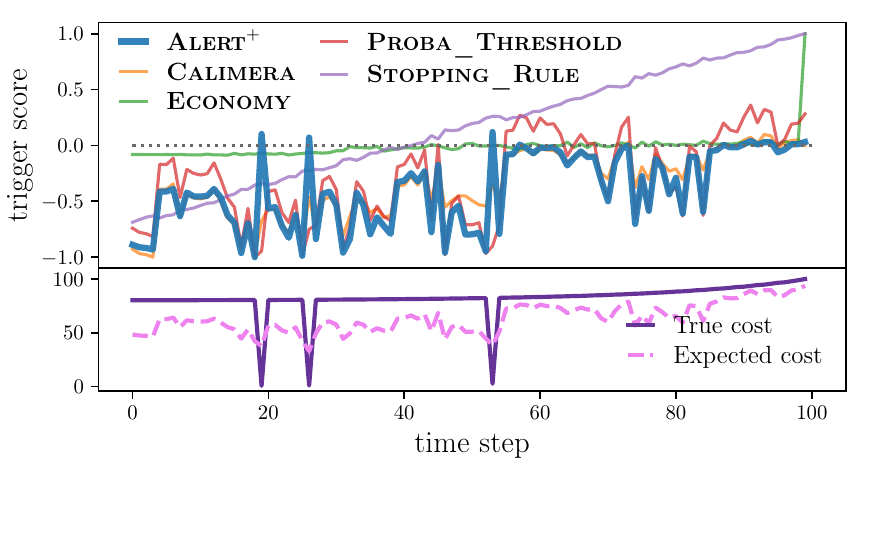}} 
    %\vspace{-1.55cm}
    \subfloat[]{\includegraphics[width=0.33\linewidth]{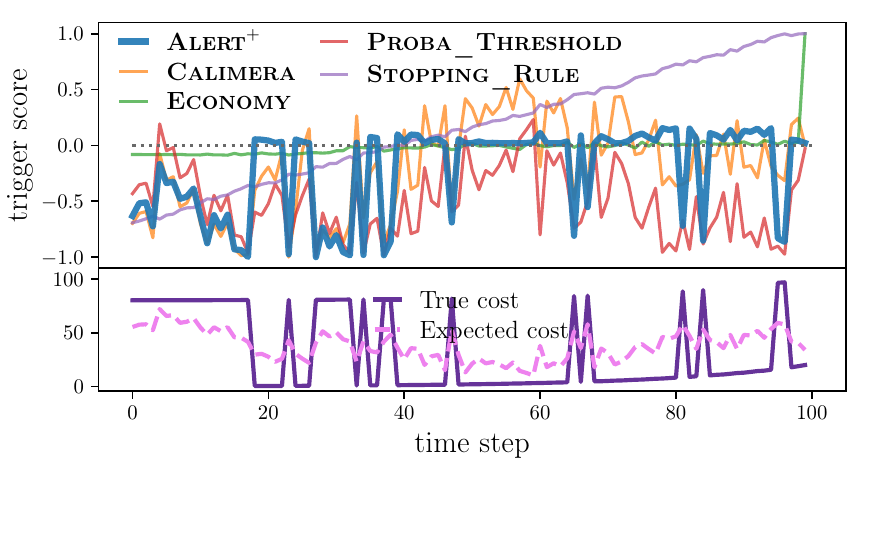}} 
    %\vspace{-1.55cm}
    \subfloat[]{\includegraphics[width=0.33\linewidth]{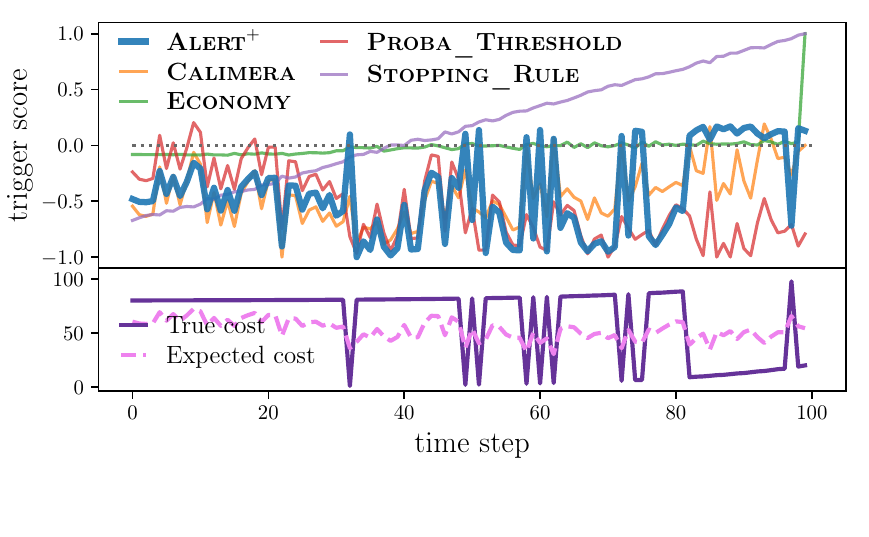}}
    \vspace{-1cm}
    \caption{MadridPM10Quality}
    \label{fig:placeholder}
\end{figure}

\begin{figure}[H]
    \centering
    \subfloat[]{\includegraphics[width=0.33\linewidth]{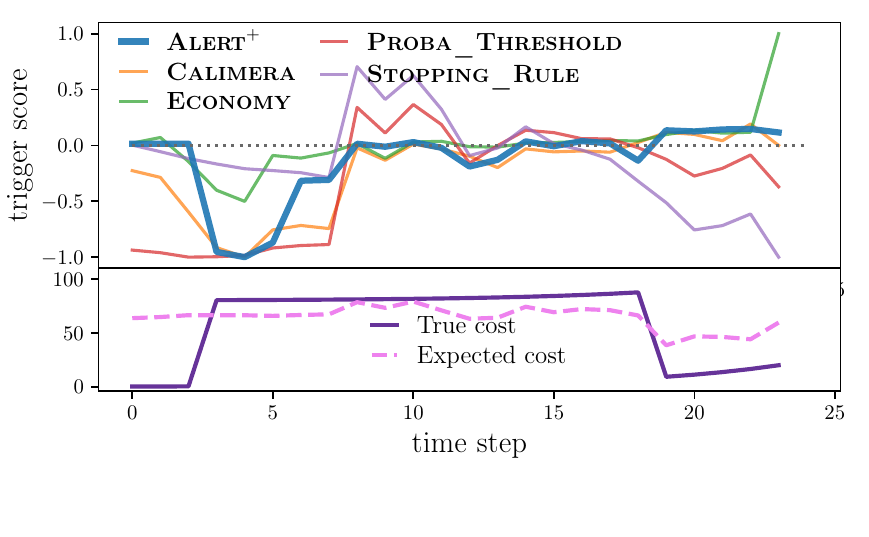}} 
    %\vspace{-1.55cm}
    \subfloat[]{\includegraphics[width=0.33\linewidth]{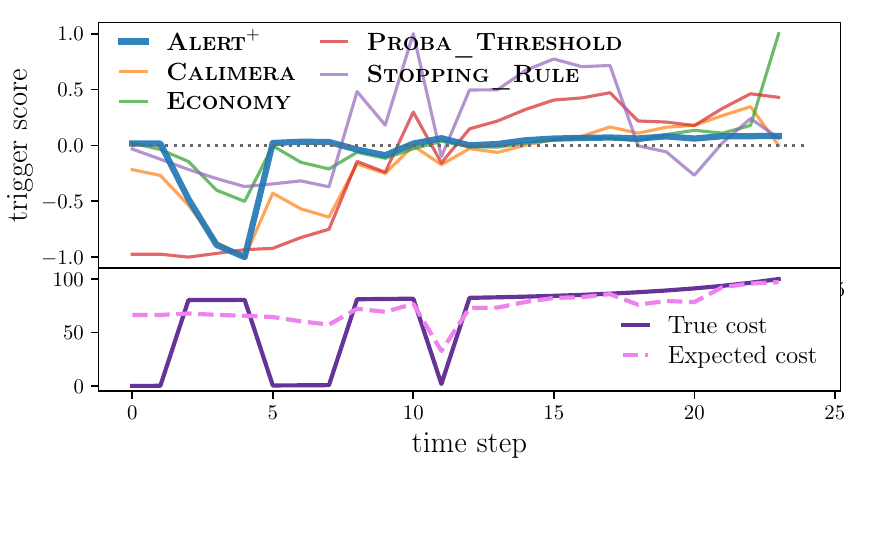}} 
    %\vspace{-1.55cm}
    \subfloat[]{\includegraphics[width=0.33\linewidth]{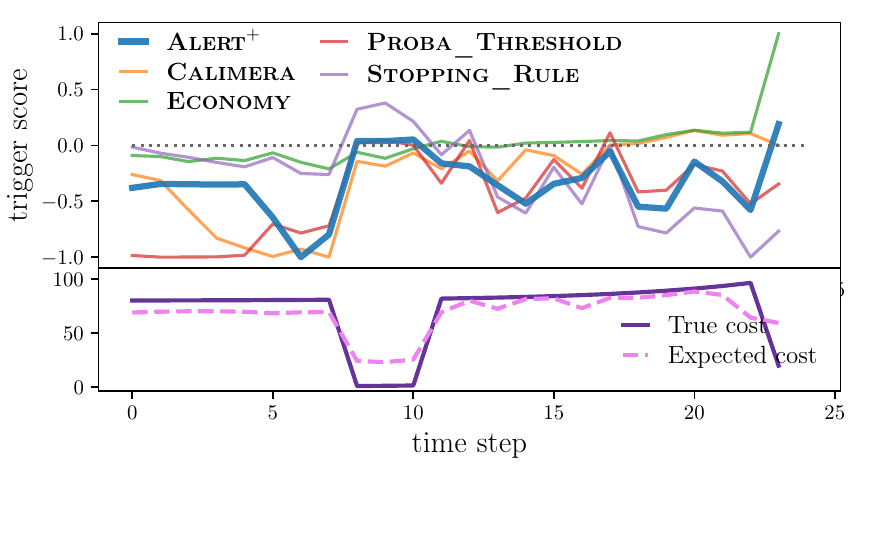}}
    \vspace{-1cm}
    \caption{MelbournePedestrian}
    \label{fig:placeholder}
\end{figure}

\begin{figure}[H]
    \centering
    \subfloat[]{\includegraphics[width=0.33\linewidth]{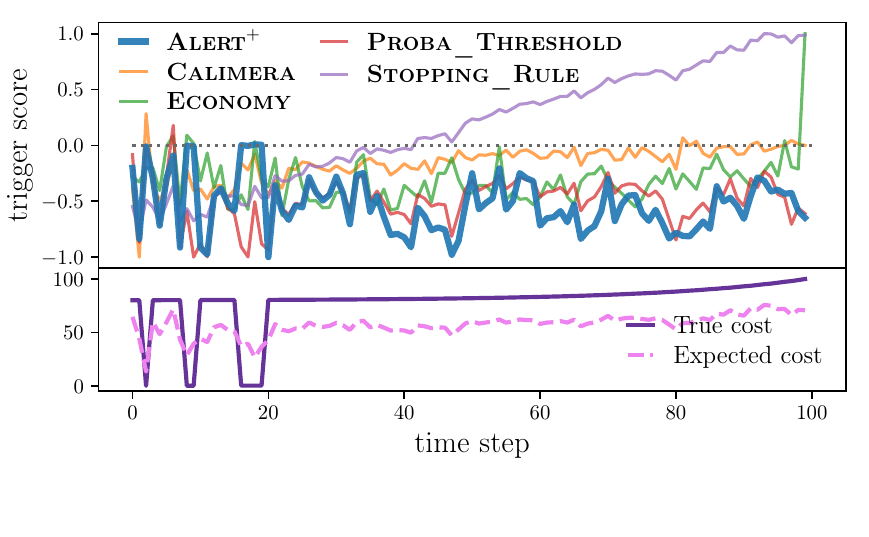}} 
    %\vspace{-1.55cm}
    \subfloat[]{\includegraphics[width=0.33\linewidth]{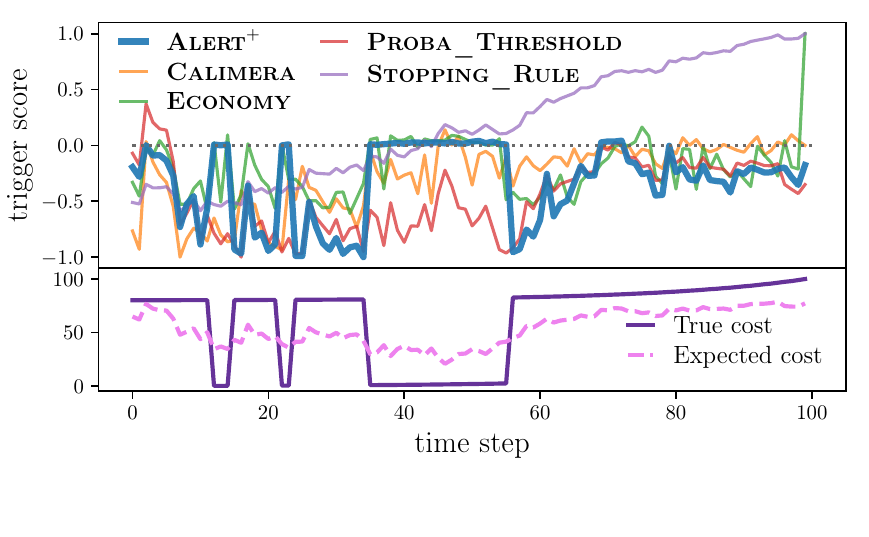}} 
    %\vspace{-1.55cm}
    \subfloat[]{\includegraphics[width=0.33\linewidth]{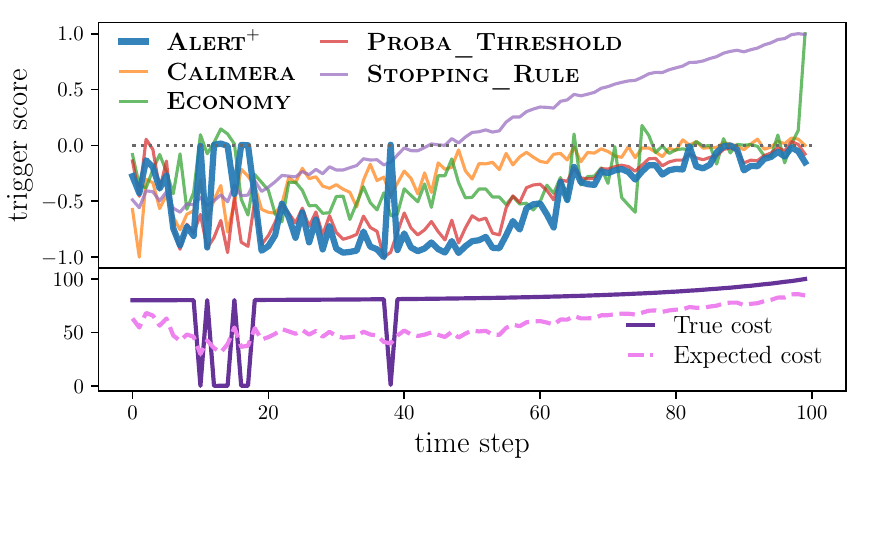}}
    \vspace{-1cm}
    \caption{PrecipitationAndalusia}
    \label{fig:placeholder}
\end{figure}

\end{document}